\documentclass[10pt,twocolumn,letterpaper]{article}

\usepackage{3dv}
\usepackage{times}
\usepackage{epsfig}
\usepackage{graphicx}
\usepackage{amsmath}
\usepackage{amssymb}

\usepackage[table]{xcolor}
\definecolor{col1}{RGB}{232, 161, 148}
\definecolor{col2}{RGB}{148, 187, 232}

\def\eg{\textit{e.g.}~}
\def\ie{\textit{i.e.}}

\def\etal{\textit{et al.}}

\usepackage{verbatim}

\usepackage[toc,page]{appendix}

\usepackage[pagebackref=true,breaklinks=true,letterpaper=true,colorlinks,bookmarks=false]{hyperref}

\threedvfinalcopy 

\def\threedvPaperID{107} 

\ifthreedvfinal\fi
\begin{document}

\title{Channel-Wise Attention-Based Network for \\ Self-Supervised Monocular Depth Estimation}

\author{Jiaxing Yan
\qquad
Hong Zhao 
\qquad
Penghui Bu 
\qquad
YuSheng Jin 
\\
\vspace{-14pt}
\\
State Key Laboratory for Manufacturing System Engineering, \\ School of Mechanical Engineering, Xi’an Jiaotong University, China\\
{\tt\small 
\{yanjxedu@stu., zhaohong@mail., bph19891027@stu.,  yushengJ@stu.\}xjtu.edu.cn
}
}

\maketitle
\thispagestyle{empty}

\begin{abstract}
Self-supervised learning has shown very promising results for monocular depth estimation. Scene structure and local details both are significant clues for high-quality depth estimation. Recent works suffer from the lack of explicit modeling of scene structure and proper handling of details information, which leads to a performance bottleneck and blurry artefacts in predicted results. In this paper, we propose the Channel-wise Attention-based Depth Estimation Network (CADepth-Net) with two effective contributions: 1) The structure perception module employs the self-attention mechanism to capture long-range dependencies and aggregates discriminative features in channel dimensions, explicitly enhances the perception of scene structure, obtains the better scene understanding and rich feature representation. 2) The detail emphasis module re-calibrates channel-wise feature maps and selectively emphasizes the informative features, aiming to highlight crucial local details information and fuse different level features more efficiently, resulting in more precise and sharper depth prediction. Furthermore, the extensive experiments validate the effectiveness of our method and show that our model achieves the state-of-the-art results on the KITTI benchmark and Make3D datasets.
\end{abstract}

\section{Introduction}

Accurate depth estimation from a single image is a fundamental task in computer vision. High quality depth information can provide useful cues for various fields, including robotics navigation \cite{desouza2002vision}, autonomous driving~\cite{menze2015object} and augmented reality~\cite{newcombe2011dtam}. Recently, the fully-supervised methods~\cite{eigen2015predicting, eigen2014depth, fu2018deep, guo2018learning} for monocular depth estimation have produced outstanding results, while they need large numbers of accurate ground truth which could only be sparsely collected from expensive LiDAR sensors~\cite{geiger2012we}. As an attractive alternative, self-supervised methods can alleviate this limitation, as they use geometrical constraints on monocular video~\cite{zhou2017unsupervised} or synchronized stereo image pairs~\cite{godard2017unsupervised} as the sole source of supervision.

In depth estimation, the most important information is the scene structure aiming at accurately obtaining the overall structure and relative depth information in 3D space. Most previous works~\cite{godard2017unsupervised, zhou2017unsupervised} just simply use convolutional neural networks to extract semantic features of input images and implicitly learn the structural information of the scene. However, the lack of explicit exploration of the robust representation of 3D scene geometry leads to an incomplete perception of overall layout for the complex scenes.

\begin{figure}
	\centering
	\resizebox{\linewidth}{!}{
		\newcommand{\imlabel}[2]{\includegraphics[width=\linewidth]{#1}%
\raisebox{58pt}{\makebox[-3pt][r]{\Large #2}}}

\begin{tabular}{@{\hskip -1.5mm}c@{\hskip 2mm}c@{\hskip 0mm}}

\centering

\imlabel{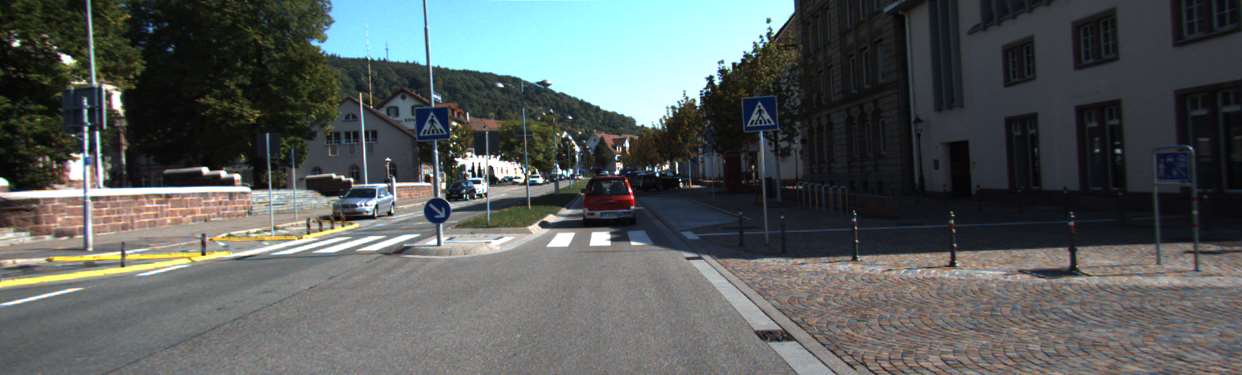} {} &
\imlabel{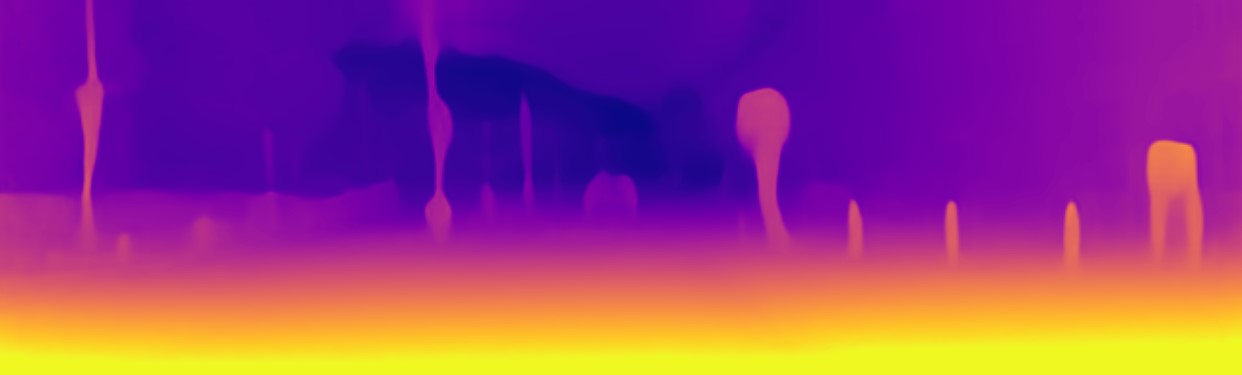} {Monodepth2~\cite{godard2019digging}} \\

\imlabel{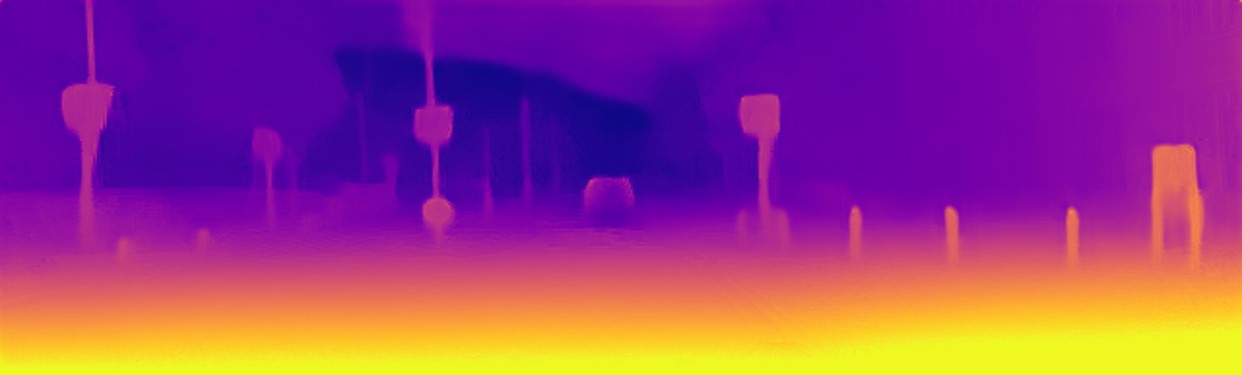} {PackNet~\cite{guizilini20203d}} &
\imlabel{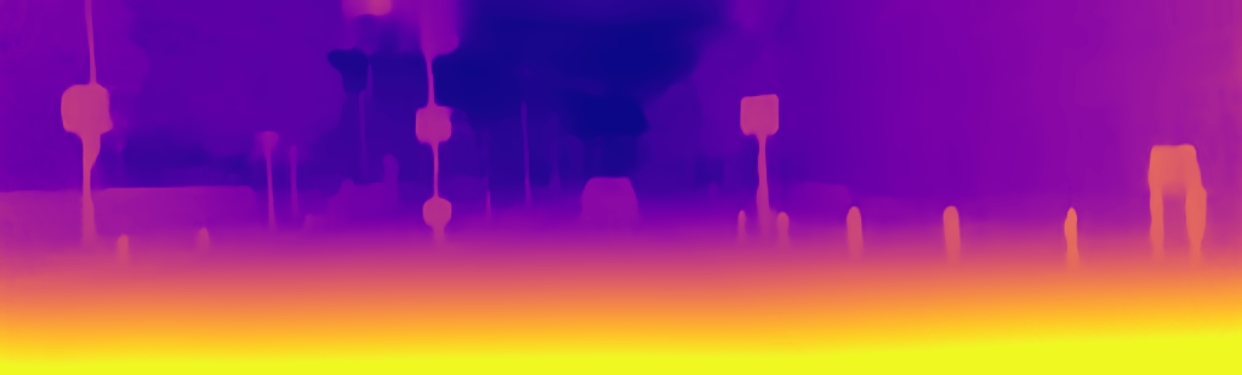} {\textbf{Ours}} \\

\imlabel{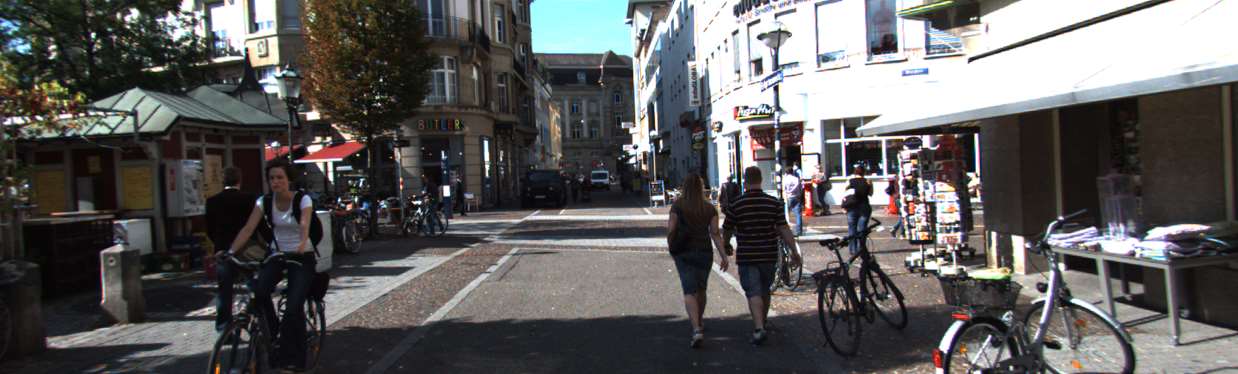}{} &
\imlabel{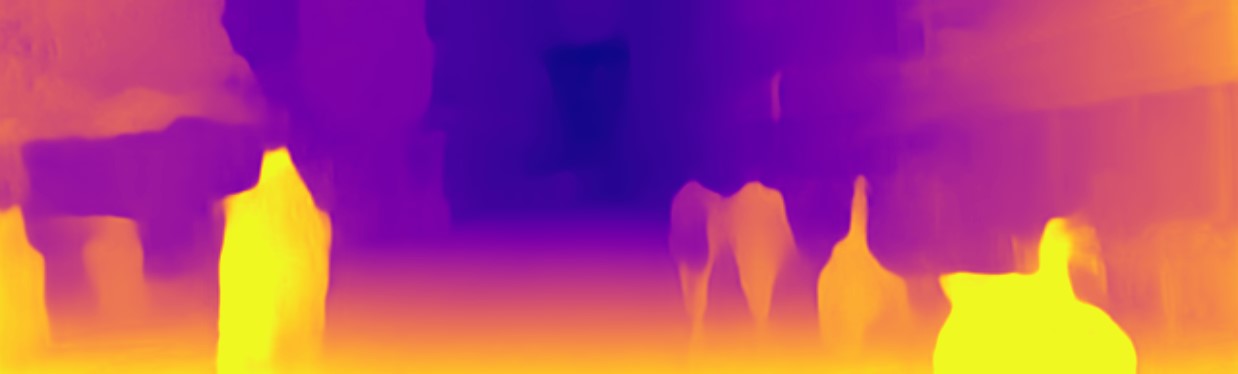} {Monodepth2~\cite{godard2019digging}} \\

\imlabel{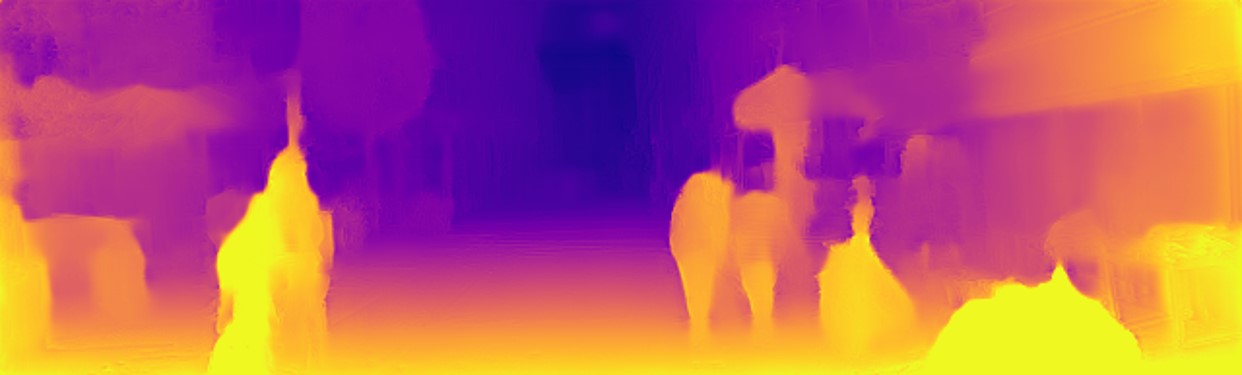} {PackNet~\cite{guizilini20203d}} &
\imlabel{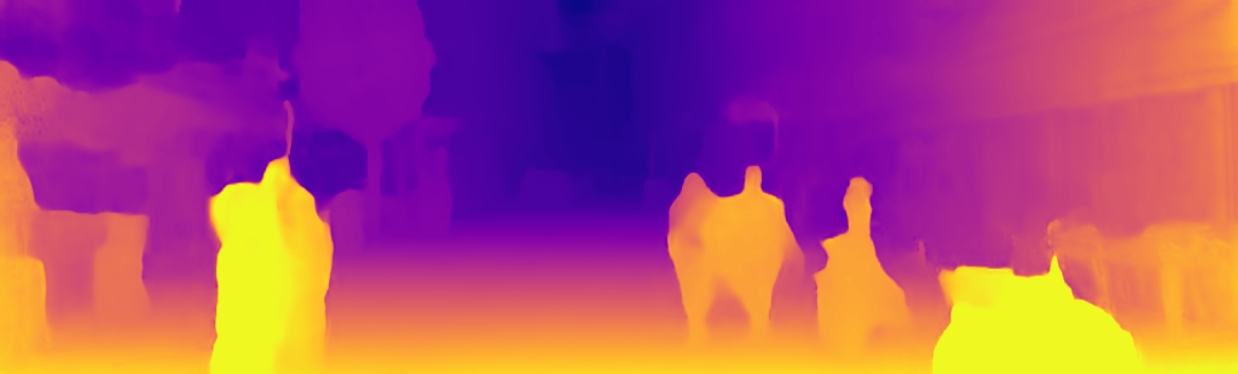} {\textbf{Ours}} \\

\end{tabular}
 }
	\caption{\textbf{Depth prediction from a single image.} Our proposed CADepth-Net produces more precise and sharper depth estimation, especially for thin structures \eg road signs and pedestrians. }
	\label{fig:teaser}
\end{figure}

Local detail is another critical feature that focuses on object boundaries and attempts to generate sharp depth maps. Most depth estimation networks are based on the U-Net~\cite{ronneberger2015u} framework, and the decoder simply leverages concatenation and a basic convolution to fuse high-level and low-level features. We found that these operations can not preserve sufficient details or precisely recover spatial information, leading to inefficient integration of different levels features and blurry artefacts at the depth discontinuous regions.

To address the above problems and efficiently handle both the overall structure and local details, we propose a novel Channel-wise Attention-based Depth Estimation Network with \textit{structure perception module} and \textit{detail emphasis module}. To better understand the 3D structure of the whole scene, we perform an in-depth analysis of semantic information in the monocular depth estimation task and conclude that each high-level feature map can be regarded as a region-specific response. Based on this observation, we provide the structure perception module to capture more contextual information of scene geometry and enhance the feature representations. Specifically, we first employ the self-attention mechanism to capture the long-range dependencies between any two channel maps, then each feature map is updated via aggregating features from all channel maps by weighted summation and fuse different local depth responses from non-contiguous regions. To generate sharper object boundaries, instead of using the above naive operations, we propose the detail emphasis module employing channel attention mechanism to re-calibrate channel-wise features and emphasize the specific semantic information. Specifically, we sequentially adopt the detail emphasis module at different scales in the decoding stage to highlight features containing crucial local details (\eg object boundaries information). To summarize our contributions in this paper:
\begin{itemize}
	\item We introduce a novel Channel-wise Attention-based Depth Network (CADepth-Net) for self-supervised monocular depth estimation employing two channel-wise attention modules to perform the information aggregation and feature re-calibration respectively.
	
	\item We propose structure perception module utilizing self-attention mechanism to obtain rich context of scene structure and better feature representation.

	\item We carefully design the detail emphasis module with channel attention mechanism to efficiently fuse different scale features and emphasize important details for sharper depth estimation.

	\item We conduct extensive experiments on KITTI and Make3D datasets, demonstrating our model significantly outperforms existing methods and achieve the state-of-the-art results on KITTI benchmark.
\end{itemize}

\section{Related Work}

\subsection{Supervised Depth Estimation}
Estimating depth from a single image is an inherently ill-posed problem as pixels in the image can have multiple plausible depths. Recently, fully supervised methods had shown the capacity of fitting predictive models to estimate depth from color input images correctly. Eigen \etal~\cite{eigen2014depth} produced dense pixel depth estimates by utilizing the multi-scale neural networks, one that estimated a coarse global depth prediction and another locally refined prediction produced by the first network. Rapidly, various fully supervised methods based on deep learning had been continuously explored~\cite{fu2018deep, eigen2015predicting, laina2016deeper, liu2015deep}. However, all the above methods required high-quality ground truth depth, which can be costly to obtain.

\subsection{Self-supervised Monocular Depth Estimation}

To overcome the limitation of supervised approaches, self-supervised methods unified depth estimation and ego-motion estimation into one framework with view synthesis as supervision signal. SfMLearner introduced by Zhou \etal~\cite{zhou2017unsupervised} simultaneously learned depth and ego-motion from monocular video by training a depth estimation network along with a separate pose network. Furthermore, Yin \etal~\cite{yin2018geonet} decomposed scene motion into rigid and non-rigid parts to account for object motion. Wang \etal~\cite{wang2018learning} incorporated Direct Visual Odometry to estimate the relative camera pose. \cite{mahjourian2018unsupervised} proposed 3D constraints loss to enforce consistency of the estimated depth and ego-motion across consecutive frames.
Guizilini \etal~\cite{guizilini20203d} learned to compress and decompress detail-preserving representations by symmetrical packing and unpacking blocks. Other published methods were based upon edge and normal~\cite{yang2018lego, yang2017unsupervised}, Competitive Collaboration~\cite{ranjan2019competitive}, semantic segmentation~\cite{guizilini2020semantically, klingner2020self} and feature representations learning~\cite{spencer2020defeat}. A state-of-the-art framework was Monodepth2 proposed by Godard \etal~\cite{godard2019digging}, which introduced a minimum re-projection loss to deal with occlusions and auto-masking scheme removing invalid pixels robustly. Our model is based on Monodepth2 extended with our contributions. 

\subsection{Self-attention Mechanism}

Self-attention mechanism had been widely used to capture long-range dependencies in various tasks. The Transformer~\cite{vaswani2017attention} was the first work that proposed the self-attention mechanism to handle long-range dependencies between words in machine translation. Wang \etal~\cite{wang2018non} modeled the spatial-temporal dependencies in video sequences and images via aggregating query-specific global context to each query position. Zhang \etal~\cite{zhang2019self} incorporated the self-attention mechanism into the GAN framework and learned a better image generator. Fu \etal~\cite{fu2019dual} enhanced the ability of feature representations for scene segmentation by designing two types of attention modules. For the monocular depth estimation task, Johnston \etal~\cite{johnston2020self} captured the context of similar disparity values at non-contiguous regions by exploring the feature similarity at spatial dimensions. Unlike previous works, we demonstrate that capturing global dependencies along the channel dimensions and aggregating discriminative features will achieve better performance for depth estimation, as each channel map gains more relative depth information from the distant regions.  

\begin{figure*}[t]
	\centering
	\includegraphics[width=\linewidth]{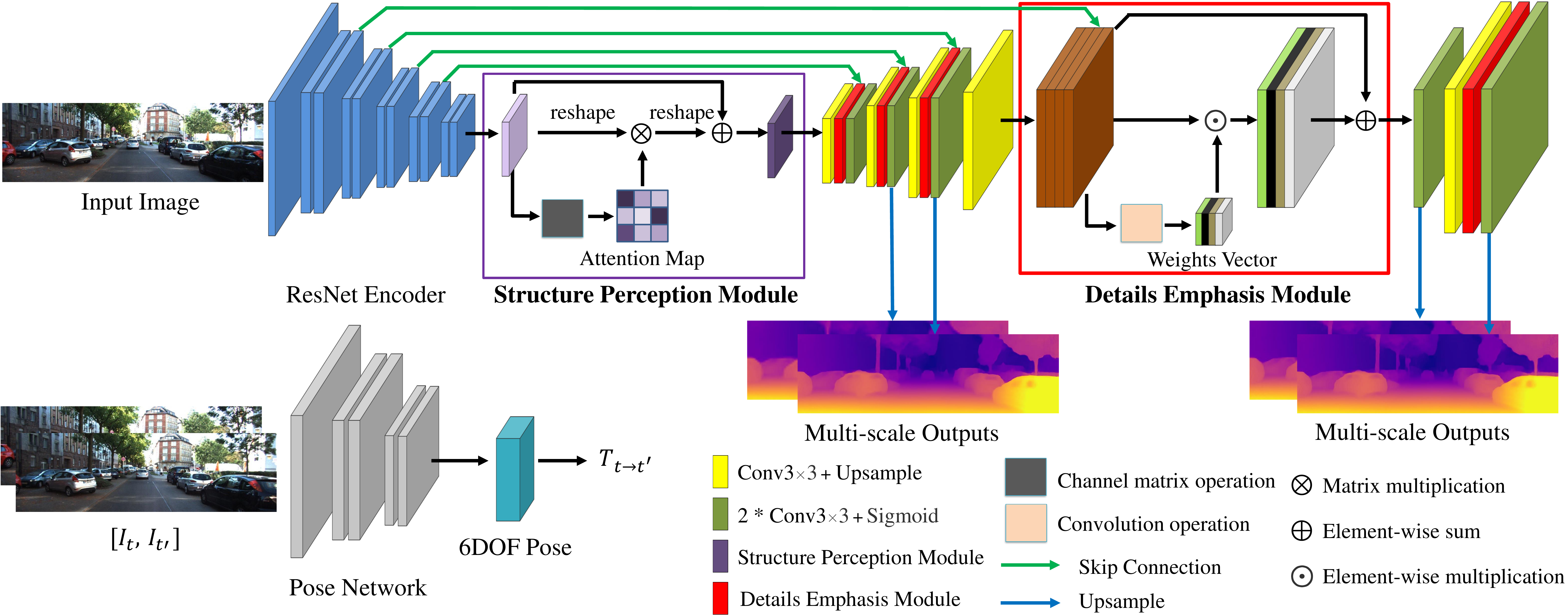}
	\caption{\textbf{Overview of Framework.} Our proposed CADepth-Net is a fully convolutional U-Net architecture. We first use a ResNet encoder to extract semantic features and input them to the \textbf{Structure Perception Module}. We perform the matrix multiplication between input features and attention maps to generate aggregated features. The low-resolution feature maps are passed through successive blocks of UpConv(upsample + convolution), as well as the \textbf{Detail Emphasis Module} which computes a weight vector to re-calibrate channel-wise features. Finally, we upsample the predicted disparities at multiple scales to original input resolutions. Besides, the pose network takes temporally adjacent images $I_t, I_{t^{\prime}}$ as input and outputs relative pose $T_{t \rightarrow t^{\prime}}$.}
	\label{fig:model_arch}
\end{figure*}

\section{Method}
In this section, we firstly review the training methods for self-supervised monocular depth estimation, then introduce the architecture of our Channel-wise Attention-based Network, finally describe our main contributions, the structure perception module and detail emphasis module.

\subsection{Self-Supervised Training}

The goal of self-supervised monocular depth estimation is to predict the depth map from a single RGB image without ground truth. Specifically, given a single input image $I_t$, the depth network predicts its corresponding depth map $D_t$, then the pose network takes temporally adjacent images as input and predicts relative pose $T_{t \rightarrow t'}$ between the target image $I_t$ and source images $I_{t'}$, $t^{\prime} \in\{t-1, t+1\}$, finally we use the predicted $D_t$ and $T_{t \rightarrow t'}$ to perform view synthesis as the supervisory signal.

At training time, both the depth network and pose network are optimized jointly by minimizing the per-pixel minimum photometric re-projection error  $L_p$~\cite{godard2019digging} 
\begin{equation}
L_{p} =\min _{t^{\prime}} pe\left(I_{t}, I_{t^{\prime} \rightarrow t}\right),
\label{eq: photometric loss}
\end{equation}
where $pe()$ denotes the photometric error which consists of L1 and the Structural Similarity (SSIM)~\cite{wang2004image}, $I_{t^{\prime} \rightarrow t}$ is the warped result from $I_{t'}$ to  $I_t$ as in
\begin{equation}
\begin{split}
pe = &\frac{\alpha}{2}\left(1-\operatorname{SSIM}\left(I_{t}, I_{t^{\prime} \rightarrow t}\right)\right) \\
&+(1-\alpha)\left\|I_{t}-I_{t^{\prime} \rightarrow t}\right\|_{1},
\end{split}
\label{eq: photometric rec loss}
\end{equation}
\begin{equation}
I_{t^{\prime} \rightarrow t}=I_{t^{\prime}}\left\langle\operatorname{proj}\left(D_{t}, T_{t \rightarrow t^{\prime}}, K\right)\right\rangle,
\label{eq: proj}
\end{equation}
where $proj()$ represents the resulting 2D coordinates of the projected depths $D_t$ in $I_{t'}$ and $\langle\rangle$ is the sampling operator. We use the differentiable bilinear sampling mechanism proposed in the STN~\cite{jaderberg2015spatial} to sample the source images. 

In a real-world scenario, situations like stationary camera and moving objects will break down the assumptions of a moving camera and a static scene. To handle this issue, we apply auto-masking method~\cite{godard2019digging} to filter out stationary pixels that remain with the same appearance between two frames in a sequence. Since the binary mask $\mu$ is computed in this form on the forward pass
\begin{equation}
\mu=\left[\min _{t^{\prime}} p e\left(I_{t}, I_{t^{\prime} \rightarrow t}\right)<\min _{t^{\prime}} p e\left(I_{t}, I_{t^{\prime}}\right)\right],
\label{eq: auto-masking}
\end{equation}
where $[ ]$ is the Iverson bracket.

In addition, in order to regularize the disparities in texture-less regions, an edges-aware smoothness regularization term $L_{s}$ is used 
\begin{equation}
L_{s}=\left|\partial_{x} d_{t}^{*}\right| e^{-\left|\partial_{x} I_{t}\right|}+\left|\partial_{y} d_{t}^{*}\right| e^{-\left|\partial_{y} I_{t}\right|},
\label{eq: smoothness loss}
\end{equation}
where $d_{t}^{*}=d_{t} / \overline{d_{t}}$ is the mean-normalized inverse depth from~\cite{wang2018learning} to discourage shrinking of the estimated depth. 

The final loss $L$ is computed as the combination of photometric loss $L_{p}$ and smoothness loss $L_{s}$ at multiple scales
\begin{equation}
L=\frac{1}{S} \sum_{i}^{S}\left(\mu L_{p}^{i}+\lambda L_{s}^{i}\right),
\label{eq: total loss}
\end{equation}
where $S$ is the number of scales, and $\lambda$ is the weighting for the smoothness regularization term.    

\begin{figure}
	\centering
	\includegraphics[width=\linewidth]{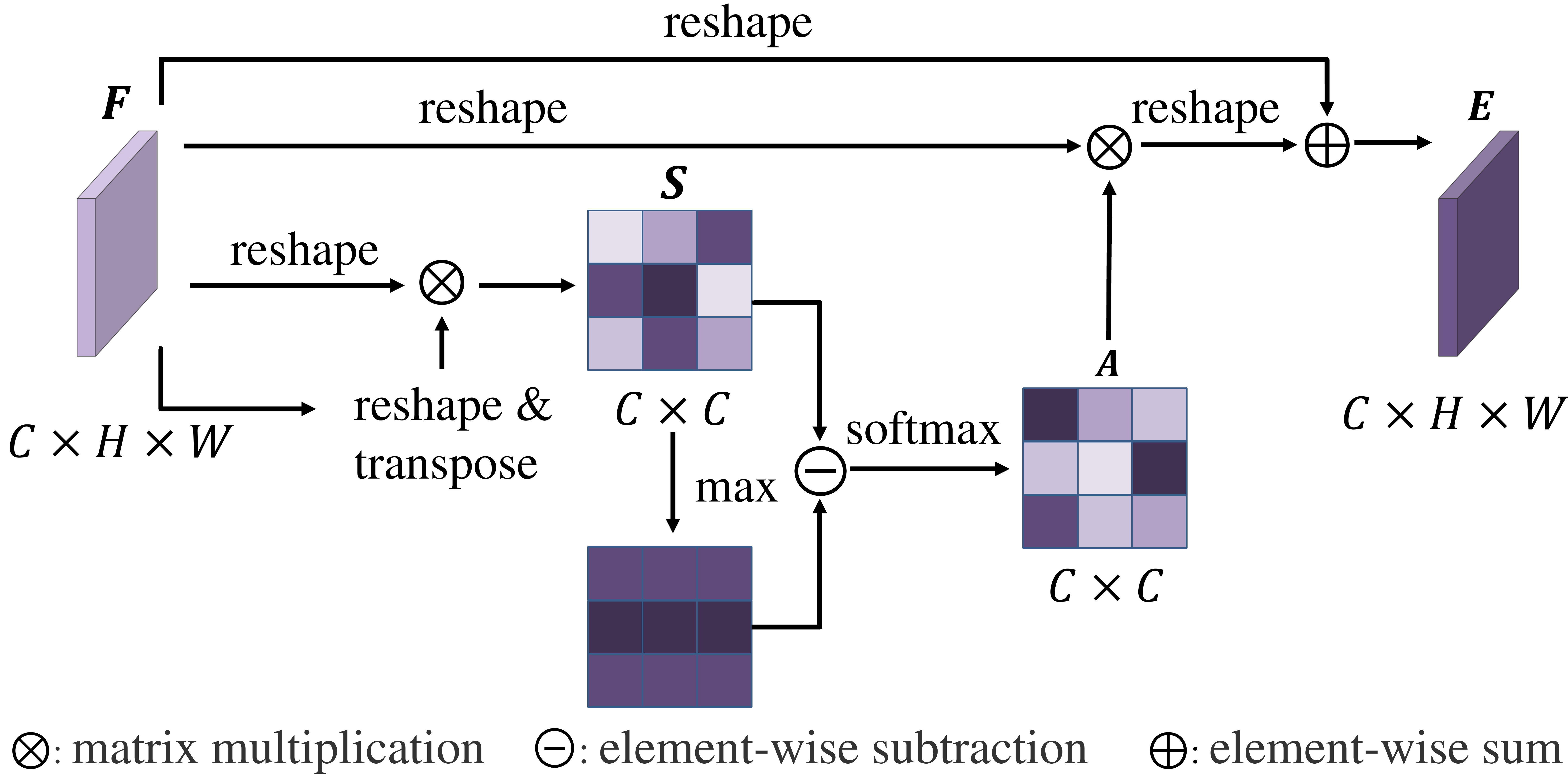}
	\caption{\textbf{Details of the structure perception module.}
	}
	\label{fig:SPM_arch}
\end{figure}

\subsection{Channel-wise Attention-based Network}

As shown in Fig. \ref{fig:model_arch}, our CADepth-Net is a fully convolutional U-Net architecture. We adopt a pretrained residual network as the backbone to extract semantic features. Then these features would be fed into the structure perception module and generate new features to explicitly enhance the perception of scene structure. Moreover, we gradually recover the spatial resolution at decoder stage, with skip-connection to facilitate the flow of gradients and information throughout the model, and sequentially employ our detail emphasis module to generate sharp edges and finer details. Finally, we successively upsample the predicted inverse depth maps until original input resolutions using nearest neighbors interpolation at multiple scales, and compute the training loss at this higher input resolution.

\subsection{Structure Perception Module}

In depth estimation, each high-level feature map can be regarded as a region-specific response as shown in Fig.~\ref{fig:SPM_vis} (b), and different region responses are associated with each other. If each channel map captures more different region responses from all the other channel maps, as shown in Fig.~\ref{fig:SPM_vis} (c), it will obtain more relative depth information from distant regions and significantly enhance the perception of scene structure. Therefore, we propose a self-attention module to model interdependencies between channels and aggregate different region responses.

The first step is to generate an attention matrix which models the relationship between any two channel maps. As illustrated in Fig.~\ref{fig:SPM_arch}, given the feature map $\mathbf {F} \in \mathbb{R}^{C \times H \times W}$ produced by ResNet encoder, we firstly reshape $\mathbf {F}$ to $\mathbb{R}^{C\times N}$, where $N=H\times W$ is the number of pixels, then perform a matrix multiplication between $\mathbf {F}$ and the transpose of $\mathbf {F}$ to compute the feature similarity $\mathbf {S} \in \mathbb{R}^{C\times C}$
\begin{equation}
S_{ij} = F_{i} \cdot F^{T}_{j}.
\label{eq: similarity}
\end{equation}

The similarity between channel maps indicates the spatial relationship of region responses \ie~any two feature maps have higher similarity means that they also have strong responses to the same region. As we need to fuse more responses from \textit{different regions}, we convert the similarity $\mathbf {S} $ to discrimination $\mathbf {D} \in \mathbb{R}^{C\times C}$ by performing the element-wise subtraction
\begin{equation}
D_{ij}=max_{i}(S) - S_{i,j},
\label{eq: discrimination}
\end{equation}
where $D_{ij}$ measures the $j^{th}$ channel's impact on the $i^{th}$ channel. For each channel map, other channels with discriminative features (\ie~different region response) will get higher scores $D_{ij}$ during feature aggregation. Then we apply a softmax layer to obtain the attention map $\mathbf {A} \in \mathbb{R}^{C\times C}$
\begin{equation}
A_{ij}=\frac{\exp \left(D_{ij} \right)}{\sum_{j=1}^{C} \exp \left(D_{ij}\right)}.
\label{eq: attention map}
\end{equation}

In addition, we perform a matrix multiplication between the transpose of $\mathbf {A}$ and $\mathbf {F}$ and reshape the result to $\mathbb{R}^{C\times H \times W}$. Finally we perform an element-wise sum operation between $\mathbf {F}$ and the result to obtain the final output $\mathbf {E} \in \mathbb{R}^{C\times H \times W}$ as follows 
\begin{equation}
E_{i}=\sum_{j=1}^{C}\left(A_{ij} F_{j}\right)+F_{i}.
\label{eq: final output}
\end{equation}

The Eq. \ref{eq: final output} shows that the final feature of each channel is the weighted sum of the features from all channels and the original feature. By capturing the long-range dependencies between feature maps, we obtain the aggregated features encoding rich context information of scene structure.

\begin{figure}
	\centering
	\includegraphics[width=\linewidth]{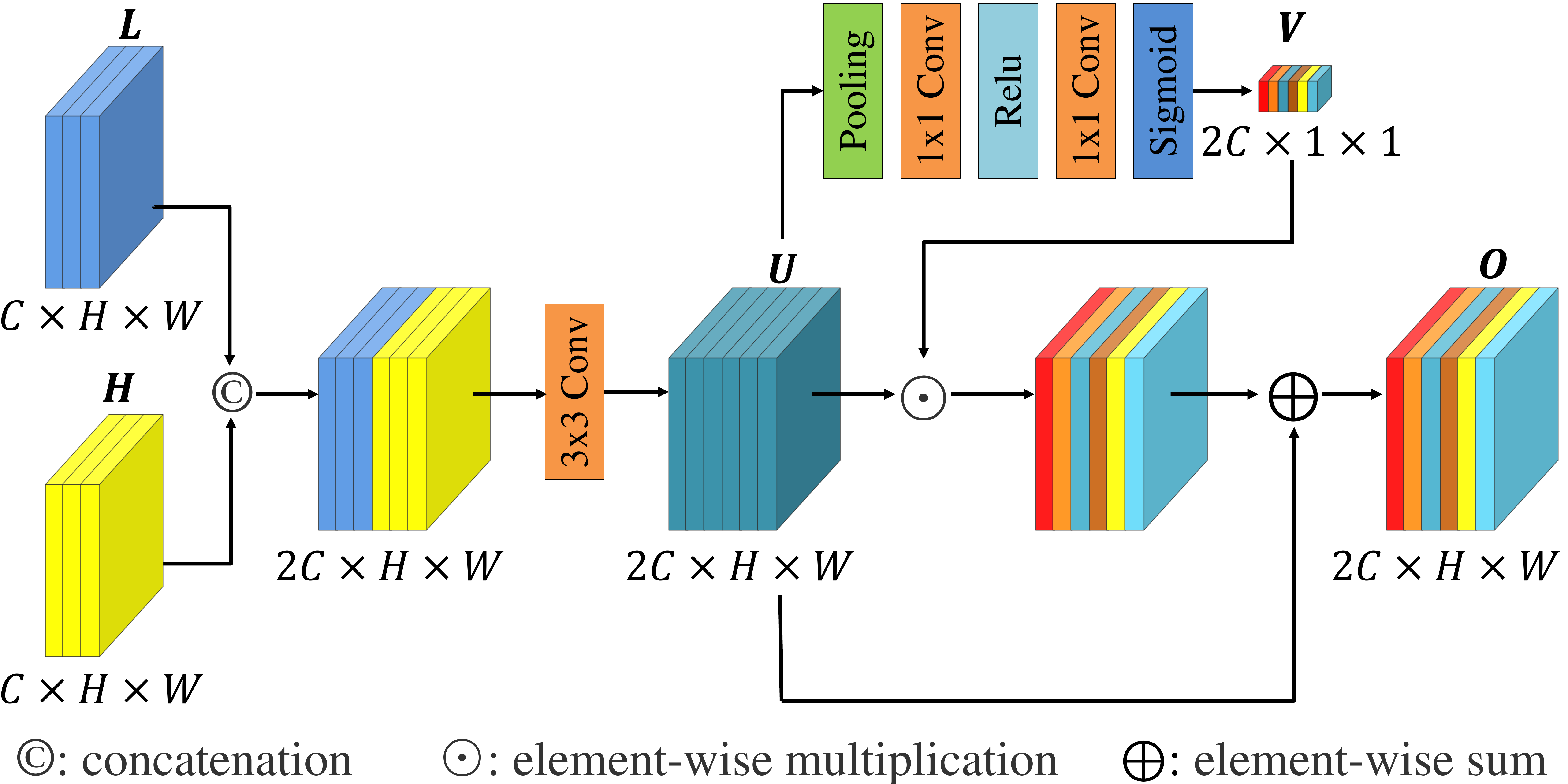}
	\caption{\textbf{Design of the detail emphasis module.} }
	\label{fig:DEM_arch}
\end{figure} 

\subsection{Detail Emphasis Module}
The decoder recovers the resolution by fusing the following features at different scales: the low-level information from skip-connection encoding rich spatial details, and the high-level information encoding more context information. The simple fusion operations like sum or concatenation lack the further processing of local details and neglect the semantic gap between different level features, leading to blurry artefacts in predicted depth maps. The core of predicting sharper edges is to properly handle local details, and it is easy for network to recover accurate depth predictions if it knows the category and location of features describing object boundaries clearly. Therefore, by using the channel attention mechanism that can make network pay attention to specific channel features, we propose a detail emphasis module to emphasize important details and efficiently fuse features at different scales.

Specifically, we first concatenate the low-level features $\mathbf{L}$ and the high-level features $\mathbf{H}$, then utilize a convolution layer to obtain $\mathbf {U}$ with the batch normalization~\cite{ioffe2015batch} to balance the scales of the features 
\begin{equation}
U = \sigma \left(BN\left(W_1 \otimes f\left(L, H\right)\right)\right),
\label{eq: U}
\end{equation}
where $ f() $ denotes concatenation and $ \otimes $ denotes the $ 3 \times 3 $ or $ 1 \times 1 $ convolution, BN refers to the batch normalization and we use the ReLU as the activation function $ \sigma() $.

Next, we squeeze $\mathbf {U}$ to a vector by global average pooling to obtain global context and use two $1 \times 1$ convolution layers followed by a sigmoid function to compute a weights vector $\mathbf {V} \in \mathbb{R}^{1 \times 1 \times C}$, to recalibrate channel-wise features and measure the importance of them in the meantime
\begin{equation}
V = \delta (W_2 \otimes \sigma (W_3 \otimes (\frac{1}{H \times W} \sum_{i=1}^{H} \sum_{j=1}^{W} U_{i,j}))),
\label{eq: V}
\end{equation}
where $H$ and $W$ refer to the height and width of $\mathbf {U}$, and $\delta()$ denotes the sigmoid function. Then we perform the element-wise multiplication between $\mathbf {V}$ and $\mathbf {U}$ to generate re-weight features. As the weights scores in $\mathbf {V}$ indicate the importance of corresponding channels, \ie~the channel maps containing critical information will get higher scores, this recalibration operation can adaptively emphasize the crucial details at multiple scales for sharp edges (Fig.~\ref{fig:DEM_vis}). Finally, we sum up $\mathbf {U}$ and re-weight features for stability
\begin{equation}
O = V \odot U + U,
\label{eq: O}
\end{equation}
where $\odot$ denotes element-wise dot product and $\mathbf {O}$ refers to the final outputs. Fig.~\ref{fig:DEM_arch} shows the design of the detail emphasis module, this design recalibrates channel-wise feature responses and produce more precise depth estimation.

\begin{table*}[t!]
	\centering
	\resizebox{\linewidth}{!}{
		\footnotesize
		\begin{tabular}{|l|c|c||c|c|c|c|c|c|c|}
			\hline
			Method & Train & Resolution & \cellcolor{col1}Abs Rel & \cellcolor{col1}Sq Rel & \cellcolor{col1}RMSE  & \cellcolor{col1}RMSE log & \cellcolor{col2}$\delta < 1.25 $ & \cellcolor{col2}$\delta < 1.25^{2}$ & \cellcolor{col2}$\delta < 1.25^{3}$\\
			\hline
			
			SfMLeaner ~\cite{zhou2017unsupervised}\textdagger & M & 416 $\times $ 128 & 0.183 & 1.595 & 6.709 & 0.270 & 0.734 & 0.902 & 0.959\\
			
			
			
			
			GeoNet~\cite{yin2018geonet}\textdagger & M & 416 $\times $ 128 & 0.149 & 1.060 & 5.567 & 0.226 & 0.796 & 0.935 & 0.975 \\
			
			DDVO~\cite{wang2018learning} & M & 416 $\times $ 128  & 0.151 & 1.257 & 5.583 & 0.228 & 0.810 & 0.936 & 0.974 \\
			
			Struct2depth `(M)'~\cite{casser2019depth} & M & 416 $\times $ 128 & 0.141 & 1.026 & 5.291 &  0.215 & 0.816 & 0.945 & \underline{0.979} \\
			
			
			SIGNet~\cite{meng2019signet} & M & 416 $\times $ 128 & 0.133 & \underline{0.905} & 5.181 & 0.208 & 0.825 & 0.947 & \bf{0.981} \\
			
			SGDepth~\cite{klingner2020self} & M & 416 $\times $ 128 & \underline{0.128} & 1.003 & \underline{5.085} & 0.206 & 0.853 & 0.951 & 0.978 \\ 
			
			Monodepth2~\cite{godard2019digging}& M & 416 $\times $ 128 & \underline{0.128} & 1.087 & 5.171 & \underline{0.204} & \underline{0.855} & \underline{0.953} & 0.978 \\ 
			
			\rowcolor{gray!30} \textbf{CADepth-Net (Ours)} & M & 416 $\times$ 128 & \bf{0.116} & \bf{0.893} & \bf{4.906} & \bf{0.192} & \bf{0.874} & \bf{0.957} & \bf{0.981} \\  
			
			\hline
			
			DF-Net~\cite{zou2018df} & M & 576 $\times $ 160 & 0.150 & 1.124 & 5.507 & 0.223 & 0.806 & 0.933 & 0.973 \\
			
			SGDepth~\cite{klingner2020self} & M & 640 $\times$ 192 & 0.117 & 0.907 & 4.844 & 0.196 & 0.875 & 0.958 &  0.980 \\ 
			
			Monodepth2~\cite{godard2019digging}& M & 640 $\times$ 192 & 0.115 & 0.903 & 4.863 & 0.193 & 0.877 &  0.959 & 0.981 \\ 
			
			PackNet-SfM~\cite{guizilini20203d} & M & 640 $\times$ 192 & 0.111 & \underline{0.785} & \underline{4.601} & 0.189 & 0.878 & 0.960 & \underline{0.982} \\   
			
			
			HR-Depth~\cite{lyu2020hr} & M & 640 $\times$ 192 & 0.109 & 0.792 & 4.632 & \underline{0.185} & 0.884 & \underline{0.962} & \bf{0.983} \\ 
			
			Johnston \etal~\cite{johnston2020self} & M & 640 $\times$ 192 & \underline{0.106} & 0.861 & 4.699 & \underline{0.185} & \underline{0.889} & \underline{0.962} & \underline{0.982} \\ 
			
			\rowcolor{gray!30} \textbf{CADepth-Net (Ours)} & M & 640 $\times$ 192 & \bf{0.105} & \bf{0.769} & \bf{4.535} & \bf{0.181} & \bf{0.892} & \bf{0.964} & \bf{0.983} \\  			
			
			\hline
			
			
			
			CC~\cite{ranjan2019competitive}  & M  & 832 $\times $ 256 & 0.140 & 1.070 & 5.326 & 0.217 & 0.826 & 0.941 & 0.975 \\
			
			
			Monodepth2~\cite{godard2019digging}& M & 1024 $\times$ 320 & 0.115 & 0.882 & 4.701 & 0.190 & 0.879 & 0.961 & \underline{0.982} \\ 
			
			TrianFlow~\cite{zhao2020towards} & M & 832 $\times$ 256 & 0.113 & 0.704 & 4.581 & 0.184 & 0.871 & 0.961 & \bf{0.984} \\   
			
			HR-Depth~\cite{lyu2020hr} & M & 1024 $\times$ 320 & 0.106 & 0.755 & \underline{4.472} & 0.181 & 0.892 & \bf{0.966} & \bf{0.984} \\ 
			
			FeatDepth~\cite{shu2020feature} & M & 1024 $\times$ 320 & \underline{0.104} & \bf{0.729} & 4.481 & \underline{0.179} & \underline{0.893} & \underline{0.965} & \bf{0.984} \\ 
			
			\rowcolor{gray!30} \textbf{CADepth-Net (Ours)} & M & 1024 $\times$ 320 & \bf{0.102} & \underline{0.734} & \bf{4.407} & \bf{0.178} & \bf{0.898} & \bf{0.966} & \bf{0.984} \\  
			
			\hline
			
			DualNet~\cite{zhou2019unsupervised} & M & 1248 $\times$ 384 & 0.121 & 0.837 & 4.945 & 0.197 & 0.853 & 0.955 & \underline{0.982} \\ 
			
			SGDepth~\cite{klingner2020self} & M & 1280 $\times$ 384 & 0.113 & 0.880 & 4.695 & 0.192 & 0.884 & 0.961 & 0.981 \\ 
			
			PackNet-SfM~\cite{guizilini20203d} & M & 1280 $\times$ 384 & 0.107 & 0.802 & 4.538 & 0.186 & 0.889 & 0.962 & 0.981 \\  
			
			HR-Depth~\cite{lyu2020hr} & M & 1280 $\times$ 384 & \underline{0.104} & \underline{0.727} & \underline{4.410} & \underline{0.179} &    \underline{0.894} & \underline{0.966} & \bf{0.984} \\
			
			\rowcolor{gray!30} \textbf{CADepth-Net (Ours)} & M & 1280 $\times$ 384 & \bf{0.102} & \bf{0.715} & \bf{4.312} & \bf{0.176} & \bf{0.900} &  \bf{0.968} & \bf {0.984} \\
			
			\hline
			
			
			
			EPC++~\cite{luo2019every} & MS & 832 $\times$ 256 & 0.128 & 0.935 & 5.011 & 0.209 & 0.831 & 0.945 & 0.979 \\
			
			Monodepth2~\cite{godard2019digging} & MS & 640 $\times$ 192 & 0.106 & 0.818 & 4.750 & 0.196 & 0.874 & 0.957 & 0.979 \\
			
			HR-Depth~\cite{lyu2020hr} & MS & 640 $\times$ 192 & 0.107 & 0.785 & \underline{4.612} & \underline{0.185} & \underline{0.887} & \underline{0.962} & \underline{0.982} \\
			
			DepthHints~\cite{watson2019self} & MS & 640 $\times$ 192 & \underline{0.105} & \underline{0.769} & 4.627 & 0.189 & 0.875 & 0.959 & \underline{0.982} \\   
			
			\rowcolor{gray!30} \textbf{CADepth-Net (Ours)} & MS & 640 $\times$ 192 & \bf{0.102} & \bf{0.752} & \bf{4.504} & \bf{0.181} & \bf{0.894} & \bf{0.964} & \bf{0.983} \\ 			
			
			\hline
			
			Monodepth2~\cite{godard2019digging} & MS & 1024 $\times$ 320 & 0.106 & 0.806 & 4.630 & 0.193 & 0.876 & 0.958 & 0.980 \\
			
			HR-Depth~\cite{lyu2020hr} & MS & 1024 $\times$ 320 & 0.101 & 0.716 & \underline{4.395} & \underline{0.179} & \underline{0.899} & \underline {0.966} & \underline{0.983} \\
			
			DepthHints~\cite{watson2019self} & MS & 1024 $\times$ 320 & 0.098 & 0.702 & 4.398 & 0.183 & 0.887 & 0.963 & \underline {0.983} \\ 
			
			FeatDepth~\cite{shu2020feature}  & MS & 1024 $\times$ 320 & \underline{0.099} & \underline{0.697} & 4.427 & 0.184 & 0.889 & 0.963 & 0.982 \\
			
			\rowcolor{gray!30} \textbf{CADepth-Net (Ours)} & MS & 1024 $\times$ 320 & \bf{0.096} & \bf{0.694} & \bf{4.264} & \bf{0.173} & \bf{0.908} & \bf{0.968} & \bf{0.984} \\   
			
			\arrayrulecolor{black}\hline
			
		\end{tabular}
	}
	\vspace{1pt}
	\caption{\textbf{Quantitative results on the KITTI Eigen Split.} Comparison of existing methods on KITTI 2015~\cite{geiger2012we} using the Eigen split for distances up to 80m.  All methods in this table are trained on the KITTI dataset without additional datasets or online refinement. Best results are in \textbf{bold}, with second-best \underline{underlined}. For Abs Rel, Sq Rel, RMSE and  RMSE$_{log}$ lower is better, and for $\delta < 1.25$, $\delta < 1.25^2$ and $\delta < 1.25^3$ higher is better. In the \emph{Train} column, S: Self-supervised stereo supervision, M: Self-supervised mono supervision. \textdagger~ refers to the newer results from Github. At test time, we scale the estimated depths with median ground-truth LiDAR information. }	
	\label{tab:kitti_eigen}
\end{table*}

\section{Experiments}

In this section, we show extensive experiments for evaluating the performance of our methods and demonstrate the effectiveness of the proposed approaches. 

\subsection{Implementation Details}

Our model are implemented based on PyTorch~\cite{paszke2017automatic}, trained for $20$ epochs on a single Nvidia $3090$ with a batch size of $12$ and an input/output resolution of $640 \times 192$. We jointly train both depth network and pose network with the Adam Optimizer \cite{kingma2014adam} with $\beta_1=0.9$, $\beta_2=0.999$. The initial learning rate is set to $1e^{-4}$ and decay to $1e^{-5}$ after $15$ epochs.  We set the SSIM weight to $\alpha = 0.85$ and smoothness term weight to $\lambda=1e-3$.

\noindent\textbf{DepthNet.} We implement our depth estimation network as an encoder-decoder architecture. Moreover, we start the ResNet50 encoder with weights pretrained on ImageNet \cite{russakovsky2015imagenet} as it has been shown to improve accuracy compared to training from scratch. We set sigmoid follow the output of network and convert result $\sigma$  to depth with $D=1/(a\sigma + b)$, where use $a$ and $b$ to constrain $D$ between $0.1$ and $100$ units. 

\noindent\textbf{PoseNet.} Our PoseNet is built on ResNet50, modified to accept six channels tensor as input, which allows the adjacent frames to feed into the network. The outputs of PoseNet is the 6-DoF relative pose consist of translation vectors and Euler angles, scaled by a factor of 0.01.

\begin{figure*}[!ht]
	\centering
	\resizebox{\linewidth}{!}{
		\newcommand{\turnheightnew}{0.2\columnwidth}
\centering

\begin{tabular}{@{\hskip -0.5mm}c@{\hskip 0.5mm}c@{\hskip 0.5mm}c@{\hskip 0.5mm}c@{\hskip -0.5mm}}

\normalsize{Input}  & \normalsize{Monodepth2~\cite{godard2019digging}} &  \normalsize {PackNet~\cite{guizilini20203d}}  & \textbf{\normalsize{Our CADepth-Net}} \\

\includegraphics[height=\turnheightnew]{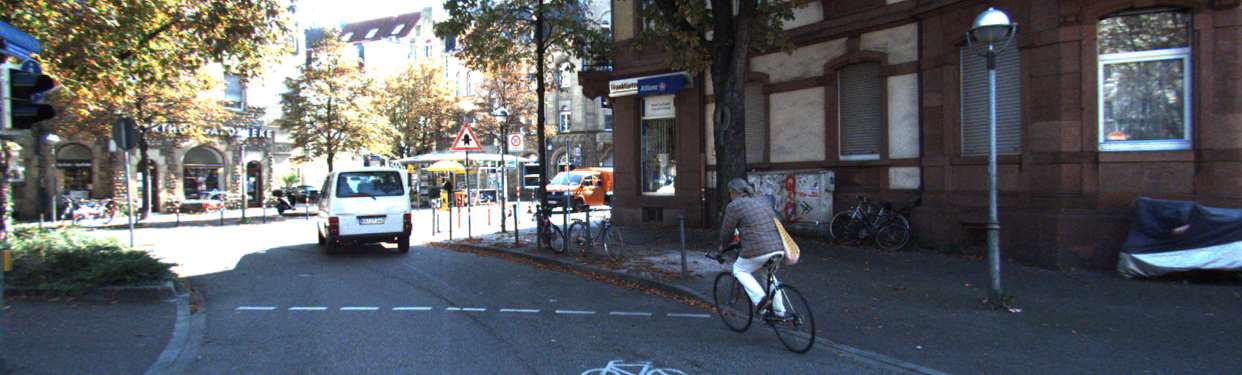} &
\includegraphics[height=\turnheightnew]{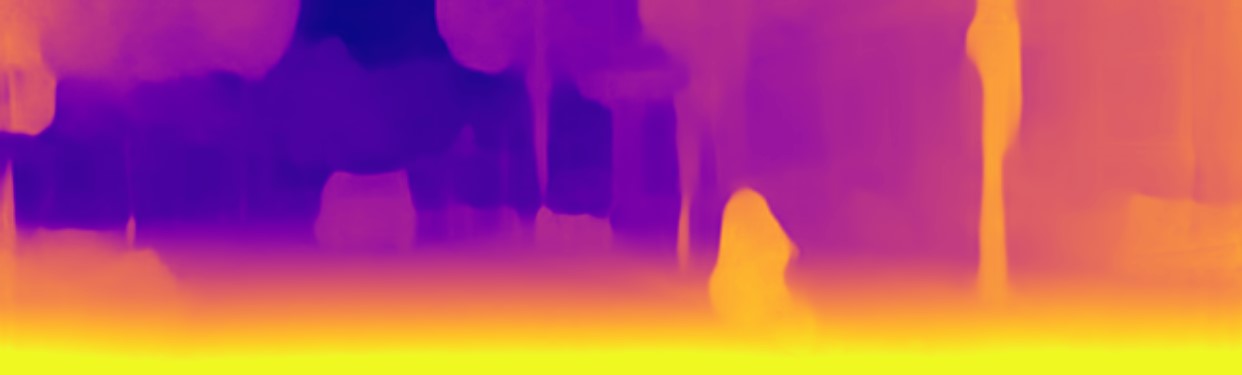} &
\includegraphics[height=\turnheightnew]{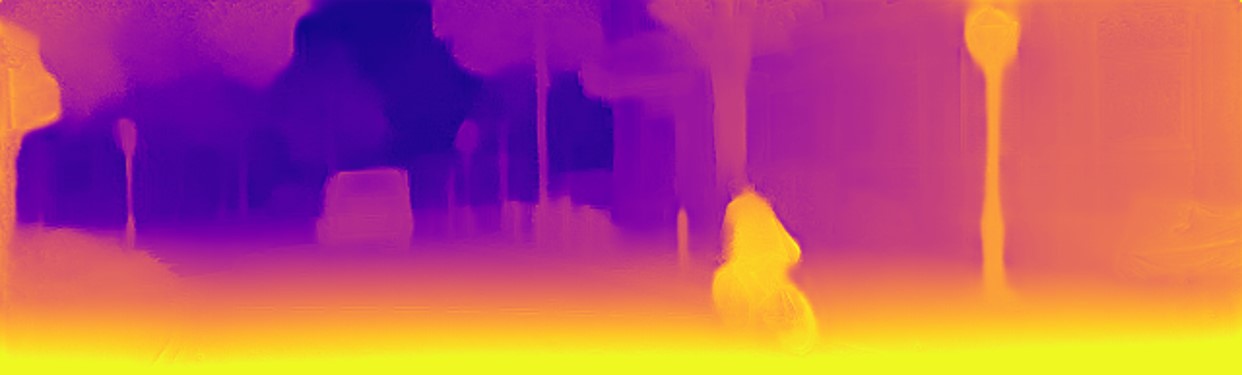} &
\includegraphics[height=\turnheightnew]{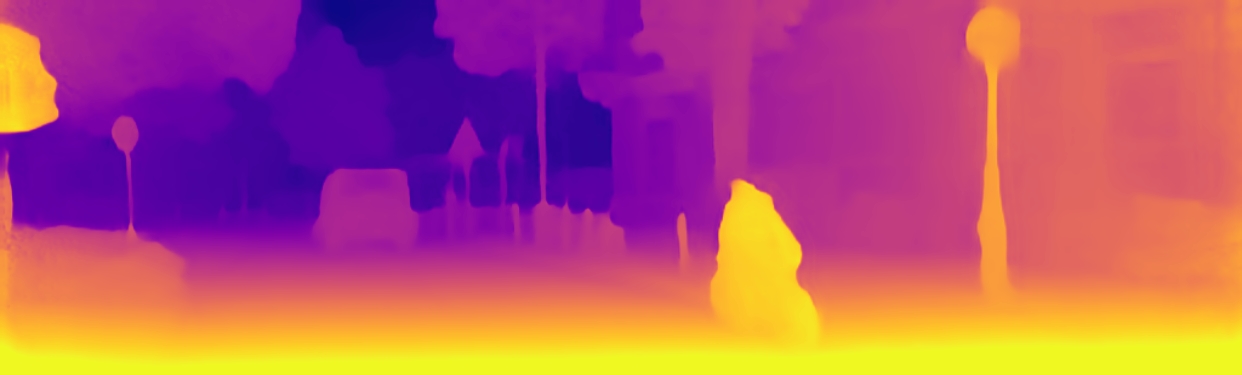}\\

\includegraphics[height=\turnheightnew]{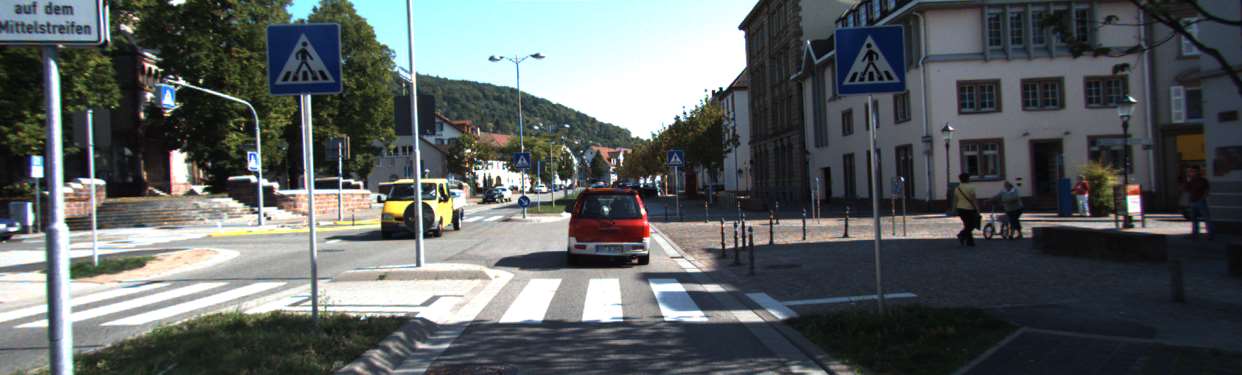} &
\includegraphics[height=\turnheightnew]{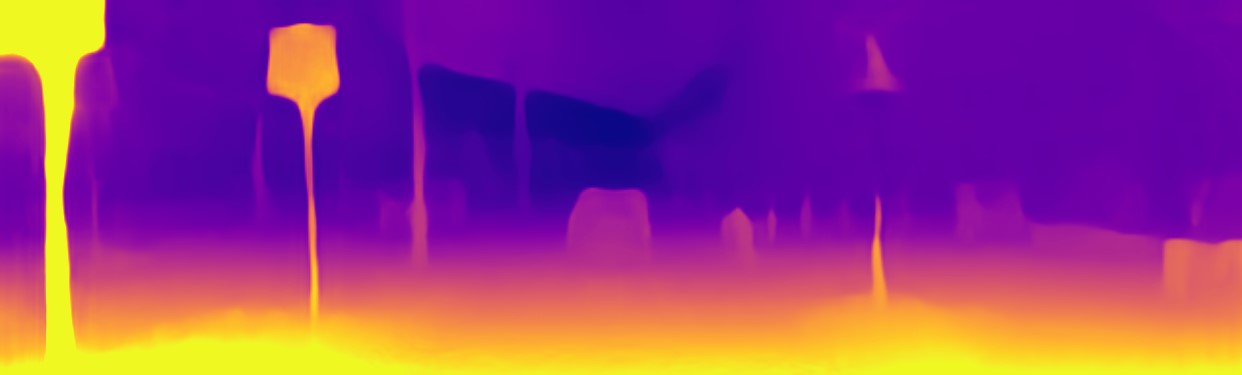} &
\includegraphics[height=\turnheightnew]{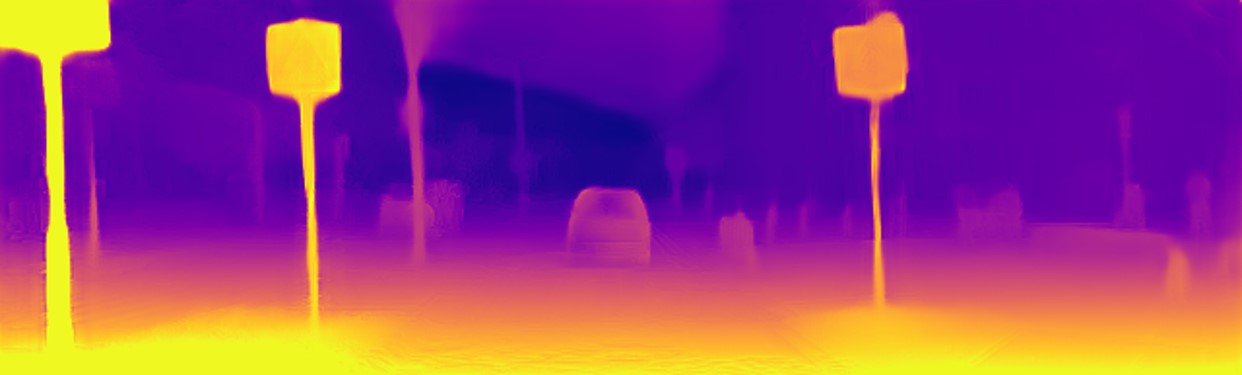} &
\includegraphics[height=\turnheightnew]{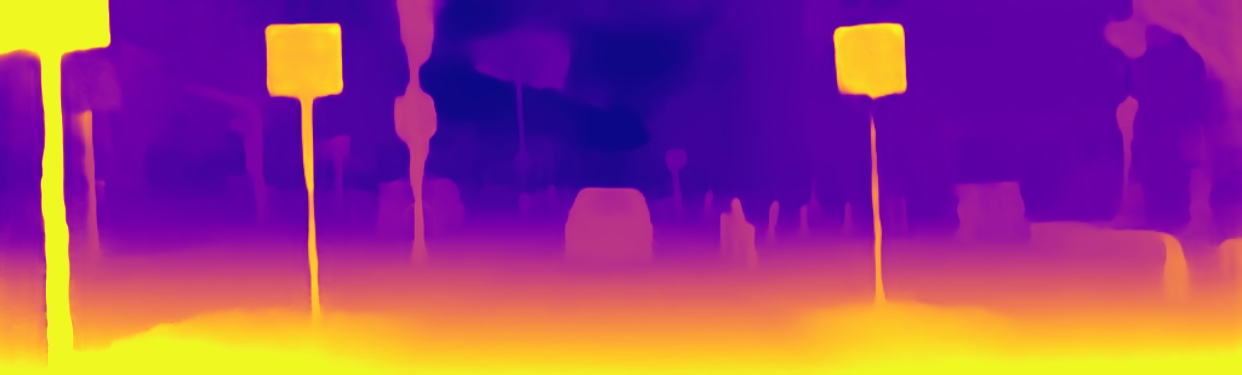}\\

\includegraphics[height=\turnheightnew]{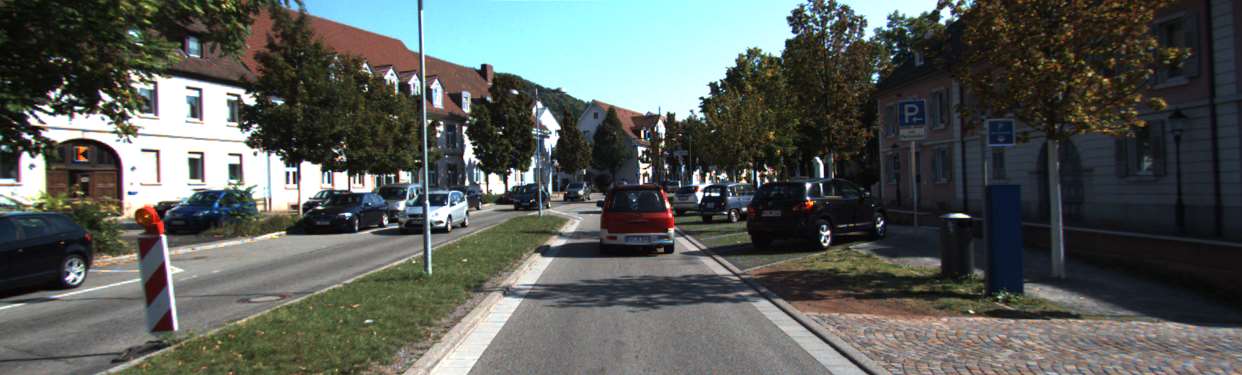} &
\includegraphics[height=\turnheightnew]{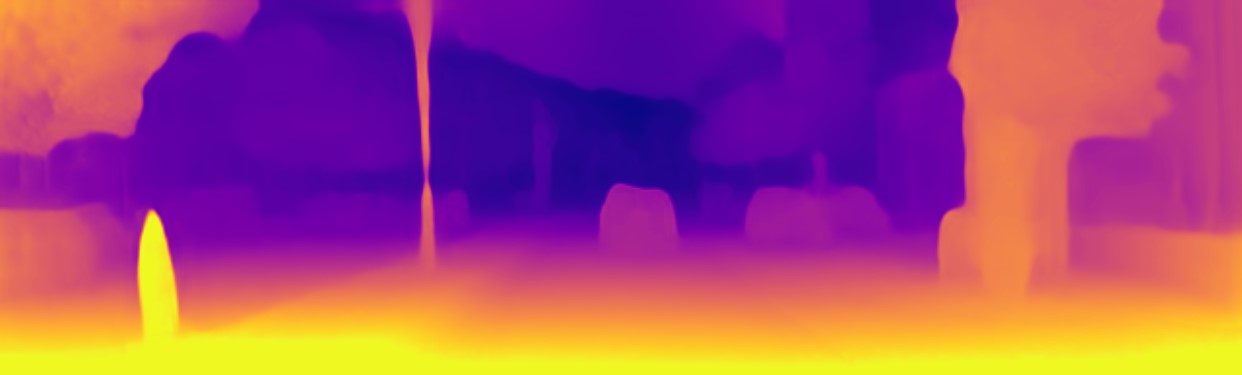} &
\includegraphics[height=\turnheightnew]{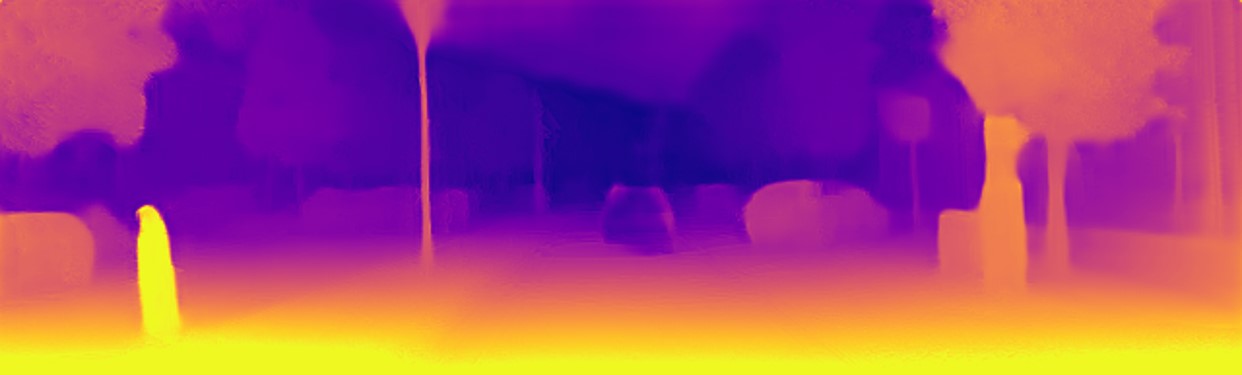} &
\includegraphics[height=\turnheightnew]{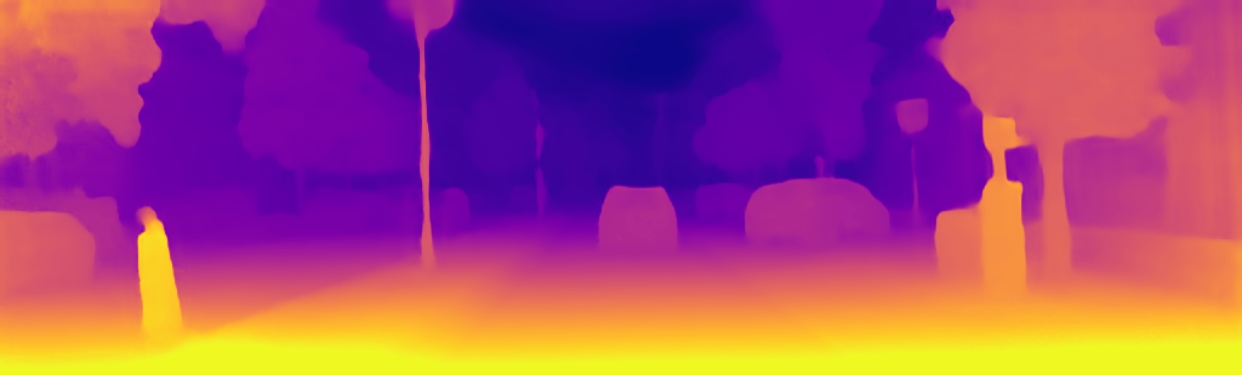}\\

\includegraphics[height=\turnheightnew]{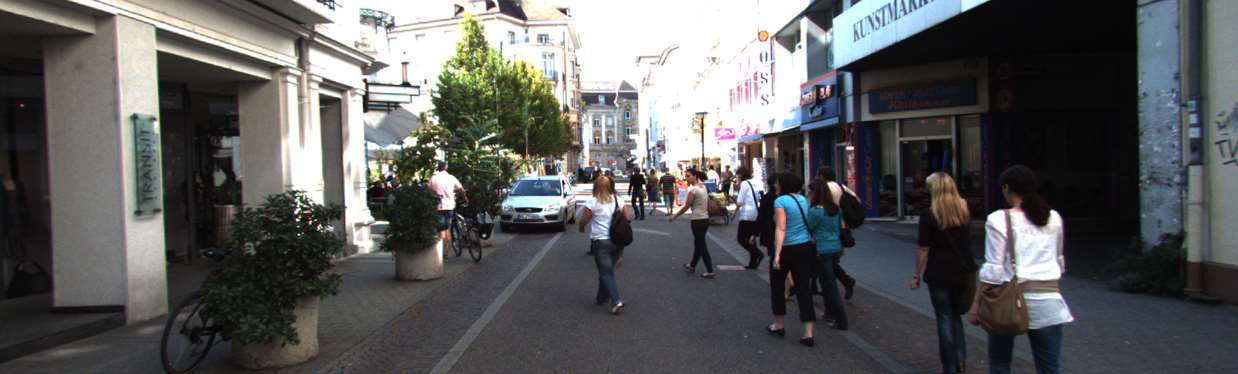} &
\includegraphics[height=\turnheightnew]{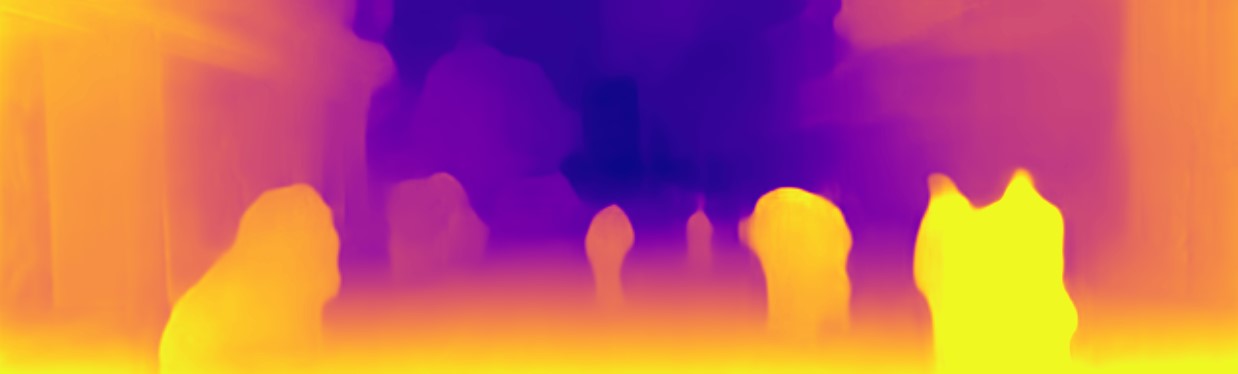} &
\includegraphics[height=\turnheightnew]{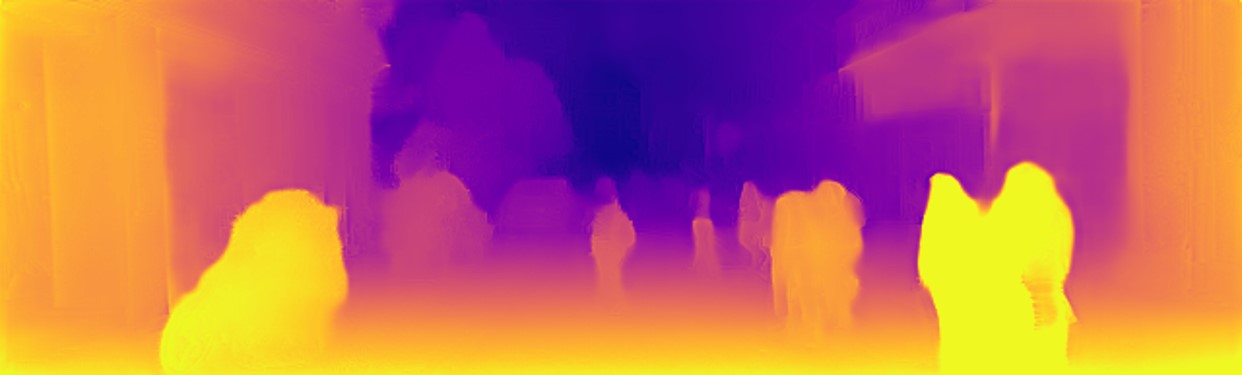} &
\includegraphics[height=\turnheightnew]{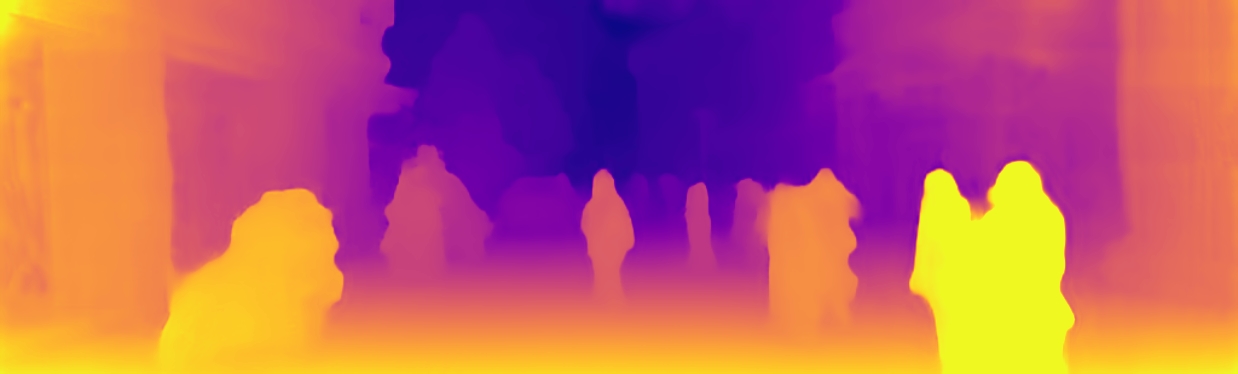}\\

\includegraphics[height=\turnheightnew]{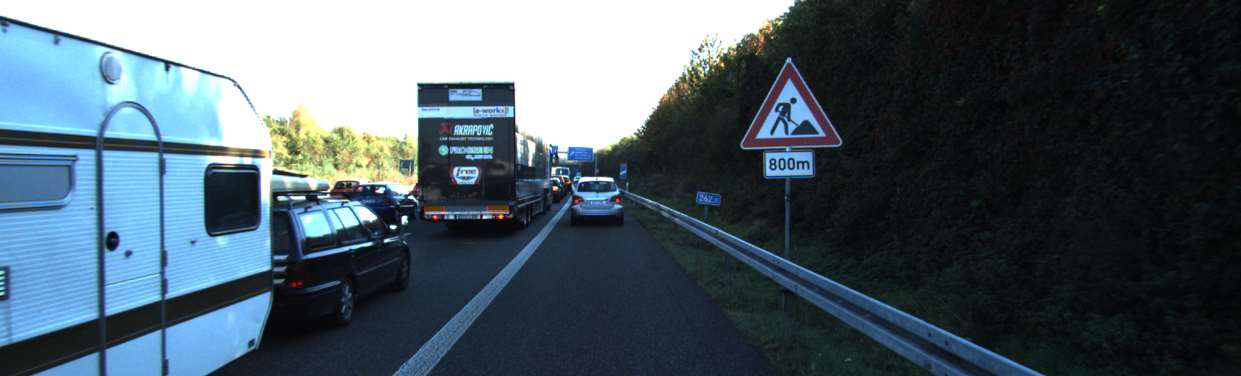} &
\includegraphics[height=\turnheightnew]{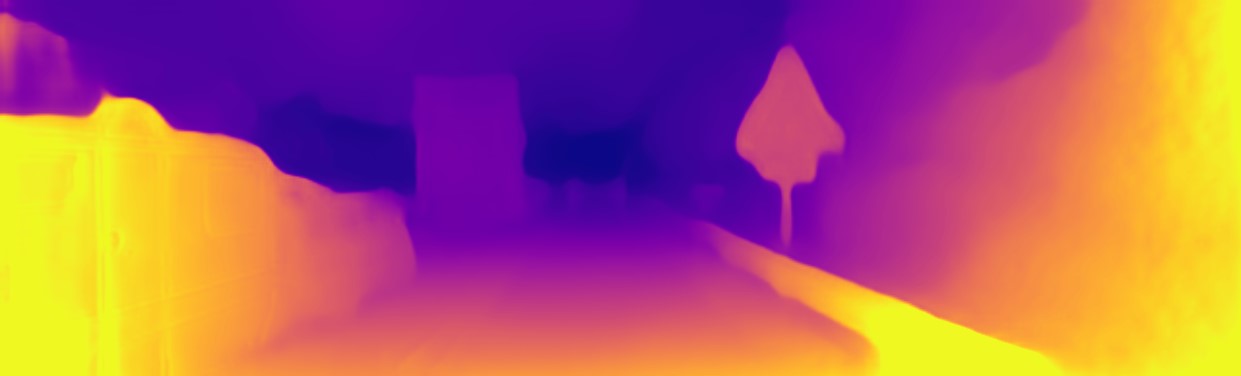} &
\includegraphics[height=\turnheightnew]{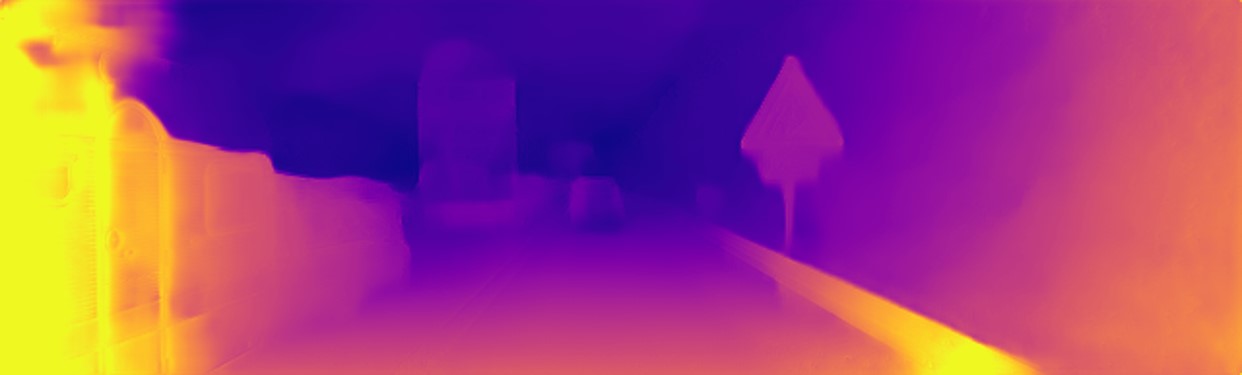} &
\includegraphics[height=\turnheightnew]{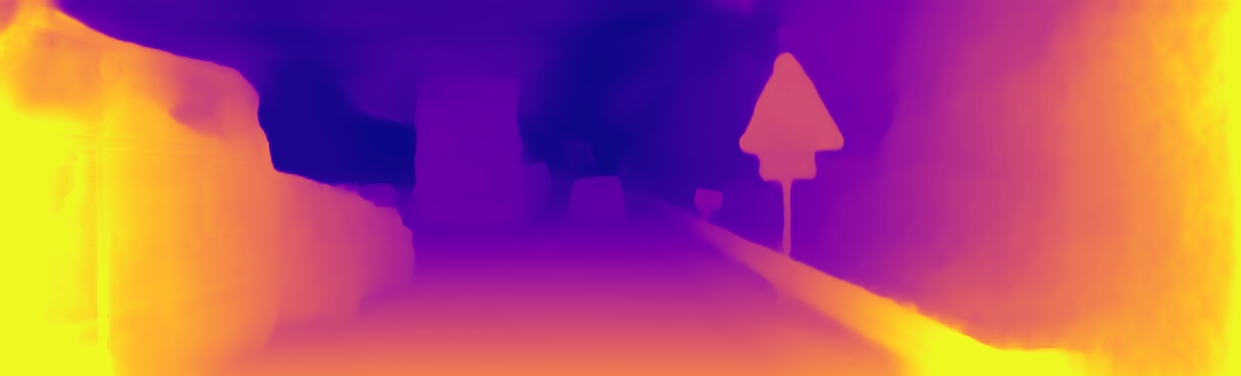}\\

\end{tabular}
 }
	\caption{\textbf{Qualitative results on the KITTI Eigen split.} Our model consistently predicts sharper boundaries and fine-gained details on thinner objects, \eg trees, pedestrians and signs.}
	\label{fig:kitti_eigen_qual}
\end{figure*}

\subsection{KITTI Results} 

We train and evaluate our methods using the KITTI 2015 stereo data set~\cite{geiger2012we}. We adopt the data split of Eigen \etal~\cite{eigen2015predicting} for distance to 80m and use pre-processing to remove static frames before training. Ultimately, this results in $39,810$ training monocular triplets, and $4,424$ for validation and $697$ for evaluation. We report results using the per-image median ground truth scaling during evaluation.

As shown in Table \ref{tab:kitti_eigen}, our proposed CADepth-Net significantly outperforms the existing SoTA self-supervised approaches in all metrics. In addition, we score dramatically higher in the hardest accuracy metric $\delta<1.25$, which indicates that our model predicts more accurate and realistically detailed depth estimation than all other competing models. For a fair comparison, we also give the results on various input resolution and training settings, and our model still improves the performance at higher image resolutions. Fig.~\ref{fig:kitti_eigen_qual} shows that our model produces sharper depth estimation on thinner structures \eg road signs and poles. Moreover, our model successfully estimates correct depth at the highly reflective car roof ($6^{th}$ row), which are the challenging problems for previous advanced methods. These improvements can be explained by the better perception of scenes and objects afforded by the structure perception module, and the further regularisation provided by detail emphasis module. Additional results can be seen in supplementary material.

\begin{figure*}[!ht]
	\centering
	\includegraphics[width=\linewidth]{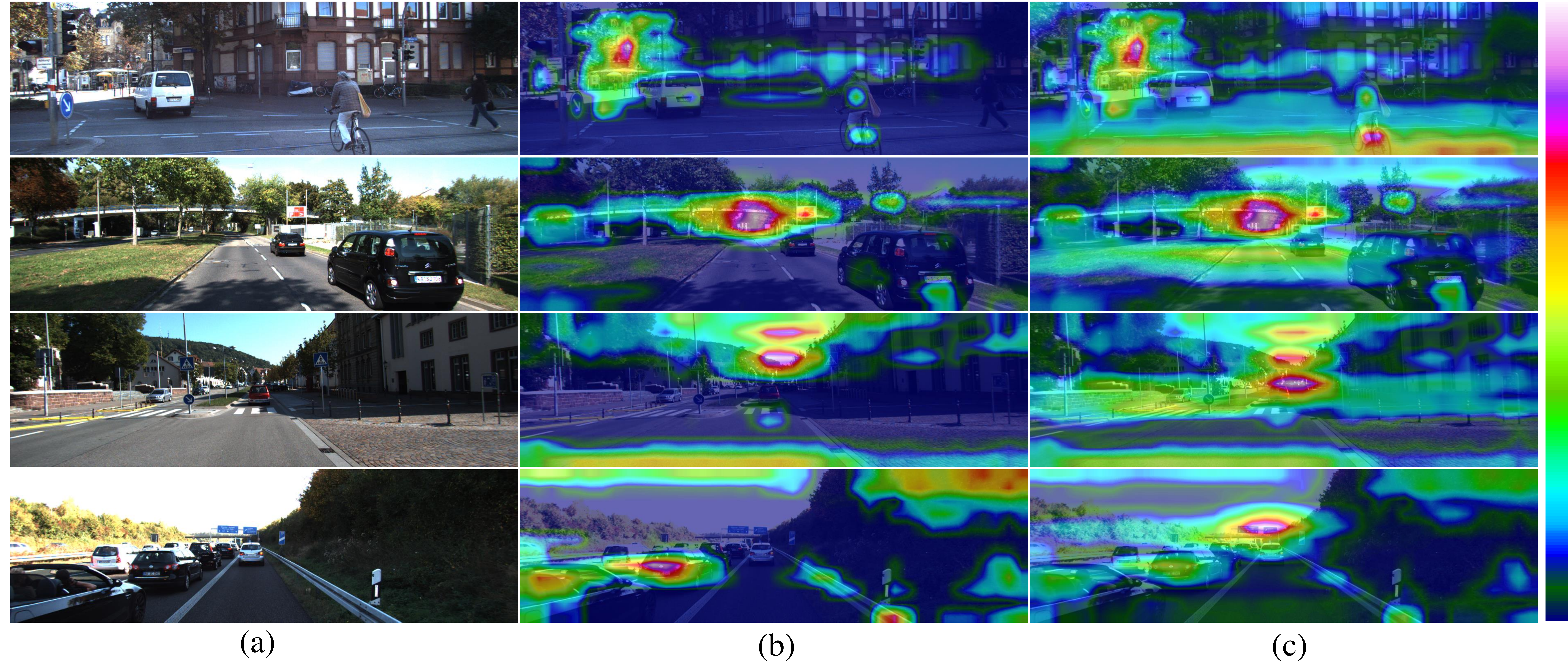}
	\caption{\textbf{The visualizations of the structure perception module.} (a) Input image. (b) Input feature maps. (c) Corresponding output of the structure perception module. All feature maps are projected onto RGB images for clear visualization. The structure perception module explicitly enhances the perception of scene structure and feature representation. 
	}
	\label{fig:SPM_vis}
\end{figure*}

\subsection{Make3D result}

\begin{figure}
	\centering
	\resizebox{\columnwidth}{!}{
		\newcommand{\turnheightnew}{0.2\columnwidth}
\centering

\begin{tabular}{@{\hskip -1mm}c@{\hskip 0.5mm}c@{\hskip 0.5mm}c@{\hskip 0.5mm}c@{\hskip -1mm}}

\Large{Input}  & \Large{MD2 (M)} \cite{godard2019digging}&  \Large {Ours (M)}  & \Large{Ground truth} \\

\includegraphics[height=\turnheightnew]{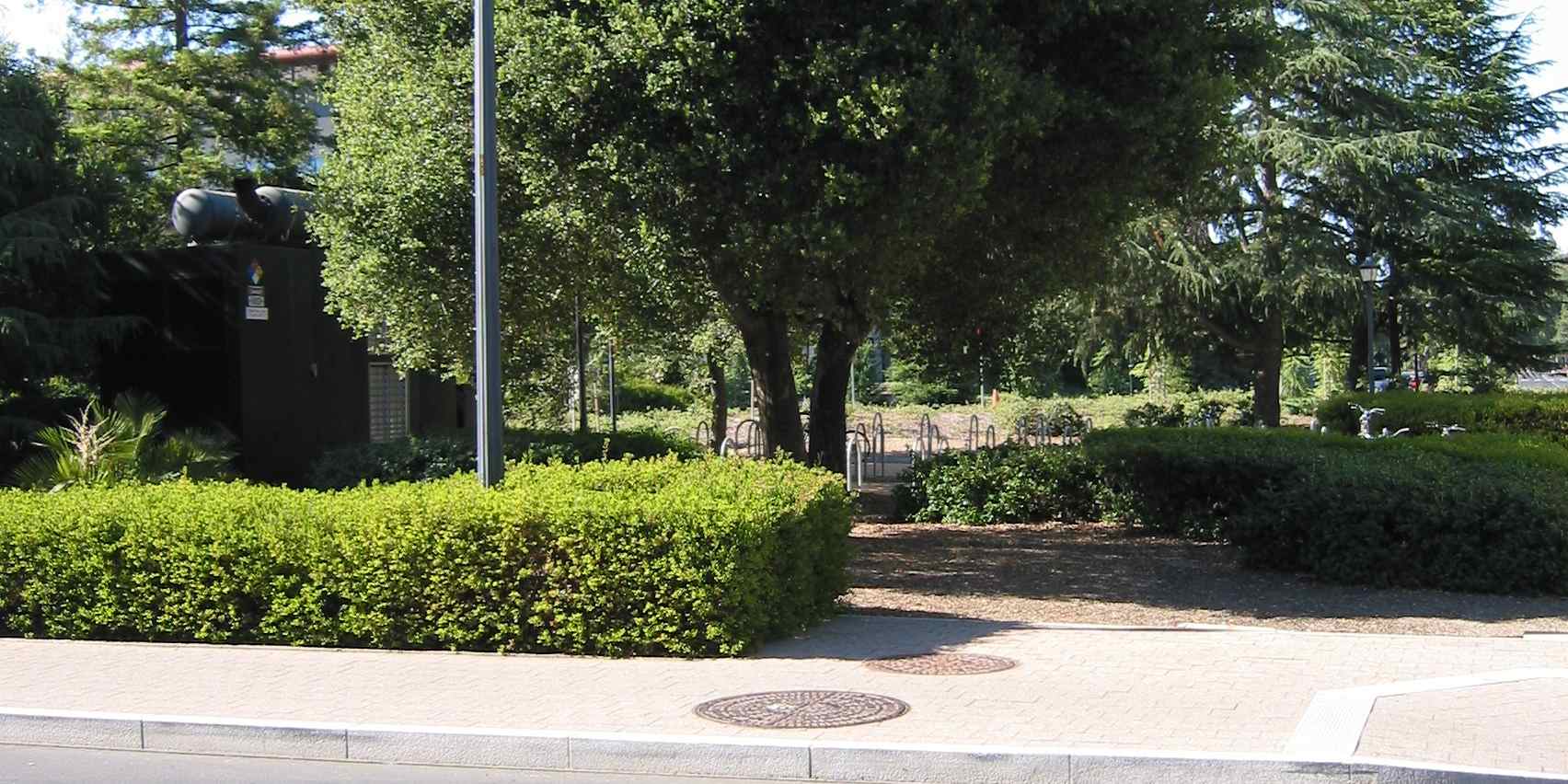} &
\includegraphics[height=\turnheightnew]{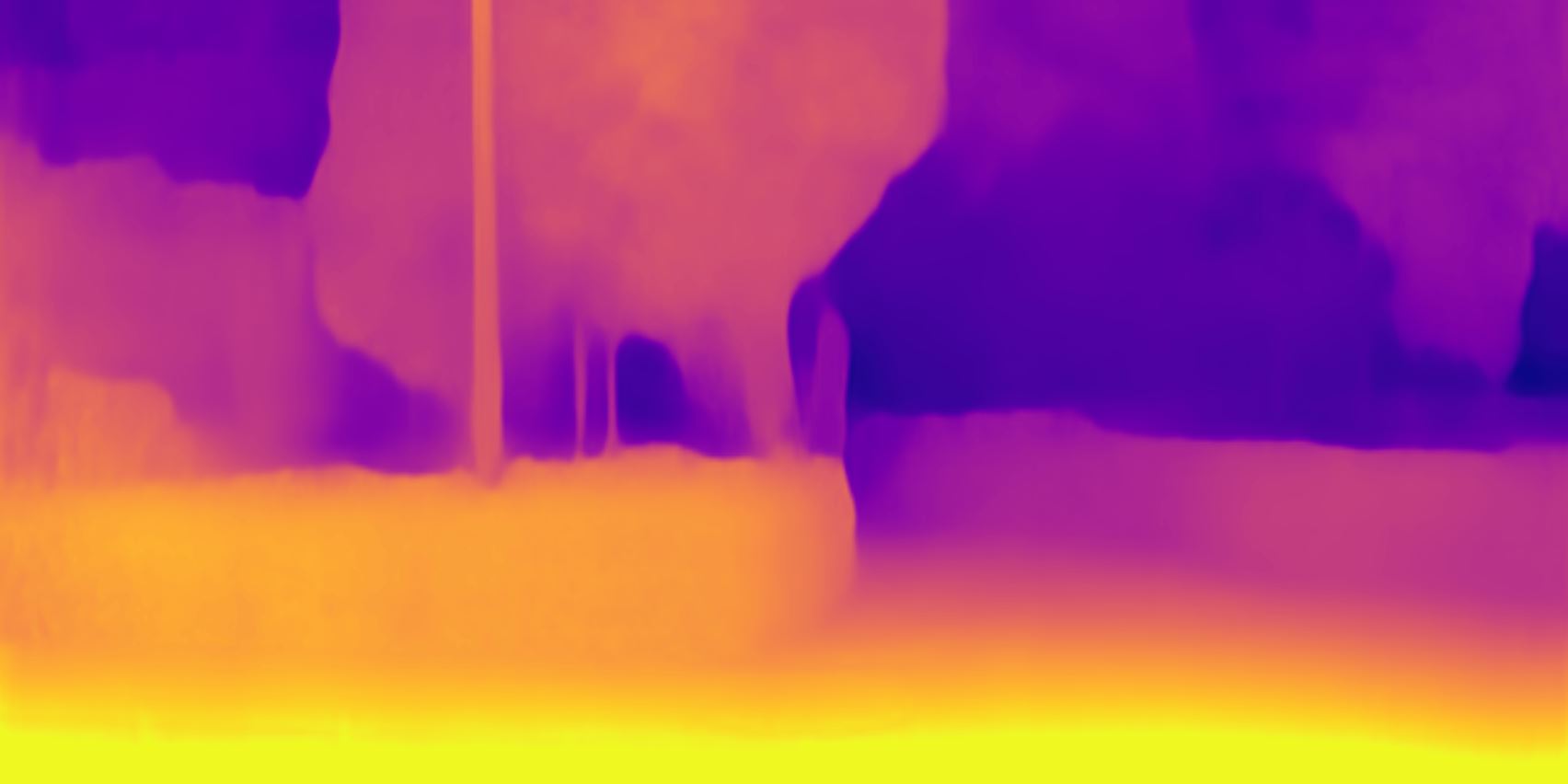} &
\includegraphics[height=\turnheightnew]{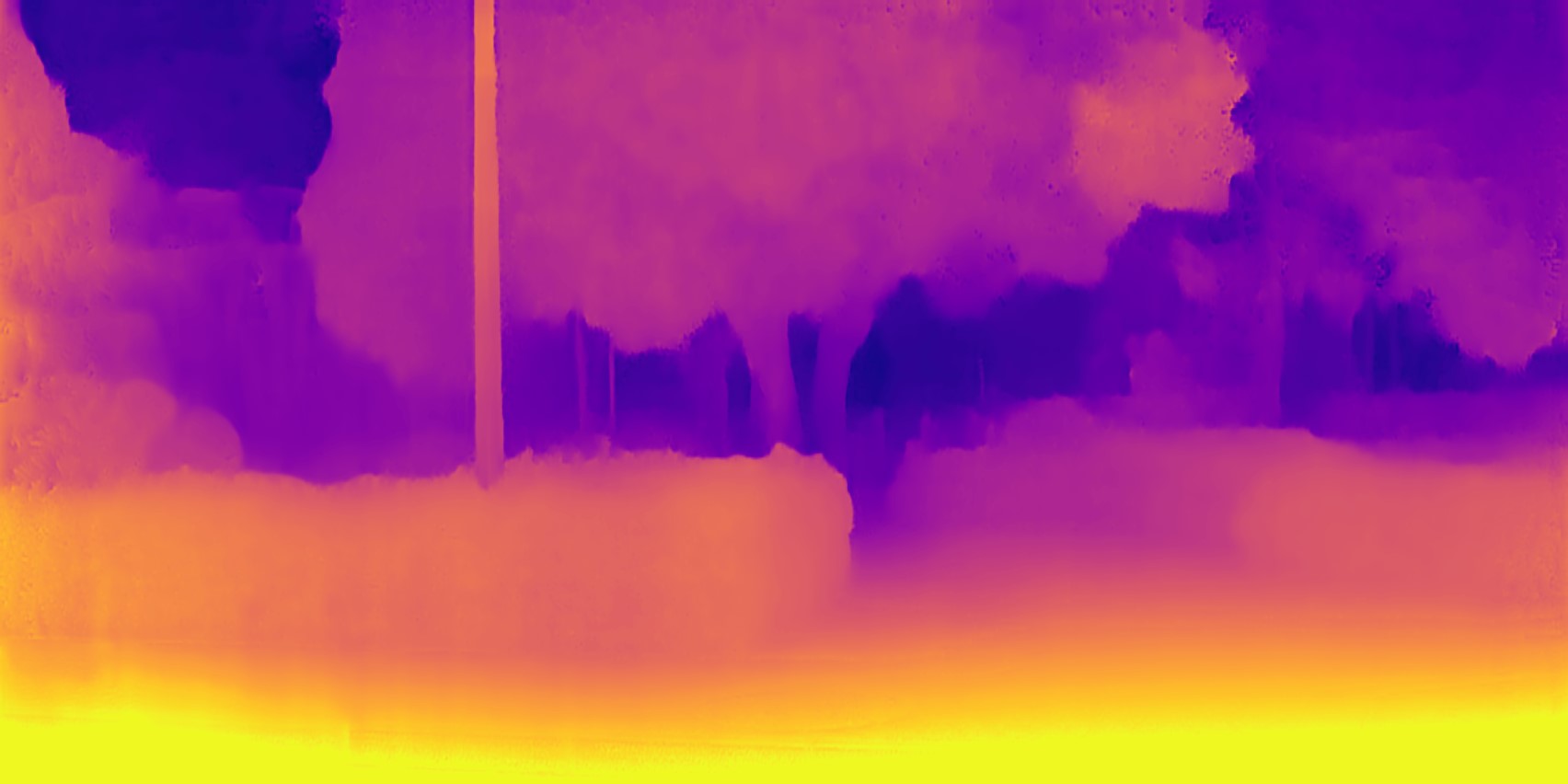} &
\includegraphics[height=\turnheightnew]{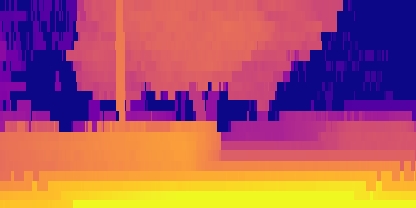}\\

\includegraphics[height=\turnheightnew]{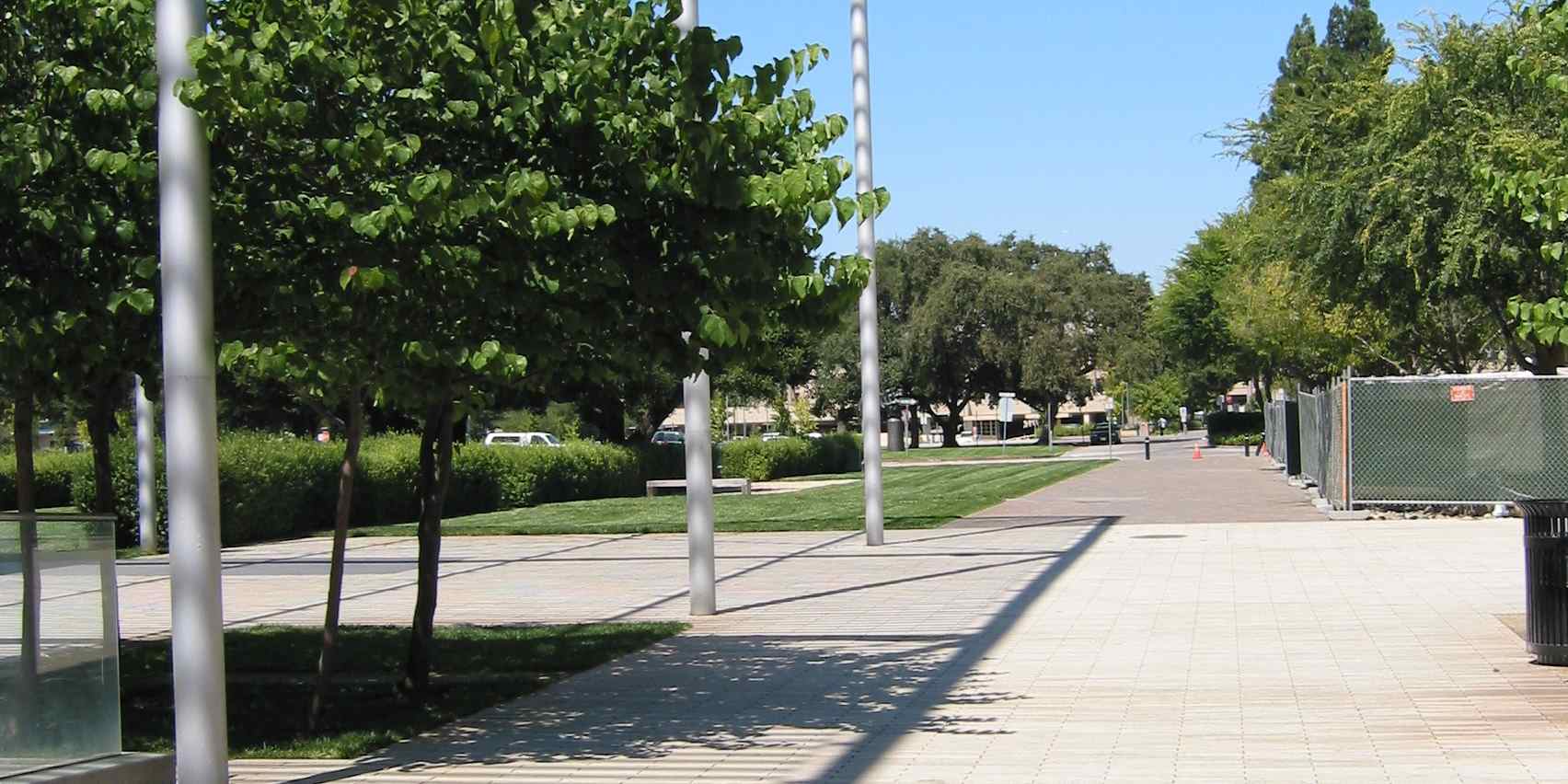} &
\includegraphics[height=\turnheightnew]{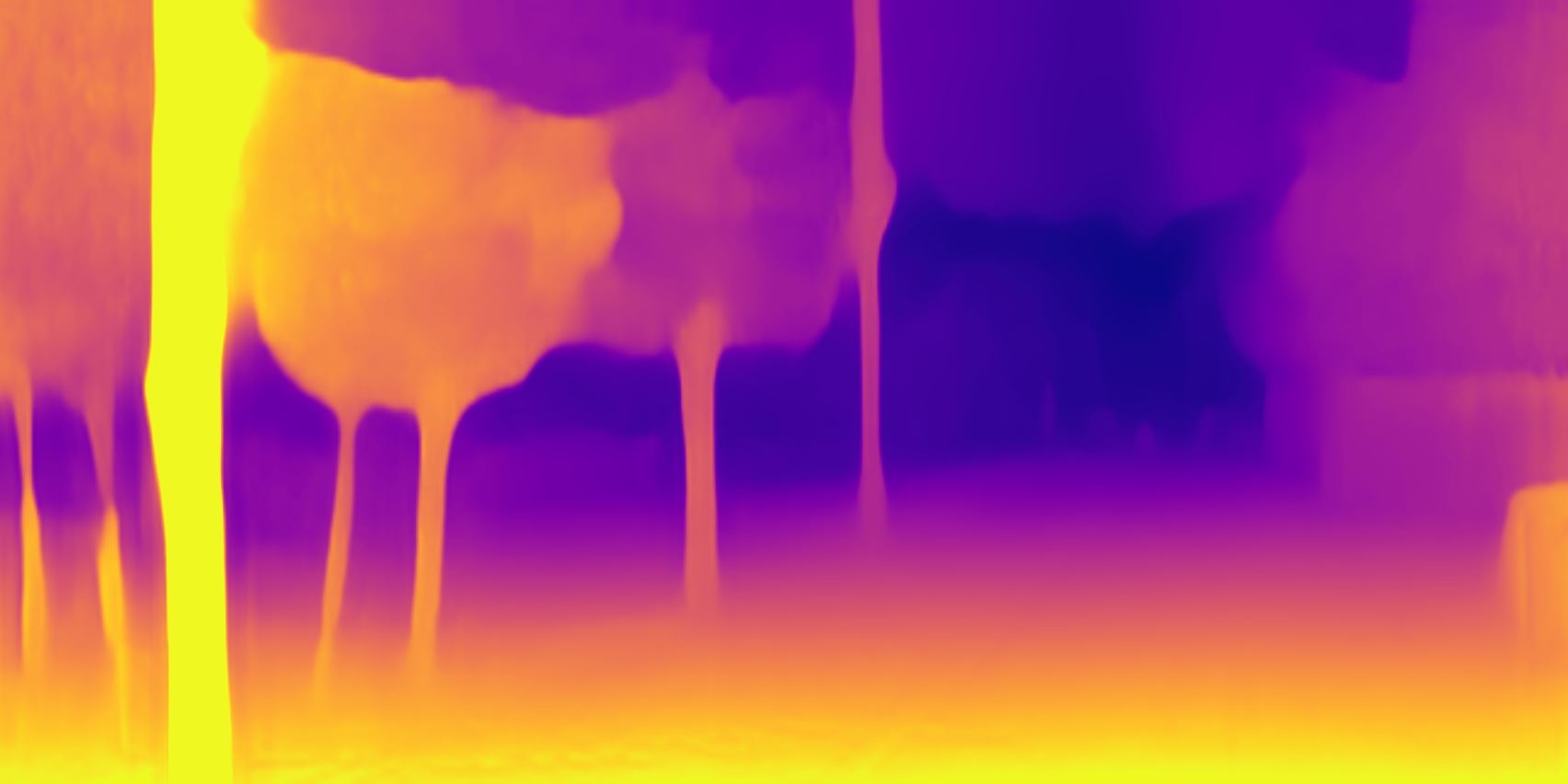} &
\includegraphics[height=\turnheightnew]{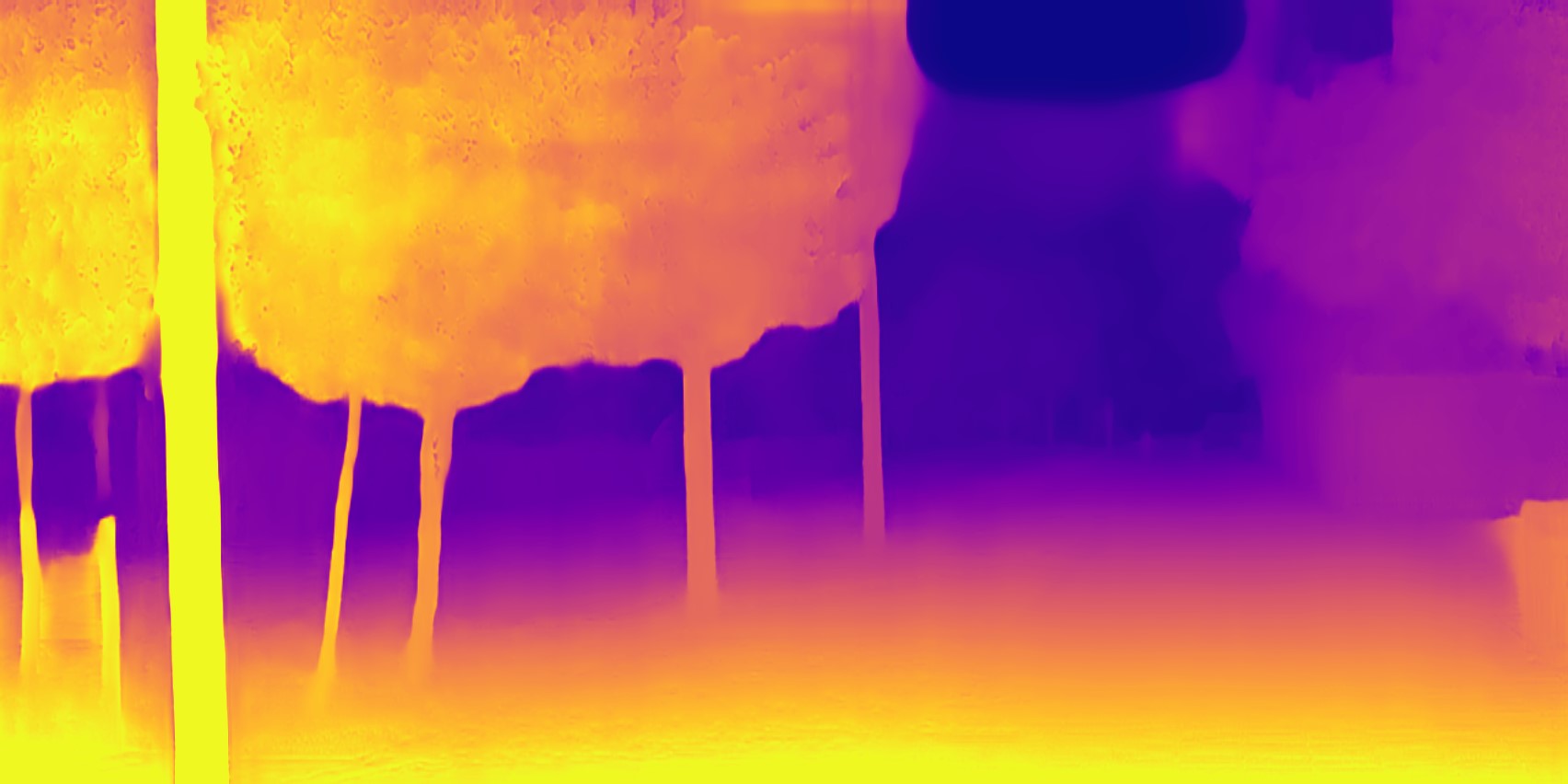} &
\includegraphics[height=\turnheightnew]{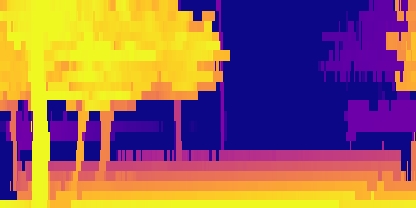}\\

\includegraphics[height=\turnheightnew]{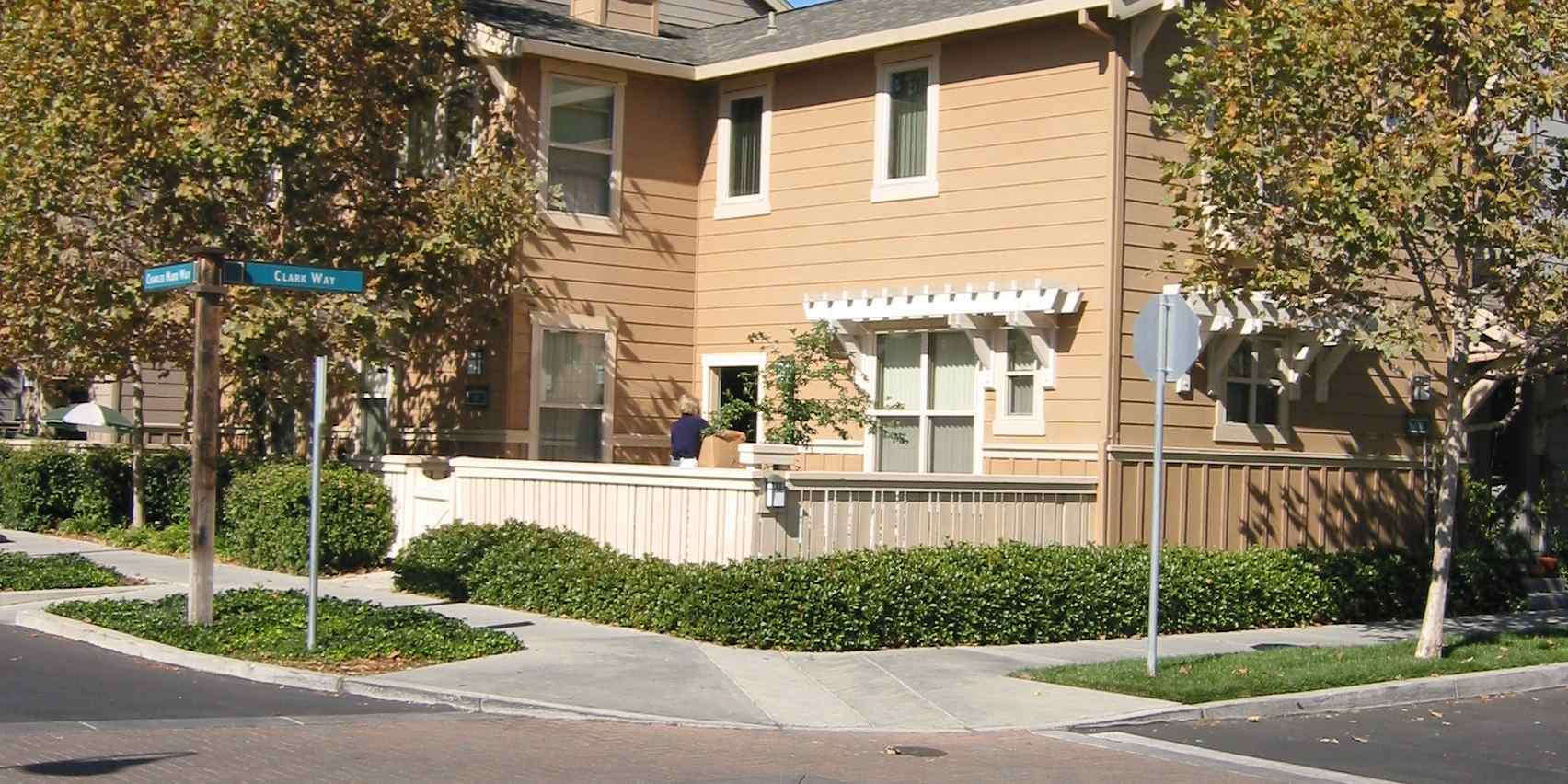} &
\includegraphics[height=\turnheightnew]{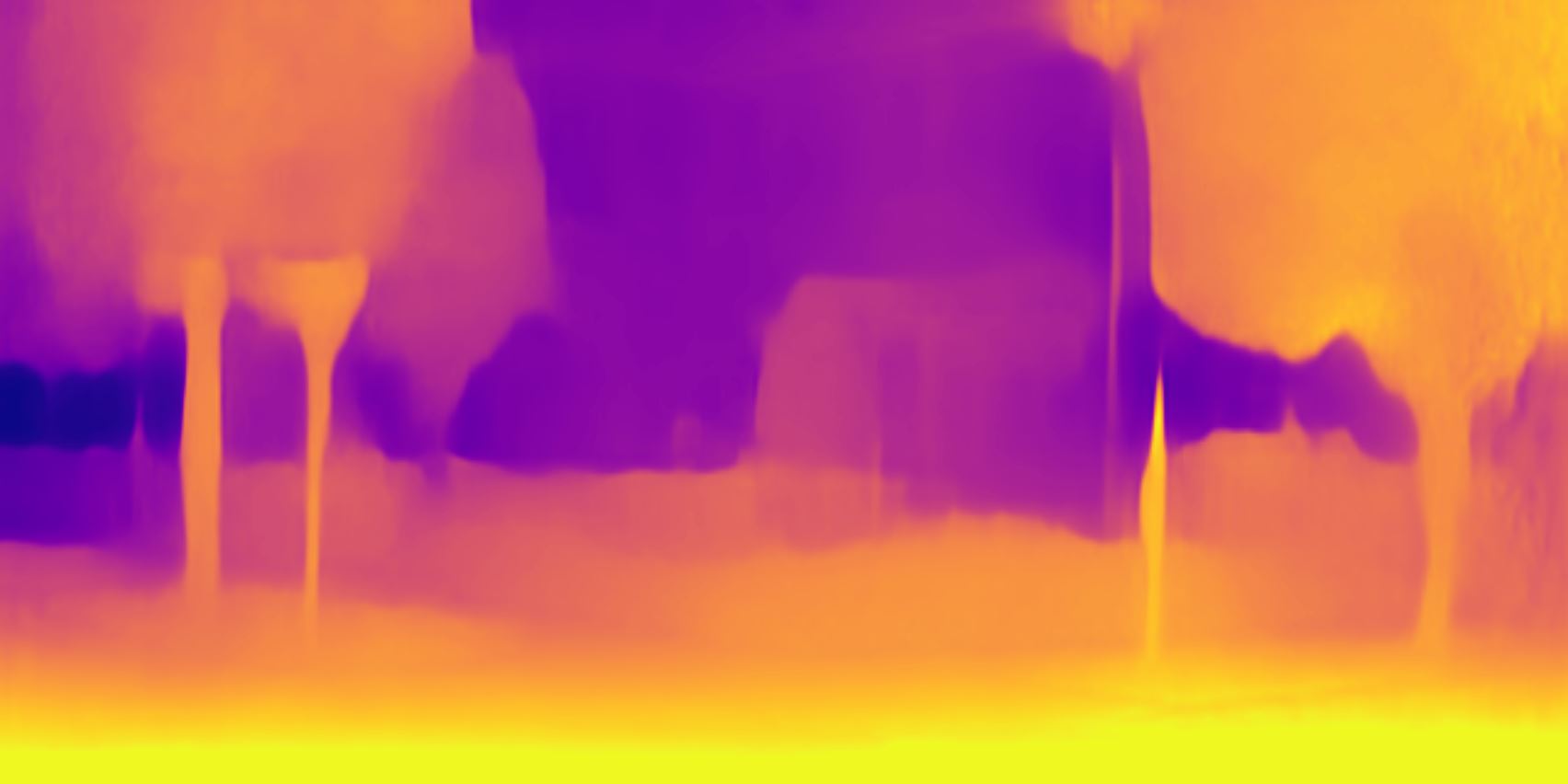} &
\includegraphics[height=\turnheightnew]{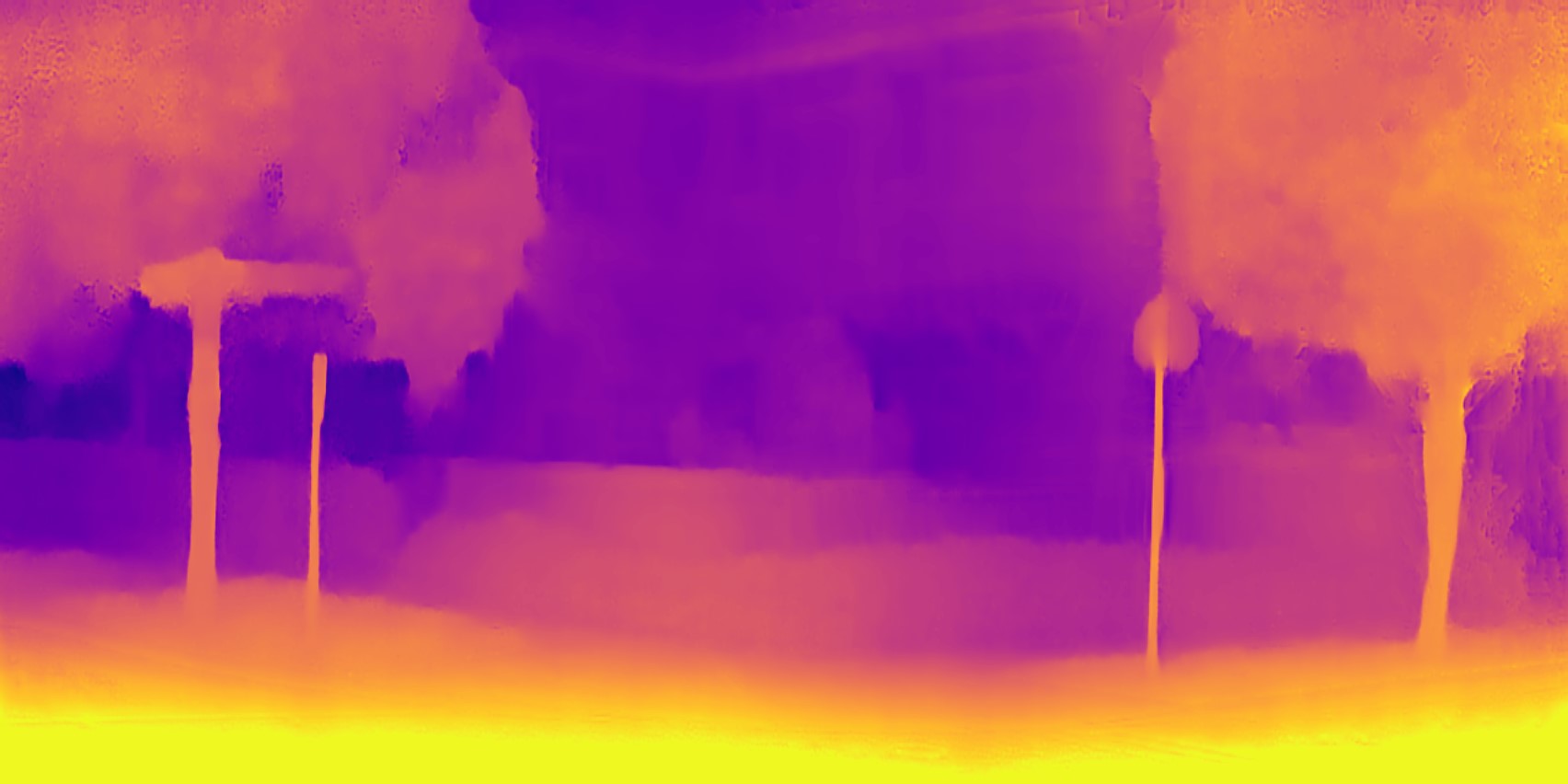} &
\includegraphics[height=\turnheightnew]{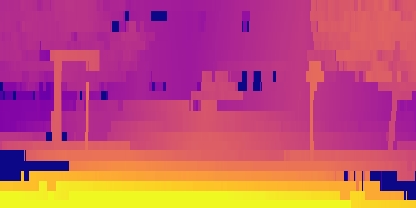}\\

\end{tabular}
 }
	\caption{\textbf{Qualitative Make3D results.} All methods were trained on KITTI using monocular supervision.}
	\vspace{1pt}
	\label{fig:make3d_results}
\end{figure}

To evaluate the generalization ability of our model on the unseen dataset, we report the quantitative results for the Make3D dataset~\cite{saxena2008make3d} using our model trained on KITTI 2015~\cite{geiger2012we}. Following the same evaluation protocol as~\cite{godard2017unsupervised}, we test on a center crop of $2 \times 1$ ratio and apply median scaling. As shown in Table \ref{tab:make3d}, our approach produces superior results compared with the other SOTA self-supervised methods. Qualitative results can be seen in Fig.~\ref{fig:make3d_results}, which show that our model generates more accurate and sharper depth estimation on the previously unseen Make3D dataset.

\subsection{Structure Perception Module}

To demonstrate the effectiveness of the structure perception module, Fig.~\ref{fig:SPM_vis} presents a comparison of features before and after treatment. For clear visualization, we map the channel maps to the RGB color cube and project them to the original RGB image. Fig.~\ref{fig:SPM_vis} (b) refers to the high-level features produced by encoder, which mainly indicates region-specific responses and various kinds of structural information of 3D scene, \eg vanish point, region with same depth range and the area with same color or texture (\eg sky) . 

As mentioned earlier, the structure perception module produces the aggregated features (Fig.~\ref{fig:SPM_vis} (c)) by performing the weighted summation of all channel maps. As the attention map describes the feature discrimination and the spatial relationship of region responses between channels, by aggregating discriminative features at channel dimension, we can make each single channel map get rich scene structure representation and more complete region responses from non-contiguous regions. By doing so, the structure perception module obtains rich contextual information of overall scene geometry perception, which results in better scene understanding and feature representation for depth estimation. 

Fig. \ref{fig:SPM_vis} adequately demonstrates that each channel map obtains more extra depth perception from distant regions \eg foreground ($1^{st}$ row) and midground ($2^{nd}$ row). In addition, it also particularly emphasizes the vanishing point regions, which are naturally a strong cue to understand the geometry of a scene. In the last row, the network originally focuses on foreground objects such as cars and obtains more relative depth relationships after adding the vanishing point information. More visualization results are described in the supplementary material.

\begin{table*}[t!]
	\centering
	\resizebox{\textwidth}{!}
	{
		\footnotesize
		\begin{tabular}{|l|c|c|c||c|c|c|c|c|c|c|c|c|}
			\hline
			& Backbone & 
			\begin{tabular}{@{}c@{}}SPM\end{tabular} & 
			\begin{tabular}{@{}c@{}}DEM\end{tabular} & 
			Para & Time &
			\cellcolor{col1}Abs Rel & \cellcolor{col1}Sq Rel & \cellcolor{col1}RMSE  &
			\cellcolor{col1}\begin{tabular}{@{}c@{}}RMSE \\ log\end{tabular} & 
			\cellcolor{col2}$\delta < $1.25 & \cellcolor{col2}$\delta < $1.25$^{2}$ & \cellcolor{col2}$\delta <$ $1.25^{3}$ \\
			
			\hline
			
			Baseline (MD2 ResNet18~\cite{godard2019digging})& R18 &  &  & 14.84 M & 11.95 ms & 0.115 & 0.903 & 4.863 & 0.193 & 0.877 & 0.959 & 0.981 \\
			
			Baseline + SPM & R18 & \checkmark & & 14.84 M & 12.13 ms & 0.110 & 0.843 & 4.739 & 0.188 & \bf{0.883} & 0.961 & 0.982 \\
			
			Baseline + DEM & R18 & & \checkmark & 18.74 M & 15.44 ms & 0.111 & 0.851 & 4.746 & 0.189 & 0.881 & 0.961 & 0.982 \\
			
			\textbf{CADepth-Net ResNet18} (full) & R18 & \checkmark & \checkmark & 18.74 M & 15.77 ms & \bf{0.110}& \bf{0.812} & \bf{4.686} & \bf{0.187} & 0.882 & \bf{0.962} & \bf{0.983} \\					
			
			\hline
			
			Baseline (MD2 ResNet50~\cite{godard2019digging}) & R50 &  &  & 34.57 M & 22.52 ms  & 0.110 & 0.831 & 4.642 & 0.187 & 0.883 & 0.962 & 0.982 \\
			
			Baseline + SPM & R50 & \checkmark & & 34.57 M & 22.80 ms  & 0.107 & 0.784 & 4.589 & 0.185 & 0.887 & 0.963 & 0.982 \\
			
			Baseline + DEM & R50 &  & \checkmark & 58.34 M & 28.01 ms  & 0.107 & \bf{0.759} & 4.557 & 0.183 & 0.884 & \bf{0.964} & \bf{0.983} \\
			
			\textbf{CADepth-Net ResNet50} (full) & R50 & \checkmark & \checkmark & 58.34 M & 28.41 ms  & \bf{0.105} & 0.769 & \bf{4.535} & \bf{0.181} & \bf{0.892} & \bf{0.964} & \bf{0.983} \\
			
			\hline
		\end{tabular}
	}
	\vspace{1pt}
	\caption{
		\textbf{Ablation Studies.} 
		We evaluate the performance of our structure perception module (SPM) and detail emphasis module (DEM) contributions with Monodepth2 (MD2)~\cite{godard2019digging} as the baseline. R: ResNet, Para: parameters. All models in this table are trained with monocular self-supervised (M) and standard resolution (640 $\times$ 192). The inference time is tested on a single RTX3090 GPU.}
	\label{tab:kitti_eigen_ablation}
\end{table*}

\begin{figure*}
	\centering
	\includegraphics[width=\linewidth]{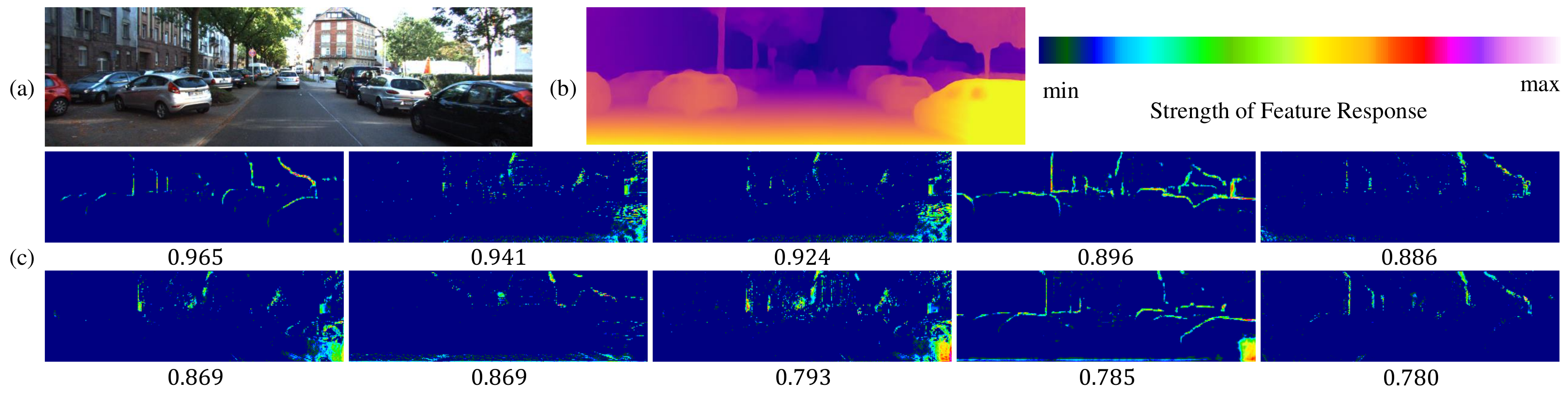} 
	\caption{\textbf{The visualizations of detail emphasis module.} (a) Input image. (b) Predicted depth map. (c) Top $n$ ($n=10$) feature maps with highest weights score that range in $[0, 1]$. Detail emphasis module mainly highlights the crucial local details. 
	}
	\label{fig:DEM_vis}
\end{figure*}

\subsection{Detail Emphasis Module}
The detail emphasis module generates a weights vector to re-calibrate channel maps and emphasizes the informative features for subsequent transformations. The weights scores produced by the sigmoid function are between 0 and 1, indicating the importance of corresponding features, \ie~the more important of the channel, the higher score is earned. As shown in Fig. \ref{fig:DEM_vis}, we report the top $n$ ($n=10$) feature maps with the highest scores to show which features the network focuses on. We observe that our model mainly highlights the crucial low-level features that describe the object boundaries precisely. This observation is consistent with the fact that the decoder mainly uses detail information to recover spatial resolution gradually. In general, the detail emphasis module adaptively selects the informative local details and helps network handle and locate the object edges for sharper depth prediction. More visualization results are provided in the supplementary material.

\subsection{Ablation Study}

For further demonstrating the performance improvements of our provided methods, Table \ref{tab:kitti_eigen_ablation} and supplemental material show the ablation study of our various components with Monodepth2~\cite{godard2019digging} as the baseline. It shows that all of our contributions achieve a steady improvement in almost all evaluation measures and obtain a consistent performance gain on different backbones, showing the robustness of our CADepth-Net to the backbone architecture capacity. Note that the structure perception module shows superior performance on metric $\delta<1.25$, with little time cost and no additional parameters, demonstrating the improvements benefit from the better scene understanding rather than an increase in network complexity. Finally, the combination of all our modules with ResNet50 achieves the best results, with 59M parameters and an inference time of 28ms on an RTX3090 GPU, meets the requirements of real-time applications.

\begin{table}
	\centering
	\resizebox{\columnwidth}{!}{
		\begin{tabular}{|l|c||c|c|c|c|}
			\hline
			& Type & Abs Rel & Sq Rel  & RMSE & $\text{log}_{10}$ \\
			\hline
			Karsch~\cite{karsch2014depth} & D & 0.428 & 5.079 & 8.389 & 0.149 \\
			Liu~\cite{liu2014discrete} & D & 0.475 & 6.562 & 10.05 & 0.165 \\
			Laina~\cite{laina2016deeper} & D & \bf{0.204} & \bf{1.840} & \bf{5.683} & \bf{0.084} \\ \hline
			Monodepth~\cite{godard2017unsupervised} & S & 0.544 & 10.94 & 11.760 & 0.193 \\
			Zhou~\cite{zhou2017unsupervised} & M &  0.383 & 5.321 & 10.470 & 0.478 \\
			DDVO~\cite{wang2018learning} & M & 0.387 & 4.720 & 8.090 & 0.204 \\
			Monodepth2~\cite{godard2019digging} & M & 0.322 & 3.589 & 7.417 & 0.163 \\
			\bf{Ours} & M & \bf{0.312} & \bf{3.086} & \bf{7.066} & \bf{0.159} \\
			\hline
		\end{tabular}
	}
	\vspace{1pt}
	\caption{{\bf Make3D results.} All self-supervised mono (M) results benefit from median scaling. }
	\label{tab:make3d}
\end{table}

\section{Conclusion}

In this paper, we introduce a novel architecture named channel-wise attention-based depth estimation network, with two effective components, the structure perception module and the detail emphasis module. The structure perception module aggregates the discriminative features by capturing the long-range dependencies to obtain the context of scene structure and rich feature representation. Additionally, the detail emphasis module employs the channel attention mechanism to highlight objects' boundaries information and efficiently fuse different level features. Furthermore, the experiments demonstrate that our CADepth-Net produces sharper depth estimation and achieves the state-of-the-art results on KITTI datasets.

{\small
\bibliographystyle{ieee_fullname}
\bibliography{main}
}

\clearpage

\begin{appendices}

This document provides additional details and results concerned with paper ”Channel-Wise Attention-Based Network for Self-Supervised Monocular Depth Estimation”.  The supplementary material is organized as follows: Section 1 provides the details of depth estimation network, Section 2 and Section 3 reports the quantitative results on the KITTI improved ground truth and online evaluation server, Section 4 collects additional ablation experiments, Section 5 provides the additional qualitative comparisons, and Section 6 shows visualization results of our methods.

\section{Network Architecture}

Except where note, for all experiments, we use a ResNet50 encoder with pretrained ImageNet weights for both depth and pose networks. For depth model, the structure perception module takes the features from encoder as input. In addition, we successively adopt the detail emphasis module at different scales in decoder stage. Note that although there is no skip-connection at the highest resolution, we still use the detail emphasis module for further regularization. For high resolution input, \eg $1024 \times 320$ and $1280 \times 384$, we employ a lightweight setup, ResNet18 and $640 \times 192$, for pose encoder at training for memory savings. The depth network architecture is shown in Table \ref{tab:network}.

\section{KITTI Improved Ground Truth}

The evaluation method proposed by Eigen \etal~\cite{eigen2015predicting} for KITTI uses the reprojected LIDAR points to generate the ground truth depth, but does not handle moving objects and occlusions. \cite{uhrig2017sparsity} created a set of high-quality depth maps for the KITTI dataset using five consecutive frames and stereo pairs to handle moving objects better. This improved ground truth depth is provided for 652 (or 93\%) of the 697 original test frames contained in the Eigen test split \etal~\cite{eigen2015predicting}. We utilize the same error metrics and evaluation strategy as the main paper, and evaluate our model on these 652 improved ground truth frames without having to retrain each method. As shown in Table \ref{tab:kitti_improved}, we observe that our CADepth-Net still outperforms all existing advanced methods on all metrics. 

\begin{table}
	\centering
	\resizebox{\columnwidth}{!}{
		\begin{tabular}[t]{|l|l|l|l|l|l|l|}
\hline
\multicolumn{7}{|l|}{\textbf{Depth network}} \\
\hline
\textbf{layer} & \textbf{k} & \textbf{s} & \textbf{chns} & \textbf{res} & \textbf{input}   & \textbf{activation}    \\ 
\hline
conv1         & 7      & 2      & 64       & 2    & image & RELU \\
maxpool       & 3      & 2      & 64       & 4    & conv1 & - \\

\hline
econv1        & 3      & 1      & 256      & 4     & maxpool & RELU \\
econv2        & 3      & 2      & 512      & 8     & econv1  & RELU \\
econv3        & 3      & 2      & 1024     & 16    & econv2  & RELU \\
econv4        & 3      & 2      & 2048     & 32    & econv3  & RELU \\
spm           & -      & -      & 2048     & 32    & econv4  & - \\
\hline

upconv5       & 3      & 1      & 256      & 32    & spm                       & ELU \cite{clevert2015fast} \\
dem5          & 3      & 1      & 1280     & 16    & $\uparrow$upconv5, econv3 & RELU \\
iconv5        & 3      & 1      & 256      & 16    & dem5                      & ELU \\ 
\hline

upconv4       & 3      & 1      & 128      & 16    & iconv5                    & ELU \\
dem4          & 3      & 1      & 640      & 8     & $\uparrow$upconv4, econv2 & RELU \\
iconv4        & 3      & 1      & 128      & 8     & dem4                      & ELU \\ 
disp4         & 3      & 1      & 1        & 1     & iconv4                    & Sigmoid \\ 
\hline

upconv3       & 3      & 1      & 64       & 8     & iconv4                    & ELU \\
dem3          & 3      & 1      & 320      & 4     & $\uparrow$upconv3, econv1 & RELU \\
iconv3        & 3      & 1      & 64       & 4     & dem3                      & ELU \\ 
disp3         & 3      & 1      & 1        & 1     & iconv3                    & Sigmoid  \\ 
\hline

upconv2       & 3      & 1      & 32       & 4     & iconv3                    & ELU \\
dem2          & 3      & 1      & 96       & 2     & $\uparrow$upconv2, conv1  & RELU \\
iconv2        & 3      & 1      & 32       & 2     & dem2                      & ELU \\ 
disp2         & 3      & 1      & 1        & 1     & iconv2                    & Sigmoid \\ 
\hline

upconv1       & 3      & 1      & 16       & 2     & iconv2                    & ELU \\
dem1          & 3      & 1      & 16       & 1     & $\uparrow$upconv1         & RELU \\
iconv1        & 3      & 1      & 16       & 1     & dem1                      & ELU \\ 
disp1         & 3      & 1      & 1        & 1     & iconv1                    & Sigmoid \\ 
\hline
\end{tabular}

 }
	\vspace{1pt}
	\caption{\textbf{Depth network architecture.} 
		For symbols in this table, \textbf{k}: kernel size, \textbf{s}: stride, \textbf{chns}: the number of output channels, \textbf{res}: the downscaling factor relative to the input image, \textbf{input}: corresponds to the input of each layer, \textbf{activation}: activation function. $\uparrow$ refers to a $2\times$ nearest-neighbor upsampling. Encoder blocks are denoted by $econv*$ naming convention. The $spm$ and $dem$ represent structure perception module and detail emphasis module.
	}
	\label{tab:network}
\end{table}

\begin{table*}[t]
	\centering
	\resizebox{\linewidth}{!}{
		\footnotesize
		\begin{tabular}{|l|c|c|c||c|c|c|c|c|c|c|}
			\hline
			Method & Train & Res&  Dataset &\cellcolor{col1}Abs Rel & \cellcolor{col1}Sq Rel & \cellcolor{col1}RMSE  & \cellcolor{col1}RMSE log & \cellcolor{col2}$\delta < 1.25 $ & \cellcolor{col2}$\delta < 1.25^{2}$ & \cellcolor{col2}$\delta < 1.25^{3}$\\
			\hline 
			SfMLearner \cite{zhou2017unsupervised}\textdagger & M & 416 $\times$ 128 & CS+K & 0.176 & 1.532 & 6.129 & 0.244 & 0.758 & 0.921 & 0.971\\
			Vid2Depth \cite{mahjourian2018unsupervised} & M & 416 $\times$ 128 & CS+K & 0.134 & 0.983 & 5.501 & 0.203 & 0.827 & 0.944 & 0.981\\
			GeoNet \cite{yin2018geonet} & M  & 416 $\times$ 128 & CS+K & 0.132 & 0.994 & 5.240 & 0.193 & 0.833 & 0.953 & 0.985\\
			DDVO \cite{wang2018learning} & M & 416 $\times$ 128 & CS+K & 0.126 & 0.866 & 4.932 & 0.185 & 0.851 & 0.958 & 0.986\\ 
			CC \cite{ranjan2019competitive}  & M & 832  $\times$ 256  & K  & 0.123 & 0.881 & 4.834 & 0.181 & 0.860 & 0.959 & 0.985\\
			EPC++ \cite{luo2019every} & M & 640 $\times$ 192 & K & 0.120 & 0.789 & 4.755 & 0.177 & 0.856 & 0.961 & 0.987\\
			Monodepth2 \cite{godard2019digging} & M & 640 $\times$ 192   & K  &  0.090  &    0.545  &   3.942  &   0.137  &  0.914  &   0.983  &  0.995 \\
			Johnston \etal~\cite{johnston2020self} & M & 640 $\times$ 192   & K  &  0.081 &   0.484  &   3.716  &   0.126  & \underline {0.927}  &   0.985  & \underline {0.996} \\
			\rowcolor{gray!30} \textbf{CADepth-Net (Ours)} & M & 640 $\times$ 192 & K & \underline{0.080} & \underline {0.442} & \underline{3.639} & \underline {0.124} & \underline {0.927} & \underline{0.986} & \underline{0.996} \\ 
			\rowcolor{gray!30} \textbf{CADepth-Net (Ours)} & M & 1280 $\times$ 384 & K & \bf{0.076} & \bf{0.374} & \bf{3.280} & \bf{0.115} & \bf{0.937} & \bf{0.990} & \bf{0.997} \\  
			
			\hline
			
			Zhan FullNYU \cite{zhan2018unsupervised} & D*MS & 608 $\times$ 160 & K & 0.130 & 1.520 & 5.184 & 0.205 & 0.859 & 0.955 & 0.981 \\ 
			EPC++ \cite{luo2019every} & MS & 640 $\times$ 192 & K & 0.123 & 0.754 & 4.453 & 0.172 & 0.863 & 0.964 & 0.989 \\
			Monodepth2 \cite{godard2019digging} & MS & 640 $\times$ 192  & K & 0.080 & 0.466 & 3.681 & 0.127 & 0.926 & 0.985 & 0.995 \\ 
			\rowcolor{gray!30} \textbf{CADepth-Net (Ours)} & MS & 640 $\times$ 192 & K & \underline {0.076} & \underline{0.417}  & \underline{3.488} & \underline {0.120} & \underline{0.933}  & \underline{0.987} & \underline{0.996}  \\ 
			
			\rowcolor{gray!30} \textbf {CADepth-Net (Ours)} & MS & 1024 $\times$ 320 & K & \bf{0.070} & \bf{0.346} & \bf{3.168} & \bf{0.109} & \bf{0.945}   & \bf{0.991} & \bf {0.997} \\ 
			\hline
			
		\end{tabular}
	}
	\vspace{1pt}
	\caption{\textbf{Quantitative results on the KITTI improved ground truth.} Comparison of the existing methods to our CADepth-Net on KITTI 2015 \cite{geiger2012we} using annotated depth maps from \cite{uhrig2017sparsity}. Best results are in \textbf{bold}, with second best \underline{underlined}. For Abs Rel, Sq Rel, RMSE and  RMSE$_{log}$ lower is better, and for $\delta < 1.25$, $\delta < 1.25^2$ and $\delta < 1.25^3$ higher is better. In the \emph{Train} column, S: Self-supervised stereo supervision, M: Self-supervised mono supervision, D*: Auxiliary depth supervision. \textdagger~ refers to the newer results from github. In \emph{Dataset} column, CS: Cityscapes dataset~\cite{cordts2016cityscapes}, K: KITTI datasets~\cite{geiger2012we}.}
	\label{tab:kitti_improved}
\end{table*}

\begin{figure}[t]
	\begin{center}
		\includegraphics[width=1\linewidth]{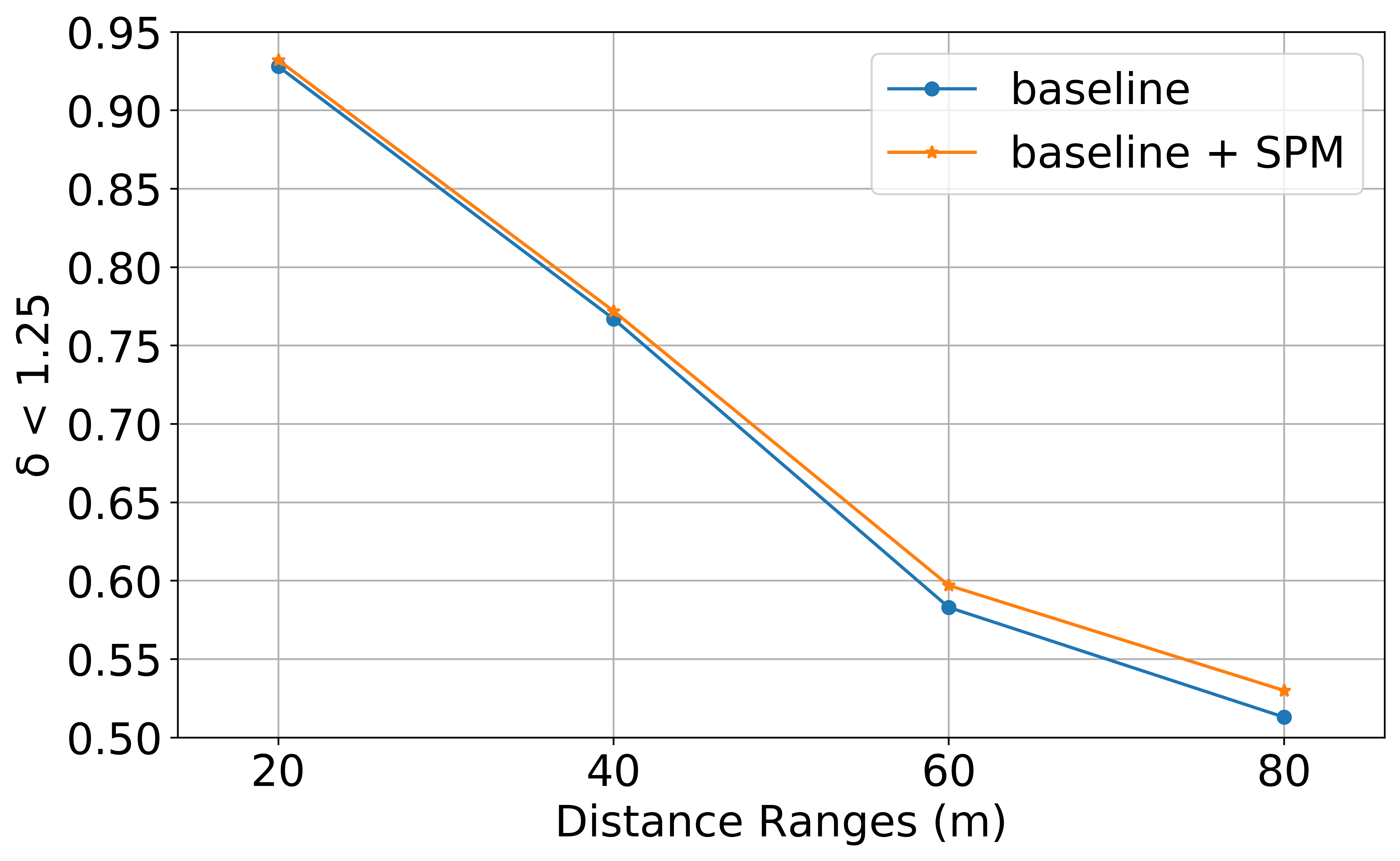}
	\end{center}
	\caption{\textbf{Depth Evaluation on KITTI binned at different intervals}, calculated independently by only considering ground-truth depth pixels in that range (0-20m, 20-40m, ...).}
	\label{fig:depth_range}
\end{figure}

\section{KITTI Evaluation Server Benchmark}

In Table \ref{tab:kitti_eval_server} we report the results of our models on the KITTI single image depth prediction benchmark~\cite{kittidepthserver} which were computed on the KITTI online evaluation server. We train a new model on the new split consisting of 72,084 training examples, 6,060 validation, and 500 test with the same training protocols mentioned in the main paper. As we cannot use median scaling for evaluation, we calculate the scale factor on the 2,000 KITTI training samples which have ground truth depths available. Table \ref{tab:kitti_eval_server} shows that our CADepth-Net outperforms the existing self-supervised approaches and significantly reduces the performance gap to the supervised methods.

\begin{table}
	\centering
	\resizebox{\linewidth}{!}{
		\begin{tabular}{| c | c || c | c | c | c | c}
			\hline
			Method & Train & \cellcolor{col1} SILog & \cellcolor{col1} sqErrorRel & \cellcolor{col1} absErrorRel  & \cellcolor{col1} iRMSE \\ 
			\hline
			DHGRL~\cite{zhang2018deep}& D & 15.47 & 4.04  & 12.52  & 15.72 \\ 
			CSWS~\cite{li2018monocular}& D & 14.85 & 3.48  & 11.84  & 16.38 \\
			APMoE~\cite{kong2018pixel} & D & 14.74 & 3.88  & 11.74  & 15.63 \\
			DABC~\cite{li2018deep}& D & 14.49 & 4.08  & 12.72  & 15.53 \\
			DORN~\cite{fu2018deep} & D & 11.77 & 2.23  & 8.78  & 12.98 \\
			\hline
			Monodepth~\cite{godard2017unsupervised} & S & 22.02	& 20.58 & 	17.79 &	21.84\\
			LSIM~\cite{goldman2019learn}  & S & 17.92	& 6.88 & 14.04 & 17.62 \\
			Monodepth2~\cite{godard2019digging} & M & 15.57 & 4.52 & 12.98 & 16.70 \\
			SGDepth~\cite{klingner2020self} & M & 15.30 & 5.00 & 13.29 & 15.80 \\
			Monodepth2~\cite{godard2019digging} & MS & 15.07 & 4.16 & 11.64 & 15.27 \\
			\rowcolor{gray!30} \textbf{Ours} & MS & \bf{13.34} & \bf{3.33} & \bf{10.67} & \bf{13.61} \\		
			\hline
		\end{tabular}
	}
	\vspace{1pt}
	\caption{\textbf{Results on KITTI depth prediction benchmark.} D refers to ground truth depth supervision, while M and S are monocular and stereo self-supervision respectively.}
	\label{tab:kitti_eval_server}
\end{table}

\begin{table}
	\centering
	\resizebox{\columnwidth}{!}
	{
		\footnotesize
		\begin{tabular}{|l|c|c|c|c|c|}
			\hline
			& Backbone  & 
			\cellcolor{col1}Abs Rel & \cellcolor{col1}Sq Rel & \cellcolor{col1}RMSE & 
			\cellcolor{col2}$\delta < $1.25 \\
			
			\hline
			SGDepth~\cite{klingner2020self} & R18  &  0.117 & 0.907 & 4.844 & 0.875 \\
			monodepth2~\cite{godard2019digging} & R18  & 0.115 & 0.903 & 4.863 & 0.877 \\
			
			\textbf{Ours} & R18 & \bf{0.110}& \bf{0.812} & \bf{4.686} & \bf{0.882} \\					
			
			\hline
			
			monodepth2~\cite{godard2019digging} & R50  & 0.110 & 0.831 & 4.642 & 0.883 \\
			Johnston \etal~\cite{johnston2020self} & R101 & 0.106 & 0.861 & 4.699 & 0.889  \\ 
			\textbf{Ours} & R50 & \bf{0.105} & \bf{0.769} & \bf{4.535} & \bf{0.892}  \\
			

			
			\hline
		\end{tabular}
	}
	\vspace{1pt}
	\caption{
		\textbf{Comparisons of methods with the same backnone on the KITTI Eigen Split.} 
		R: ResNet. All models are trained with the same settings.}
	\label{tab:backbones}
\end{table}

\section{Additional Ablation Experiments}

\begin{figure*}
	\centering
	\includegraphics[width=\linewidth]{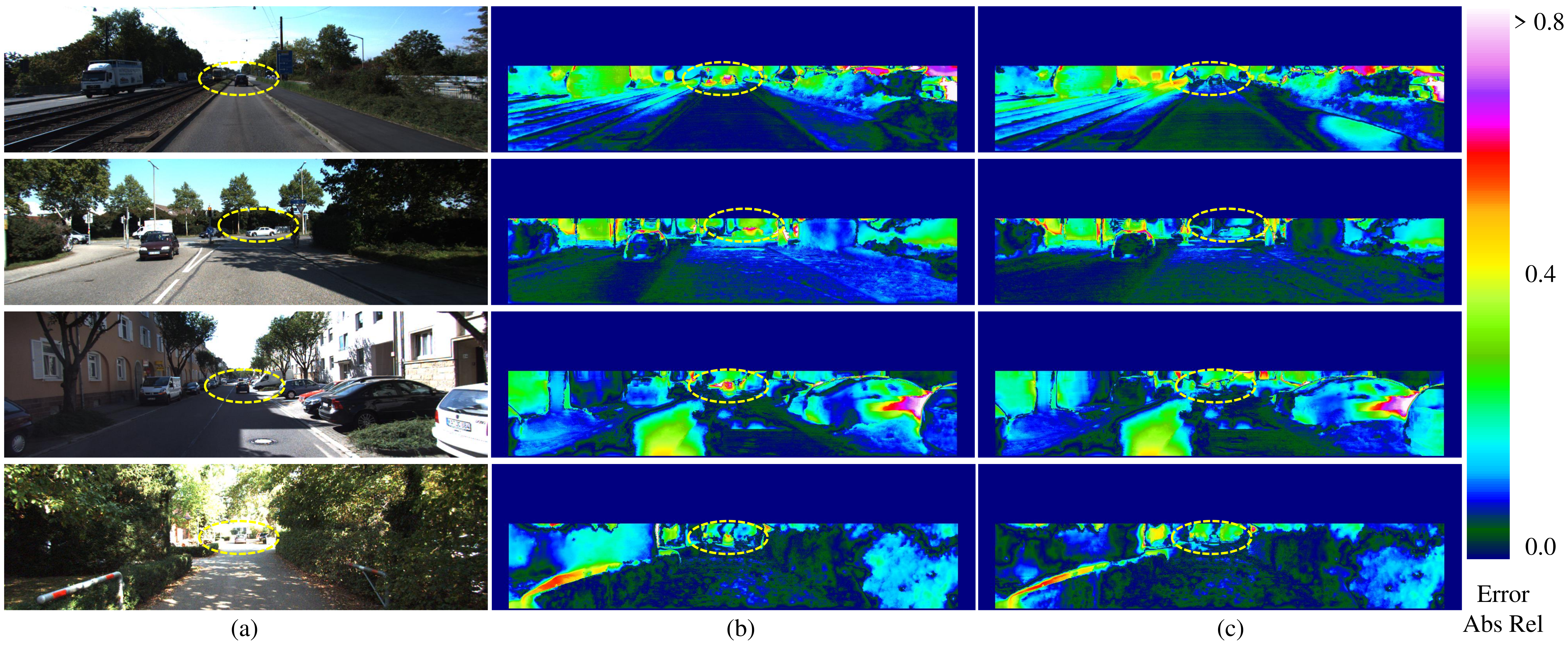} 
	\caption{\textbf{Qualitative ablation study of Structure Perception Module (SPM).} (a) Input Image. (b) Error maps without SPM. (c) Error maps with SPM. The error induced by distance objects (yellow circles) is improved by the structure perception module.
	}
	\label{fig:ablation_spm}
\end{figure*}

\begin{figure*}
	\centering
	\includegraphics[width=\linewidth]{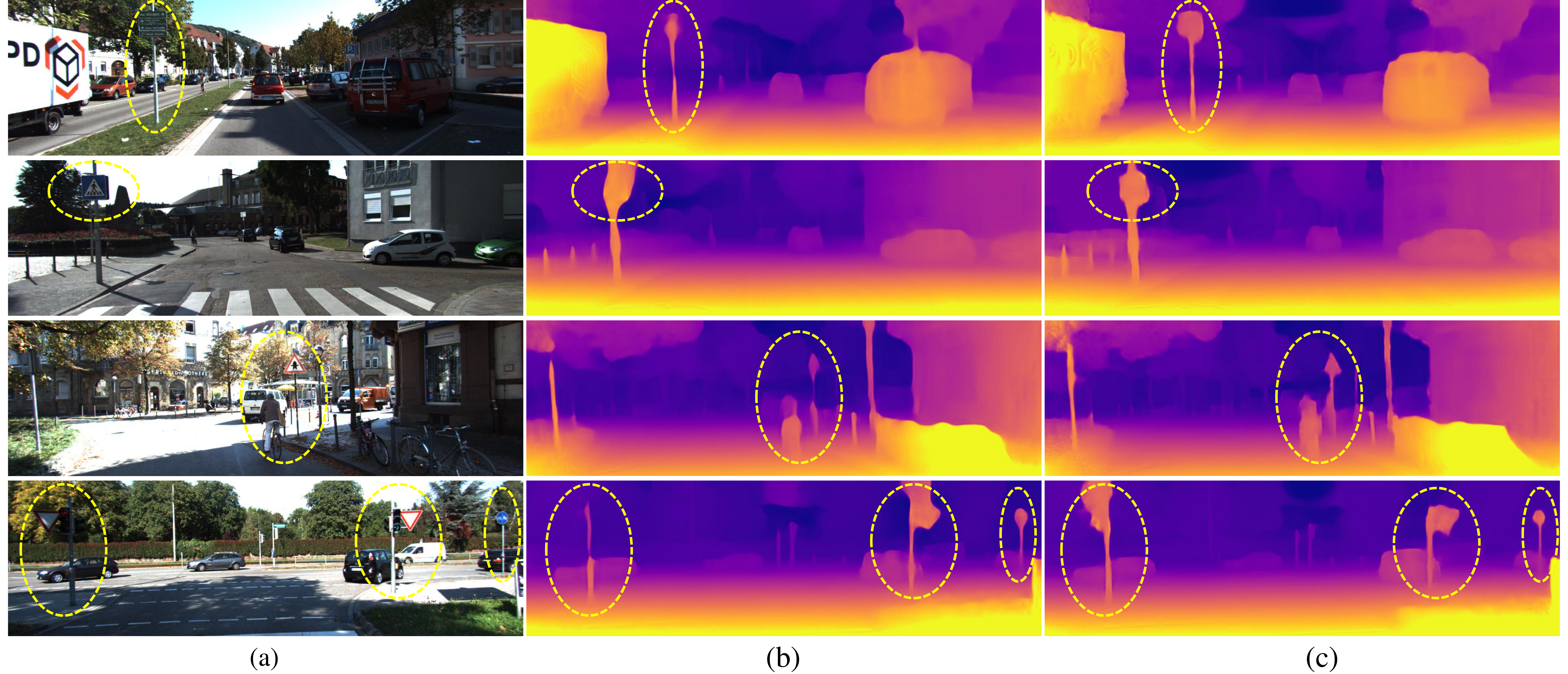} 
	\caption{\textbf{Qualitative ablation study of Detail Emphasis Module (DEM).} (a) Input image. (b) Predicted depth maps without DEM. (c) Predicted depth maps with DEM. Thanks to the detail emphasis module, we could obtain the more precise and sharper depth estimation.
	}
	\label{fig:ablation_dem}
\end{figure*}

To further demonstrate the effectiveness of our structure perception module, we evaluate the performance of distant objects with the absolute relative error, Fig. \ref{fig:ablation_spm} shows that we can effectively reduce the error produced by distant objects. Besides, Fig. \ref{fig:depth_range} reports that our model improves the accuracy at all depth intervals, and the performance gap consistently increases when larger distances are considered, thanks to the better 3D scene geometric perception introduced by the structure perception module. Fig. \ref{fig:ablation_dem} shows that our method generates more fine-gained details and accurate object boundaries \eg pedestrians and thin road signs by using detail emphasis module individually. For a fair comparison, Table  \ref{tab:backbones} reports that our model outperforms other methods with the same backbone, which means that the gain in performance shown in our experiments is mainly due to the effectiveness of our proposed methods.

\begin{figure}
	\centering
	\resizebox{\columnwidth}{!}{
		\newcommand{\turnheightnew}{0.2\columnwidth}
\centering

\begin{tabular}{@{\hskip -1mm}c@{\hskip 0.5mm}c@{\hskip 0.5mm}c@{\hskip 0.5mm}c@{\hskip -1mm}}

\Large{Input}  & \Large{MD2 (M)}~\cite{godard2019digging}&  \Large {Ours (M)}  & \Large{Ground truth} \\

\includegraphics[height=\turnheightnew]{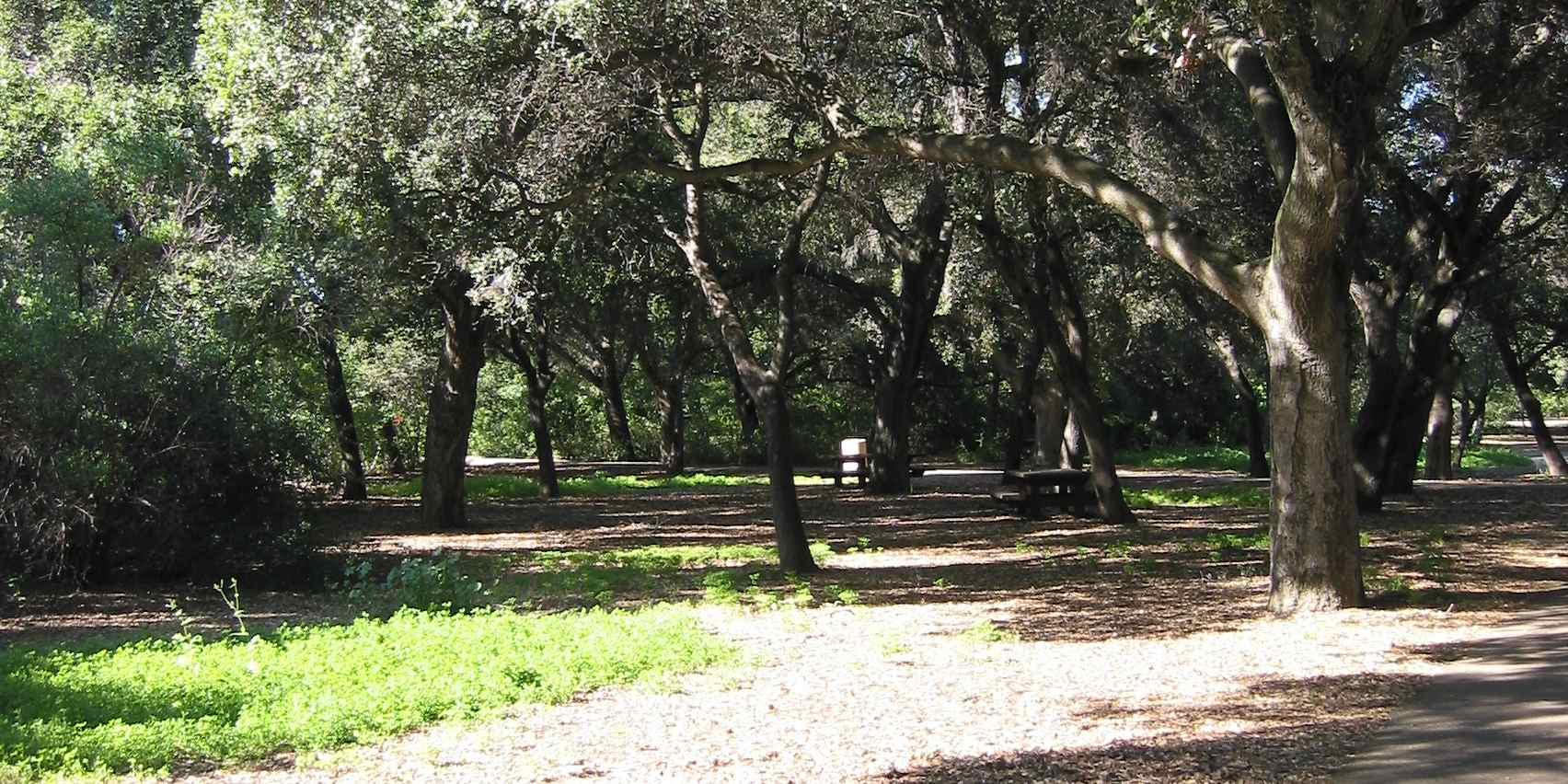} &
\includegraphics[height=\turnheightnew]{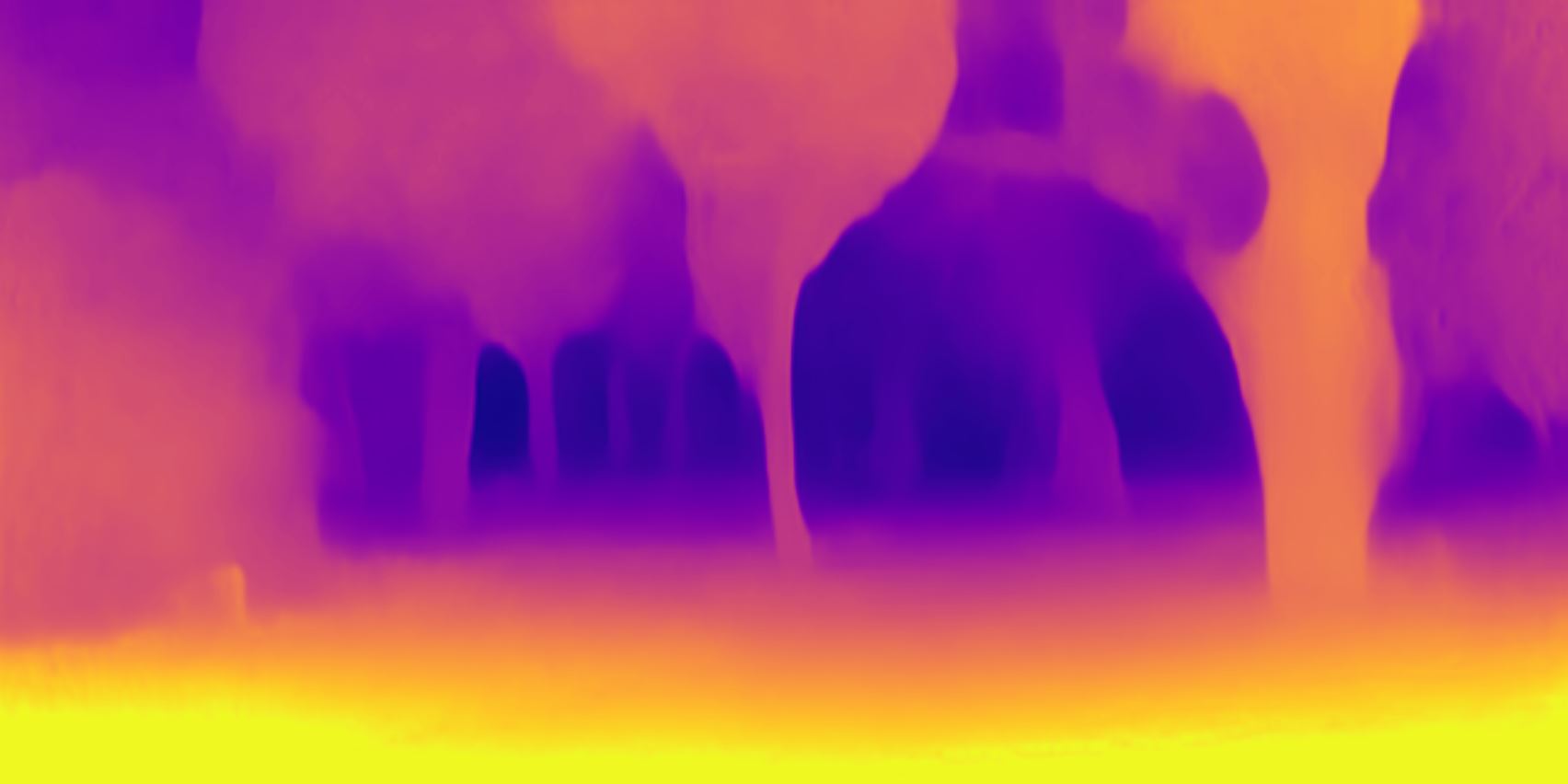} &
\includegraphics[height=\turnheightnew]{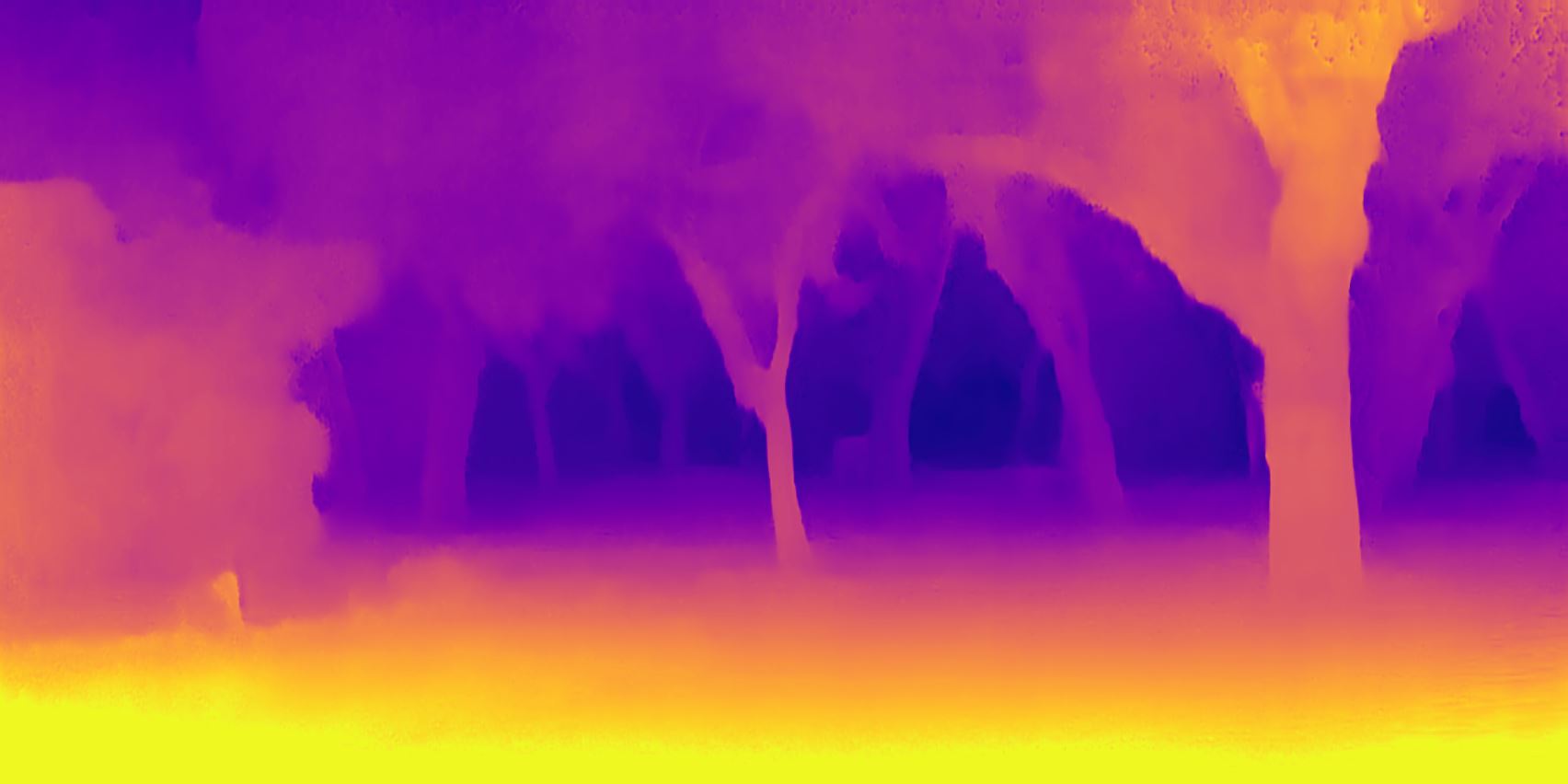} &
\includegraphics[height=\turnheightnew]{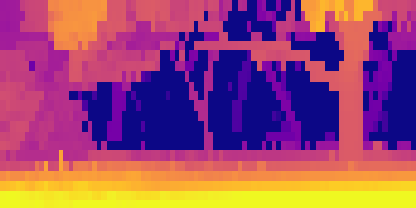}\\

\includegraphics[height=\turnheightnew]{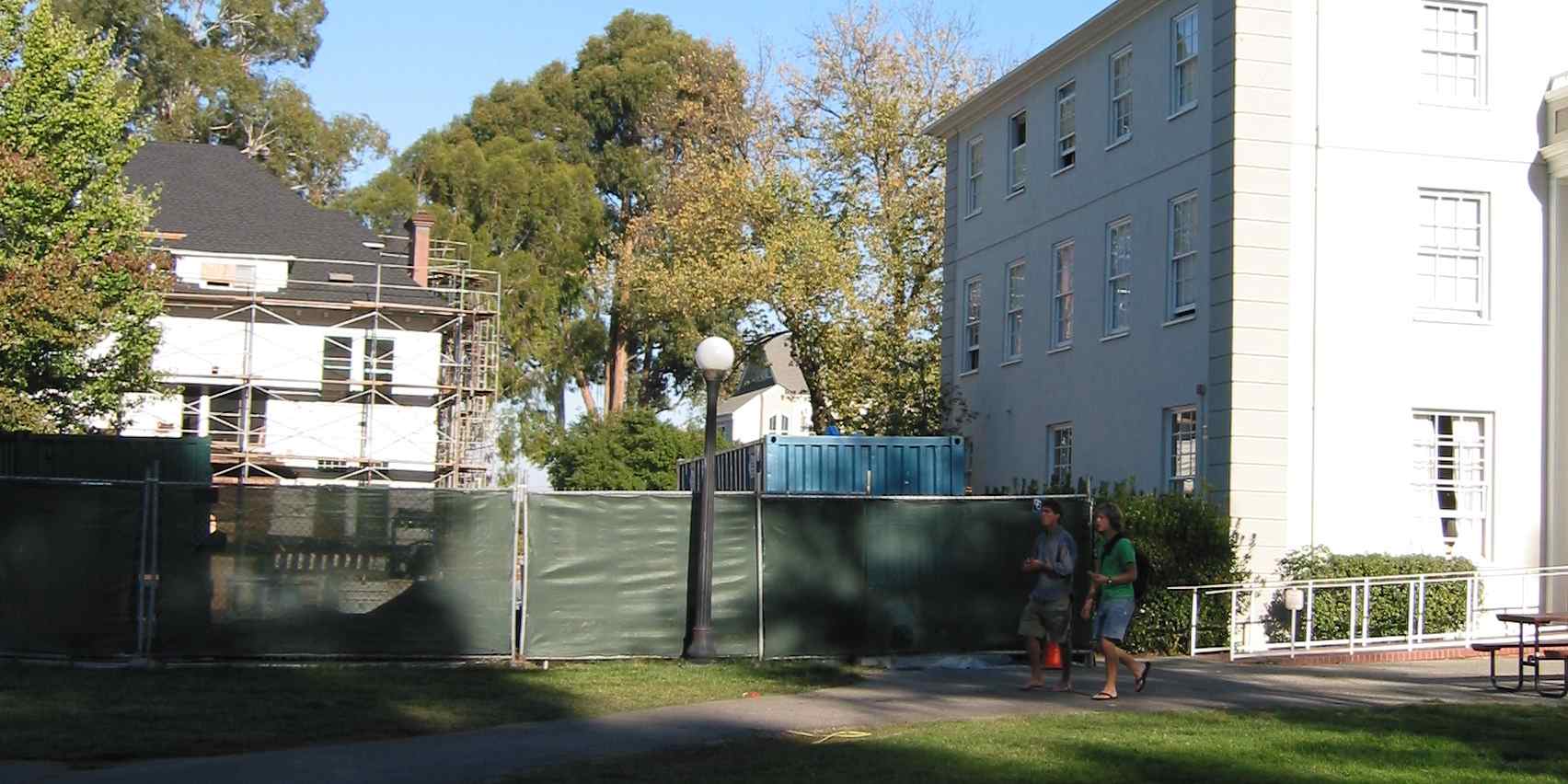} &
\includegraphics[height=\turnheightnew]{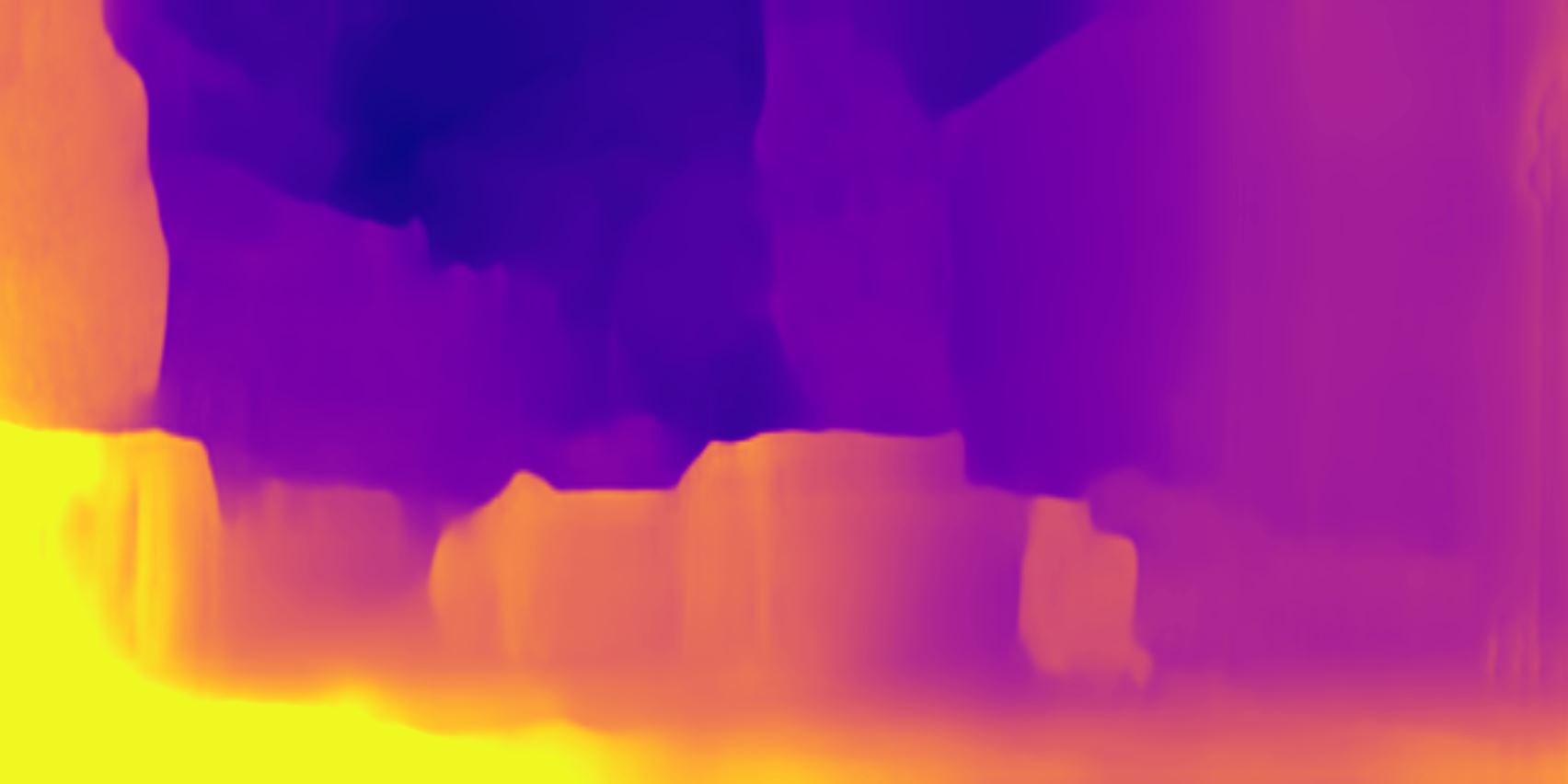} &
\includegraphics[height=\turnheightnew]{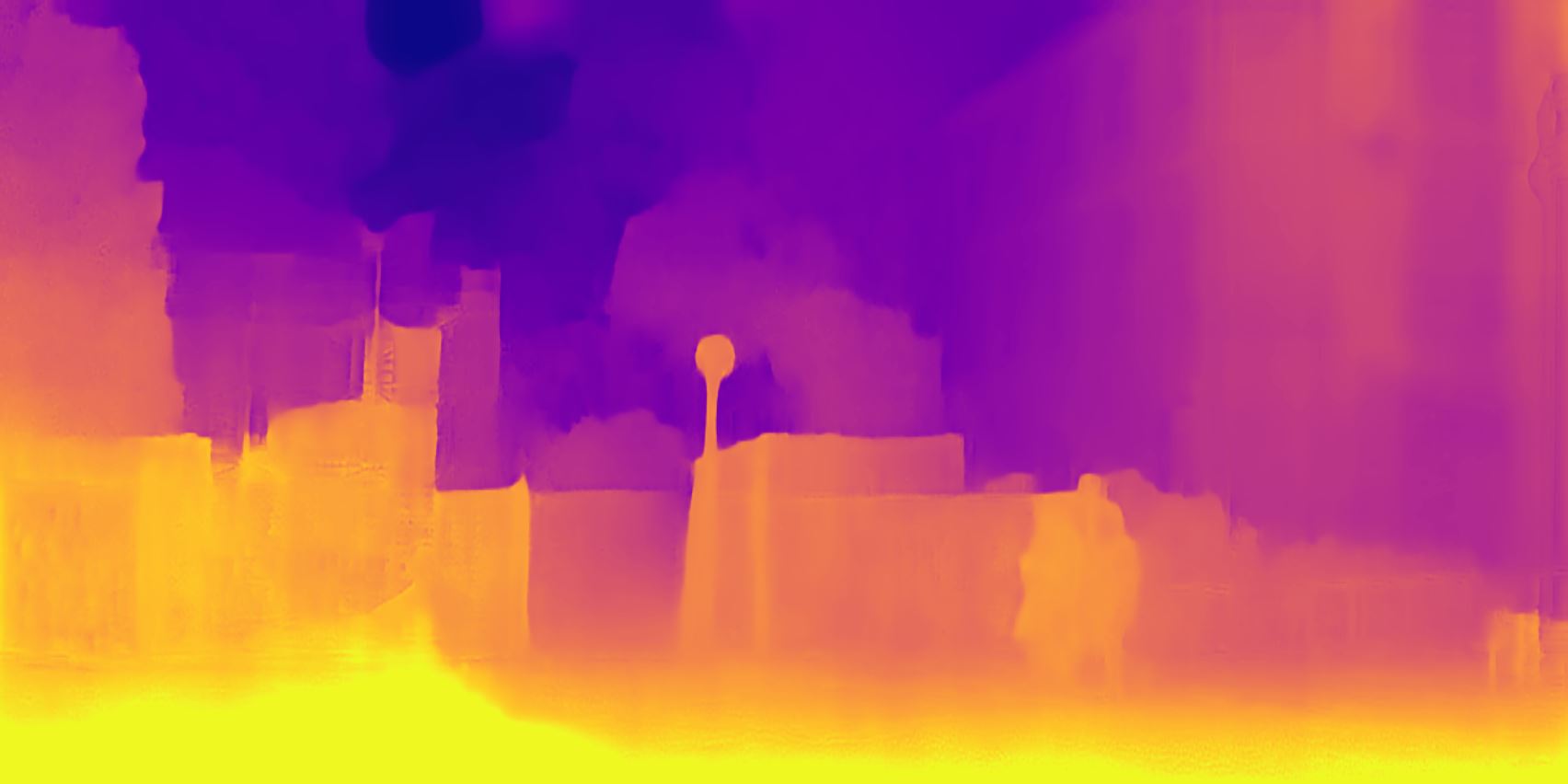} &
\includegraphics[height=\turnheightnew]{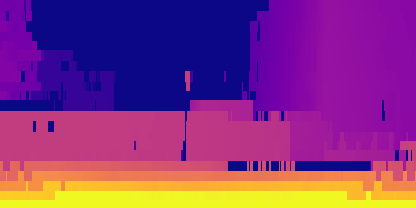}\\

\includegraphics[height=\turnheightnew]{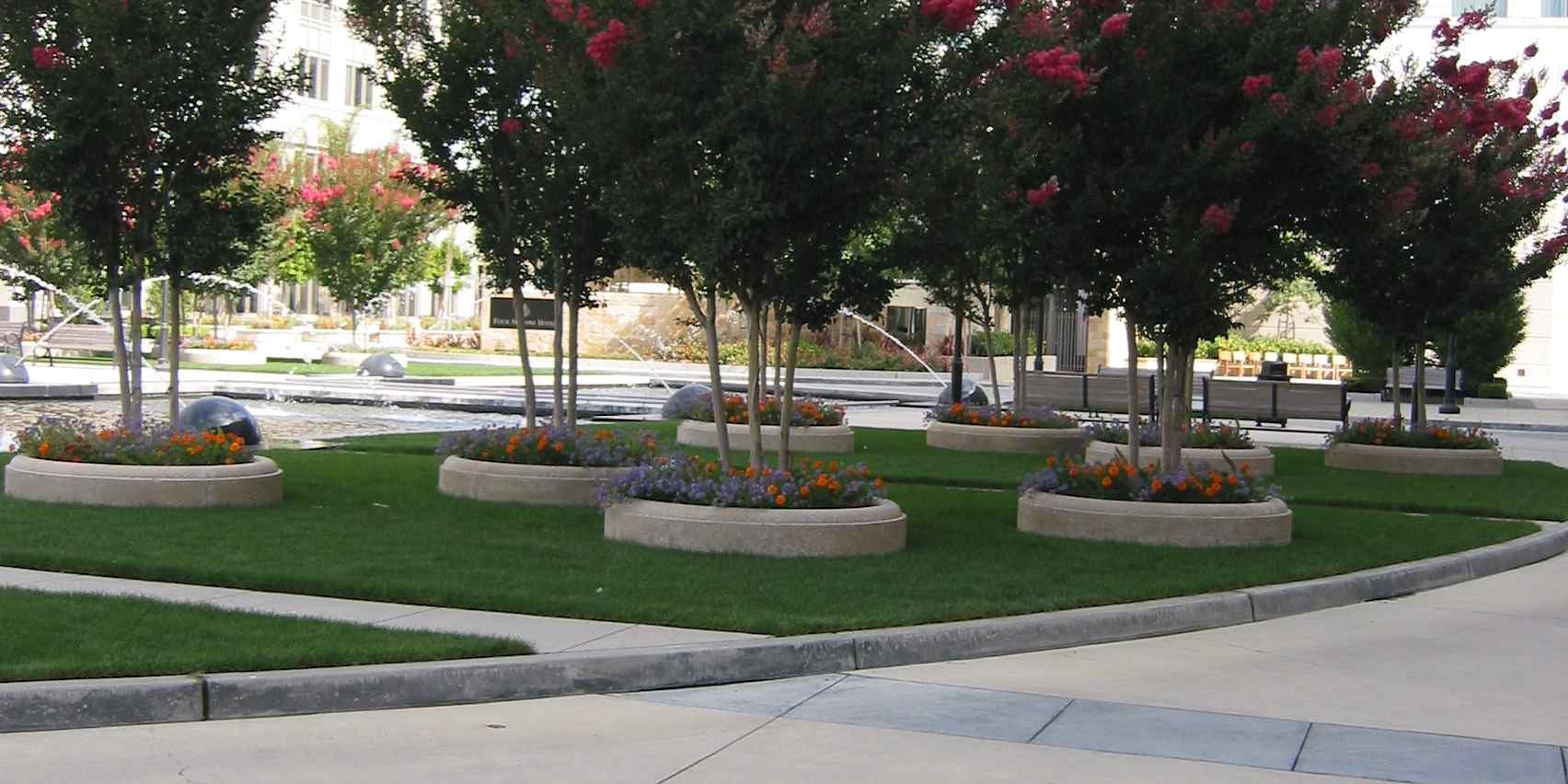} &
\includegraphics[height=\turnheightnew]{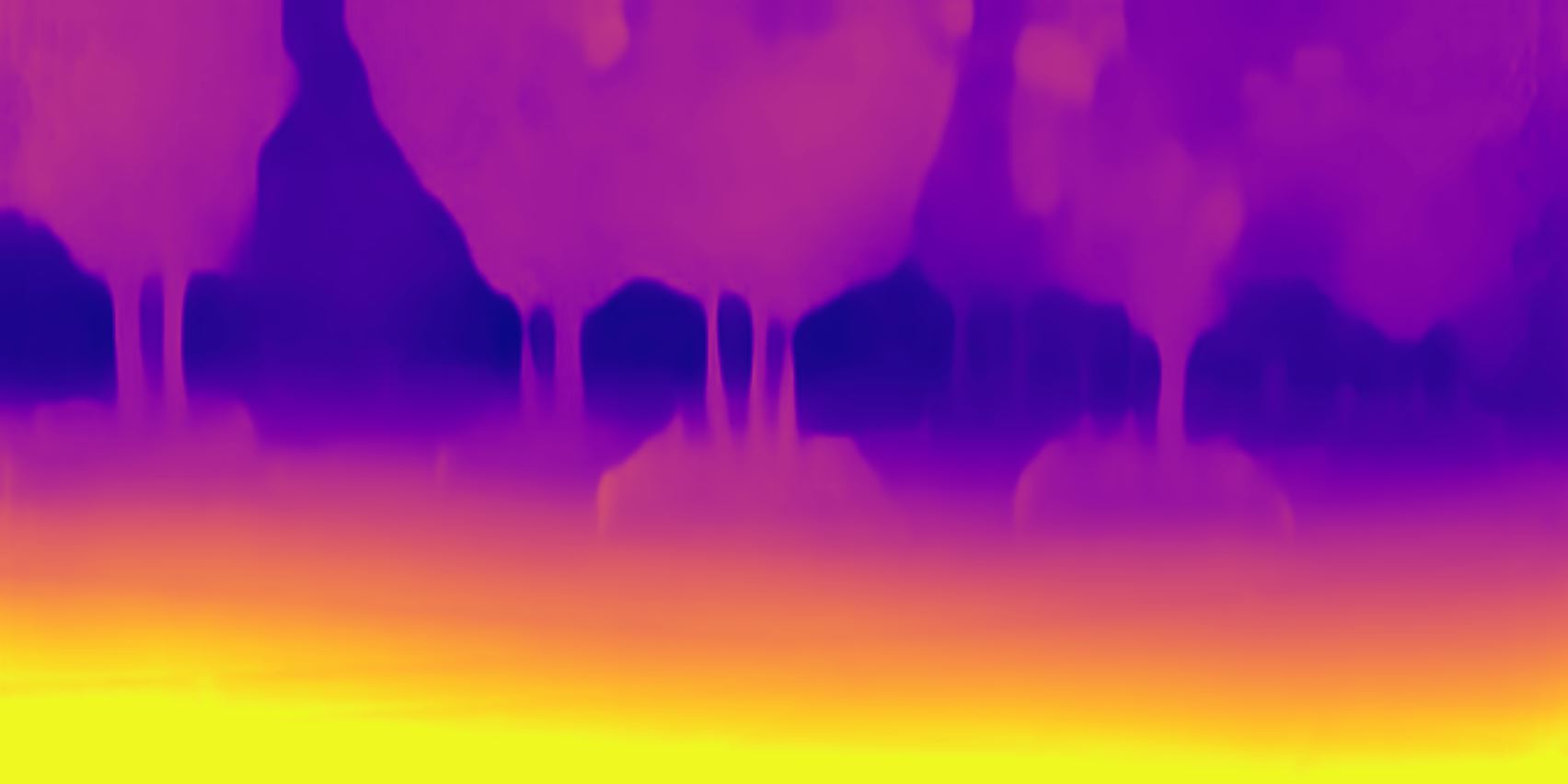} &
\includegraphics[height=\turnheightnew]{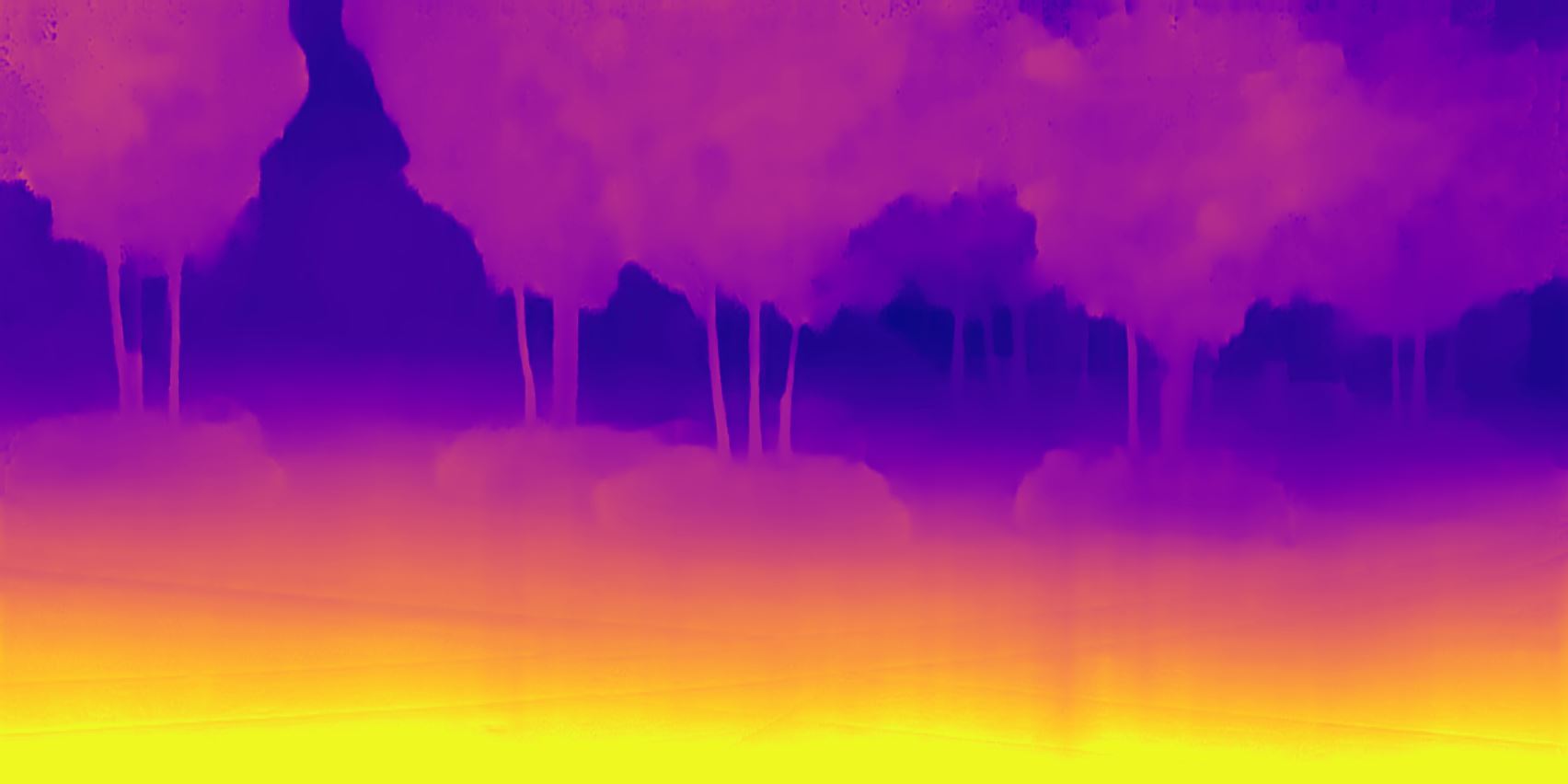} &
\includegraphics[height=\turnheightnew]{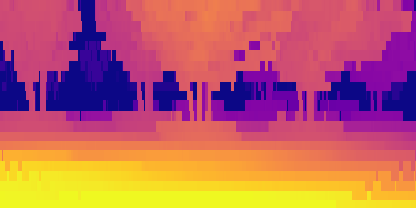}\\

\includegraphics[height=\turnheightnew]{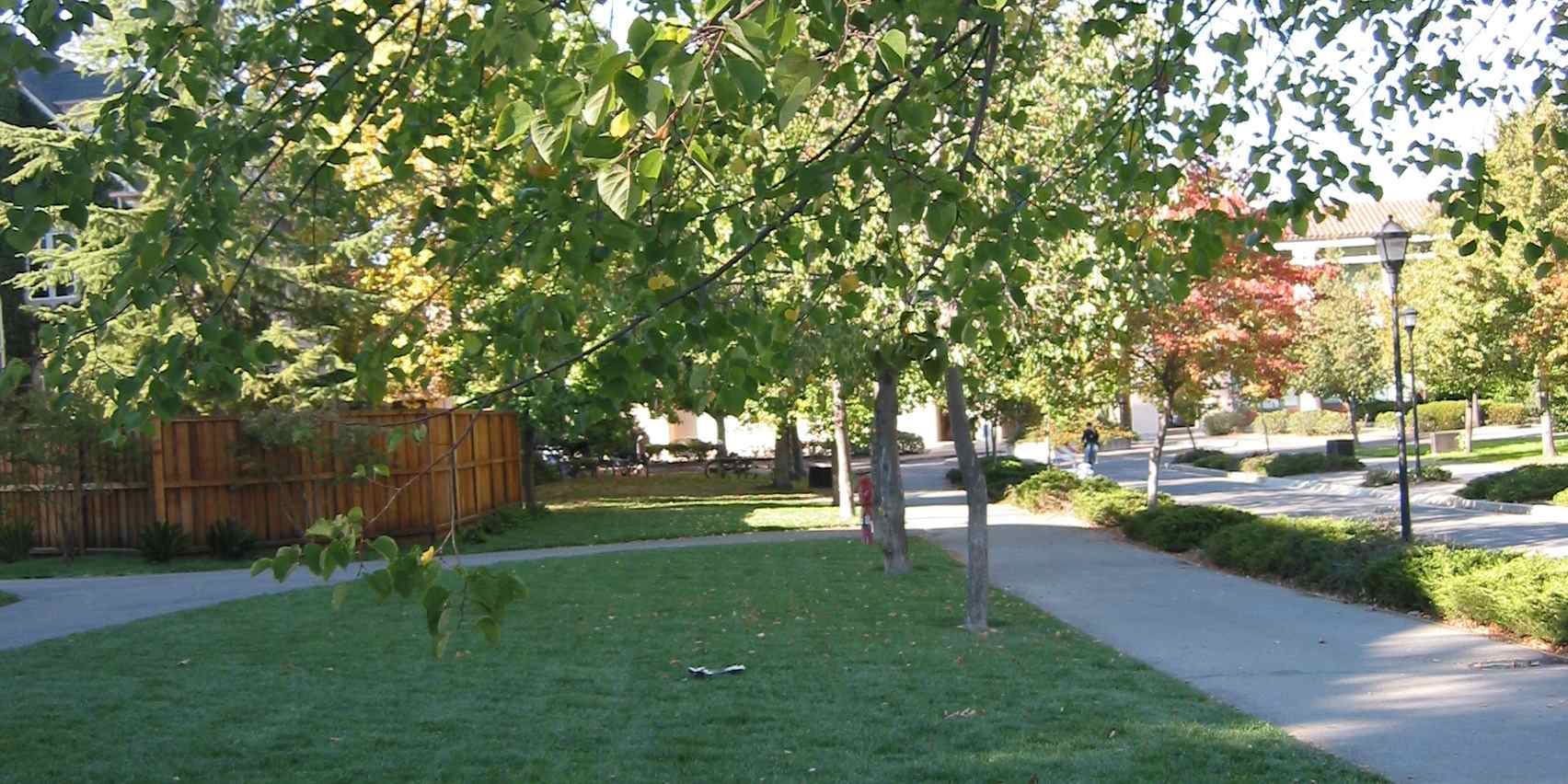} &
\includegraphics[height=\turnheightnew]{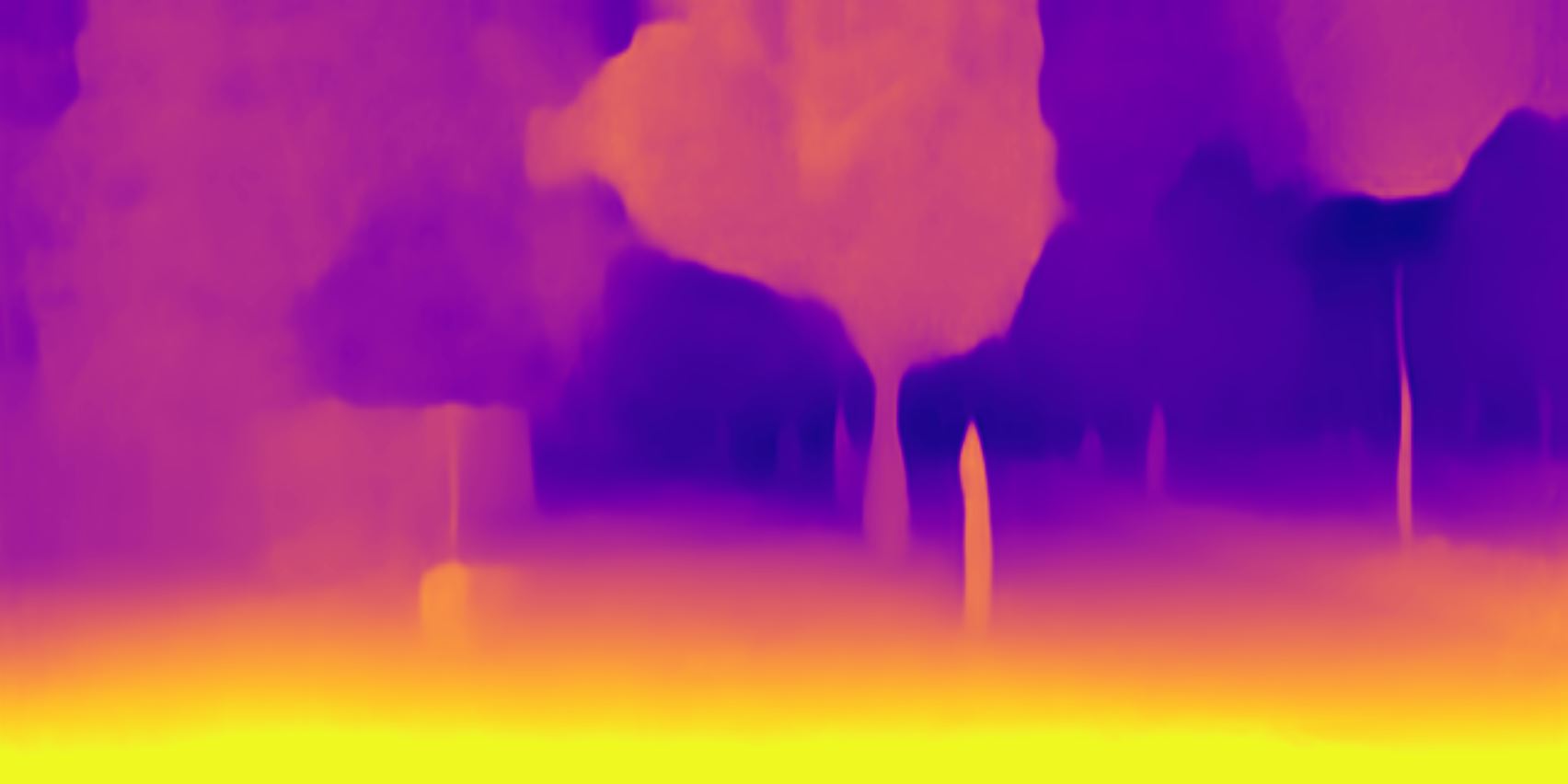} &
\includegraphics[height=\turnheightnew]{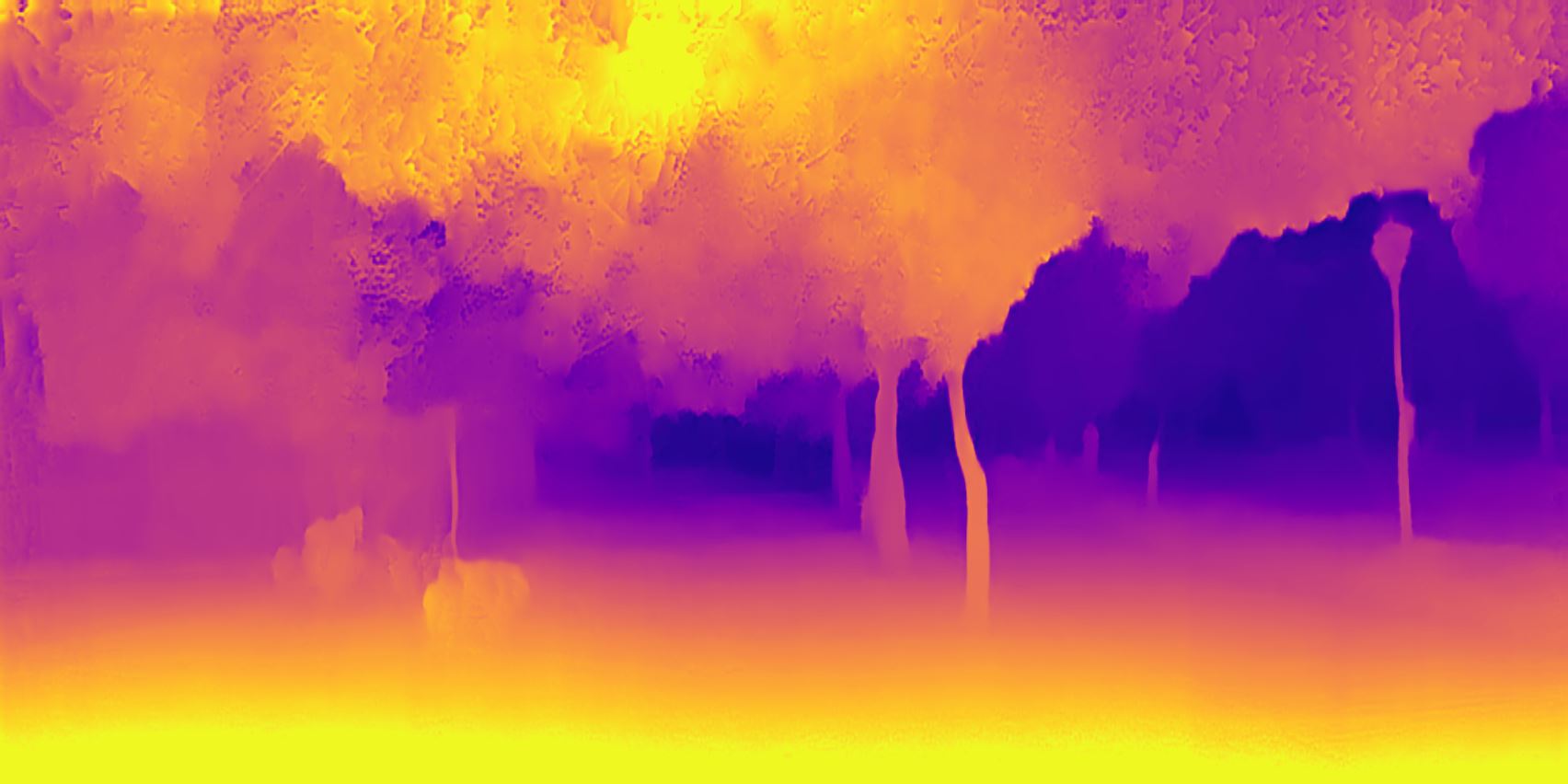} &
\includegraphics[height=\turnheightnew]{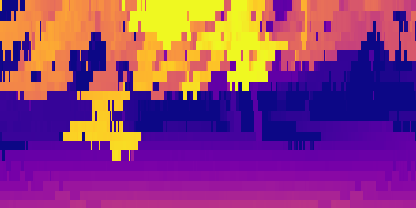}\\

\includegraphics[height=\turnheightnew]{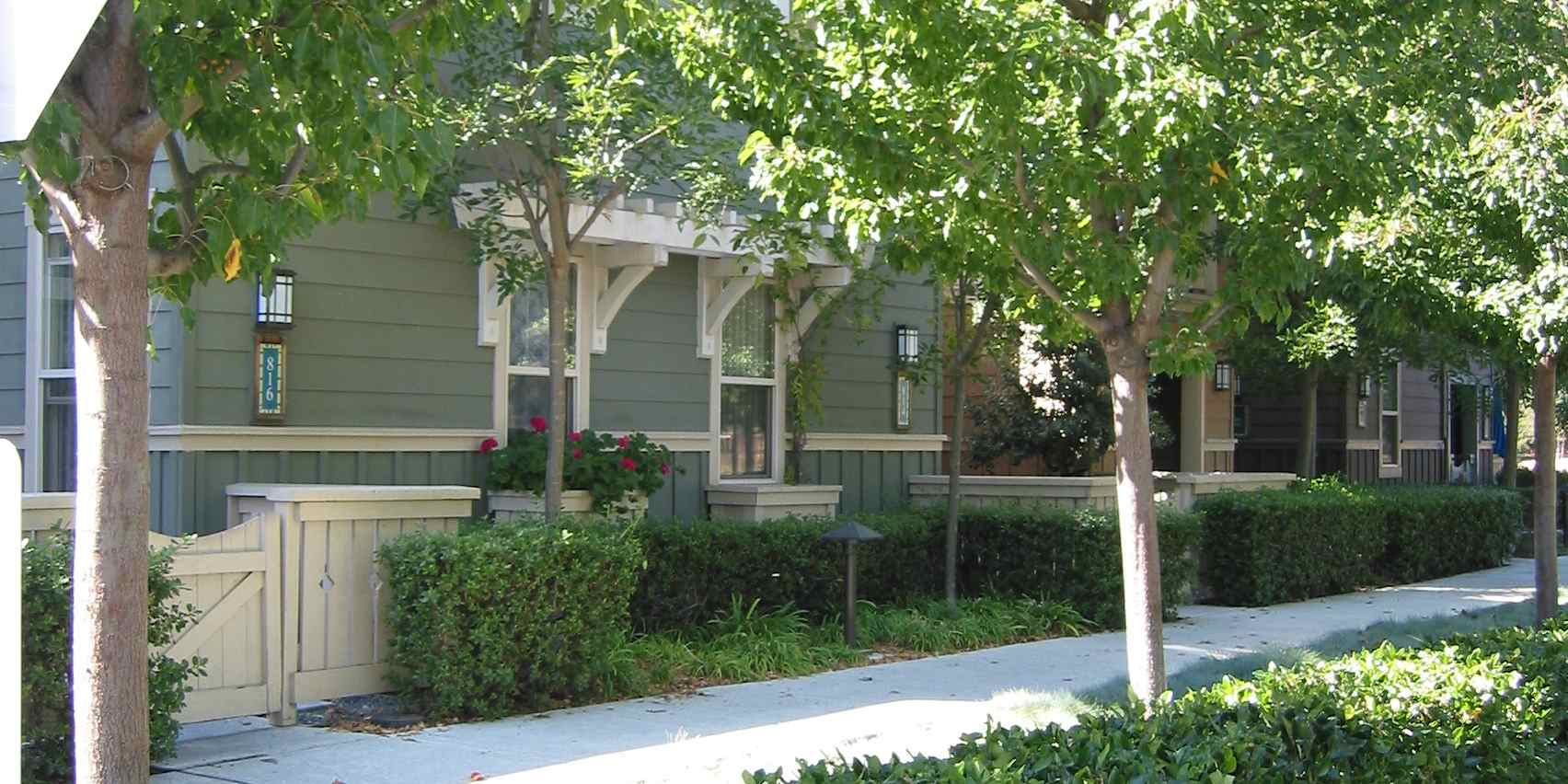} &
\includegraphics[height=\turnheightnew]{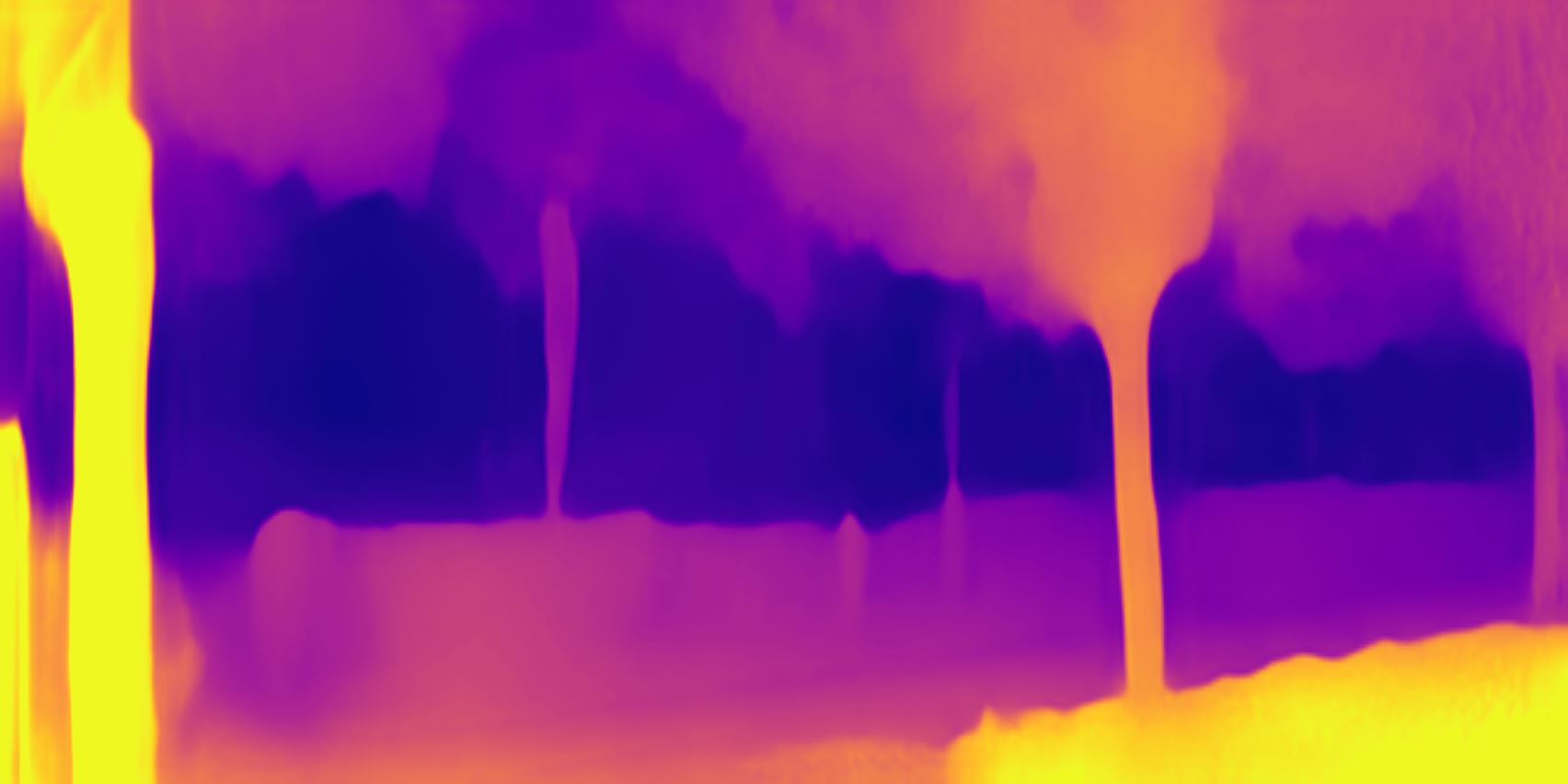} &
\includegraphics[height=\turnheightnew]{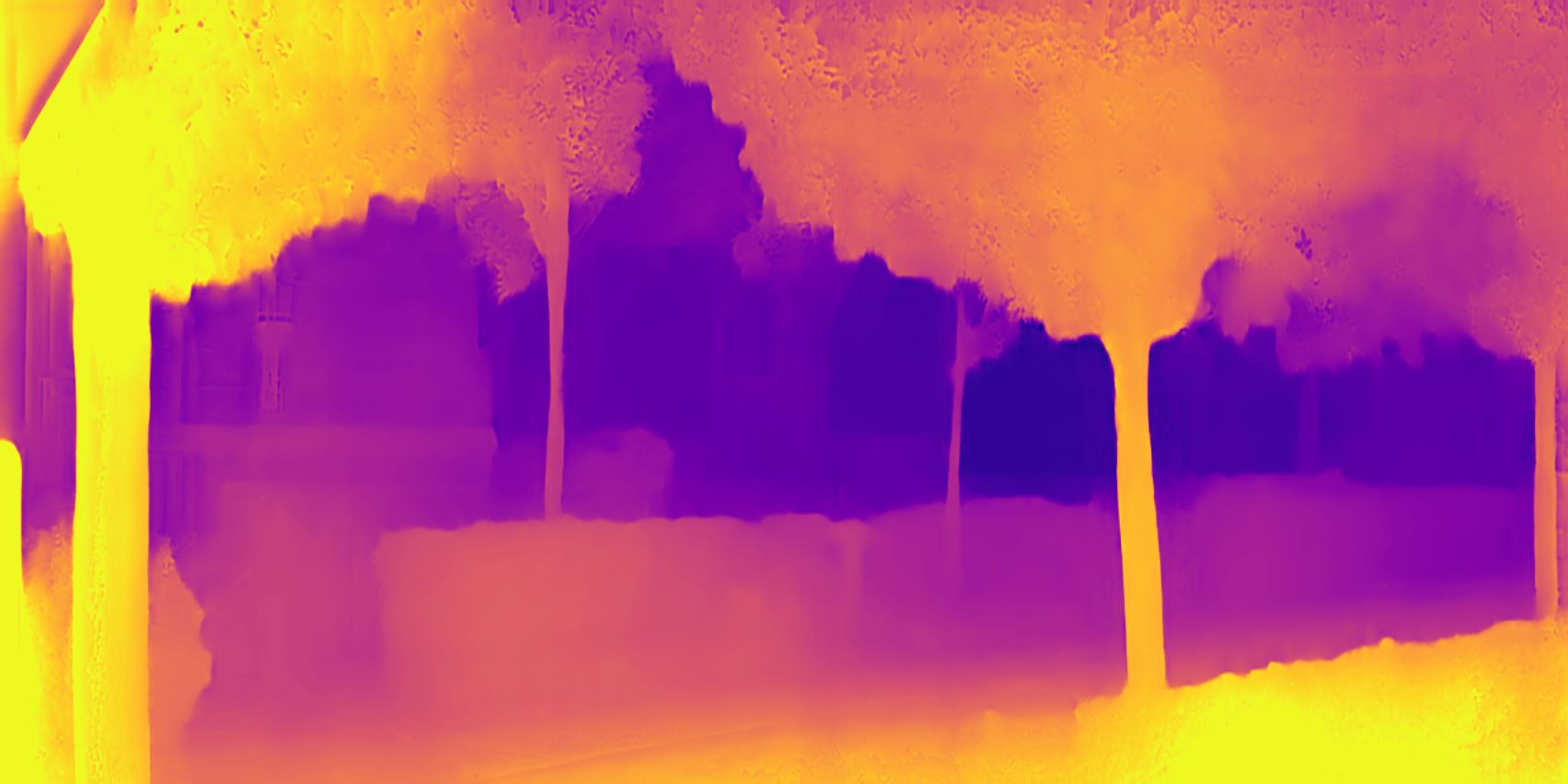} &
\includegraphics[height=\turnheightnew]{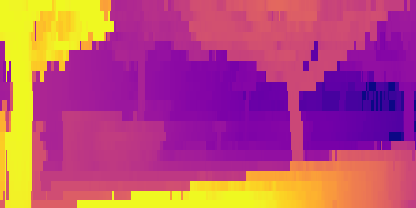}\\
\end{tabular}
 }
	\vspace{1pt}
	\caption{\textbf{Qualitative Make3D results.} Our CADepth-Net generates more fine-gained details compared to other method. }
	\label{fig:supp_make3d_results}
\end{figure}

\section{Additional Qualitative Comparisons}

\begin{figure*}[!ht]
	\centering
	\resizebox{\linewidth}{!}{
		\newcommand{\turnheightnew}{0.2\columnwidth}
\centering

\begin{tabular}{@{\hskip -0.5mm}c@{\hskip 0.5mm}c@{\hskip 0.5mm}c@{\hskip 0.5mm}c@{\hskip -0.5mm}}

\normalsize{Input}  & \normalsize{Monodepth2~\cite{godard2019digging}} &  \normalsize {PackNet~\cite{guizilini20203d}}  & \textbf{\normalsize{Our CADepth-Net}} \\

\includegraphics[height=\turnheightnew]{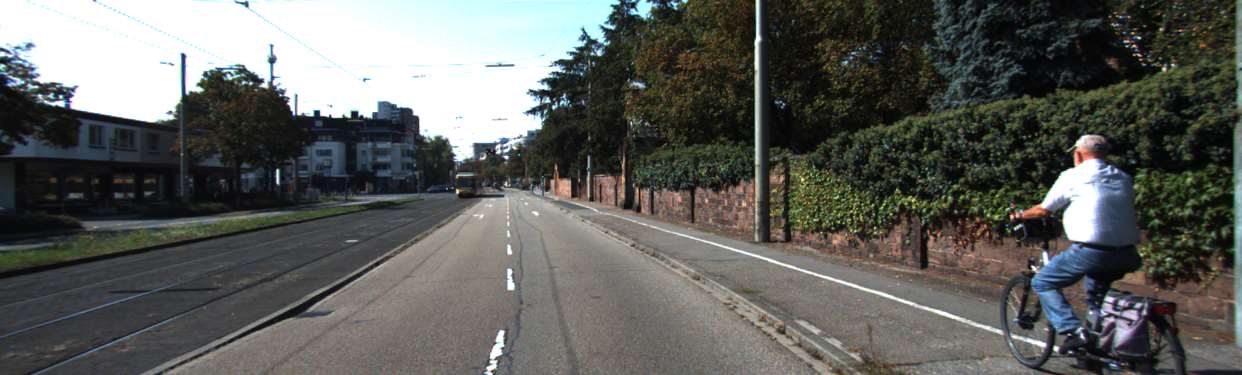} &
\includegraphics[height=\turnheightnew]{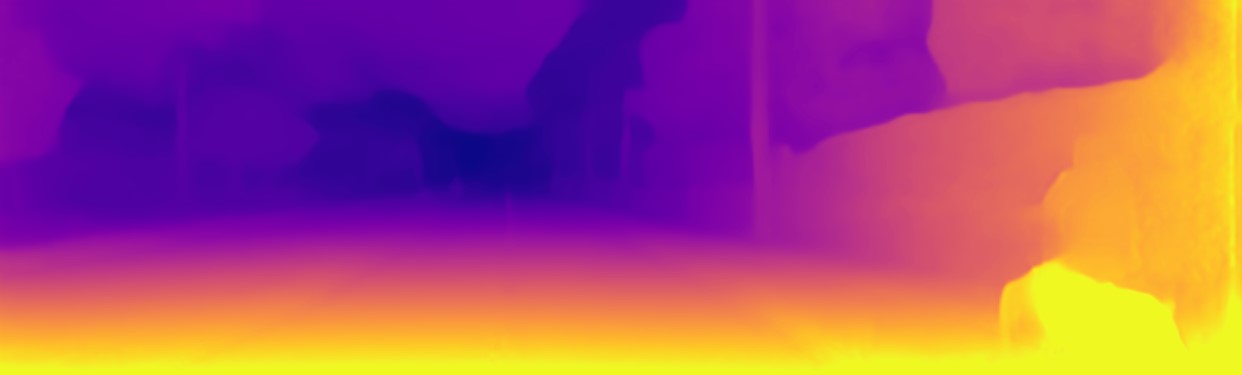} &
\includegraphics[height=\turnheightnew]{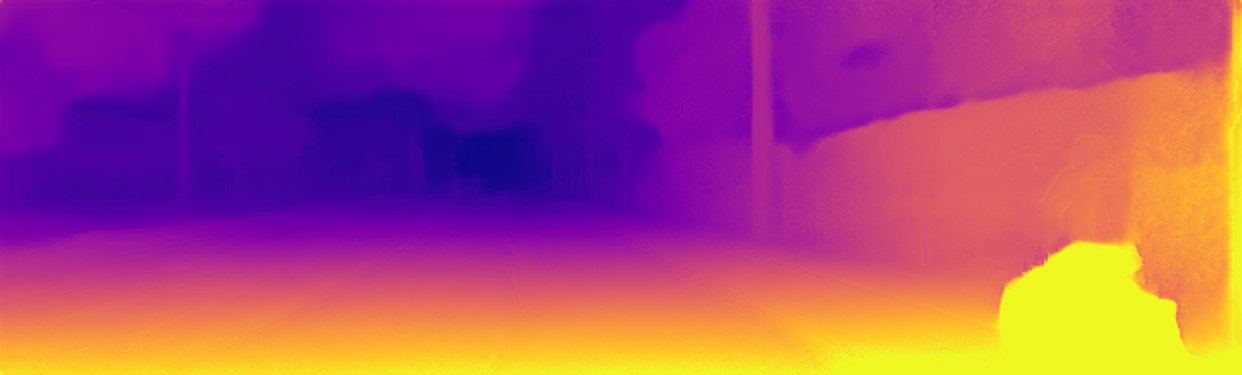} &
\includegraphics[height=\turnheightnew]{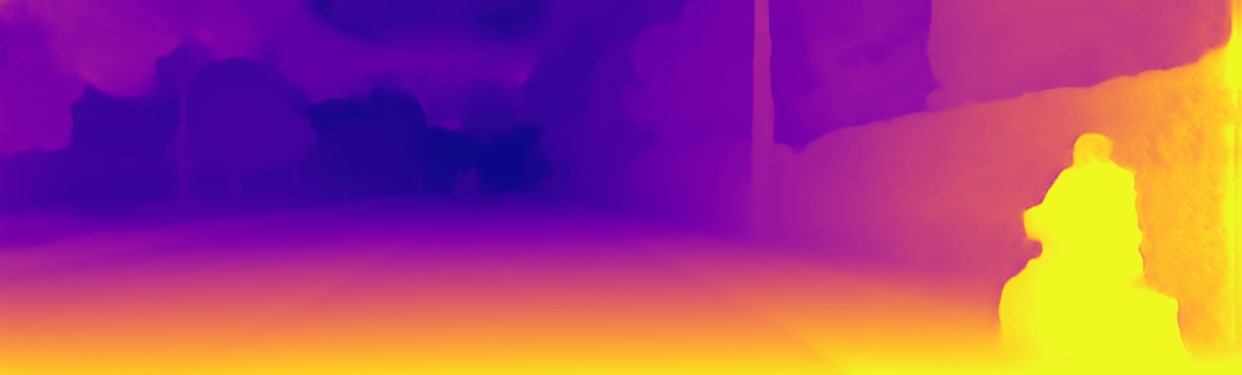}\\

\includegraphics[height=\turnheightnew]{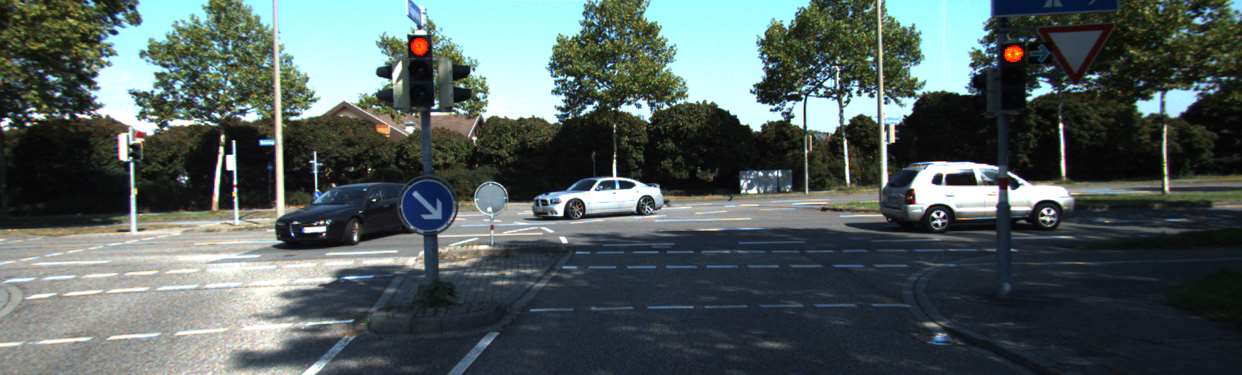} &
\includegraphics[height=\turnheightnew]{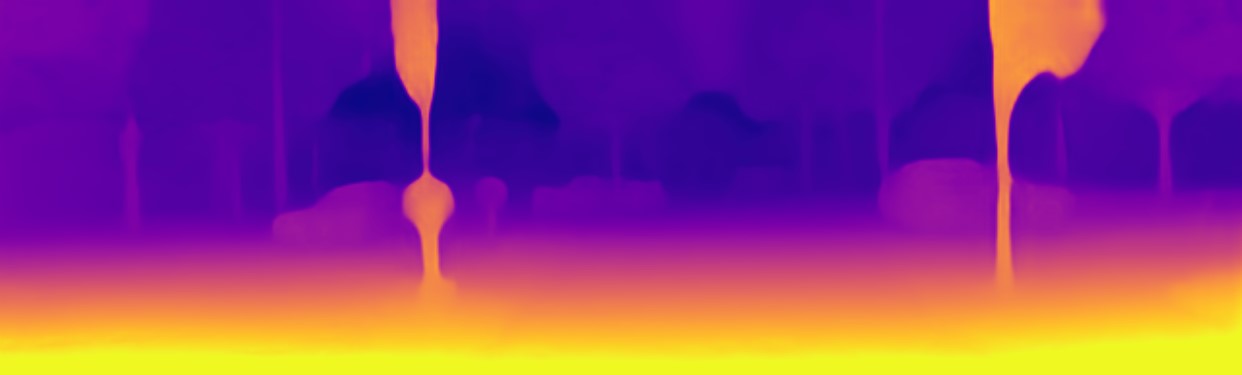} &
\includegraphics[height=\turnheightnew]{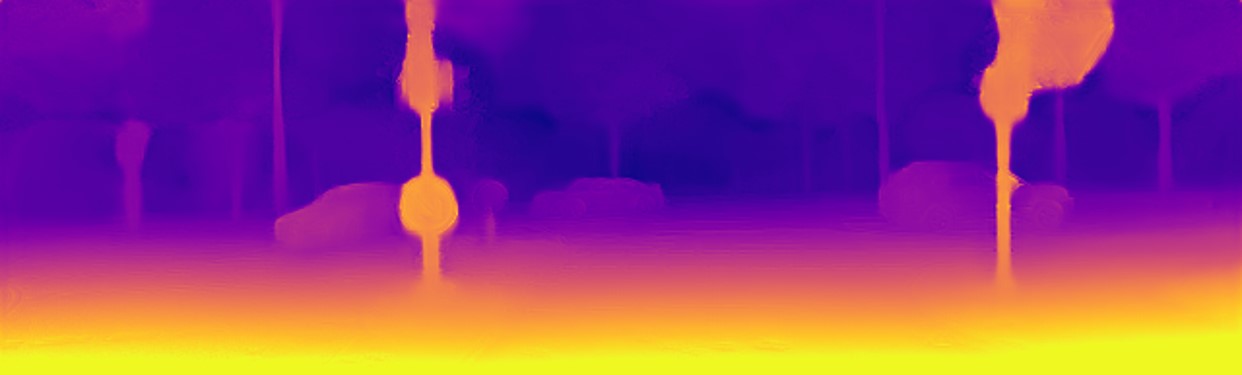} &
\includegraphics[height=\turnheightnew]{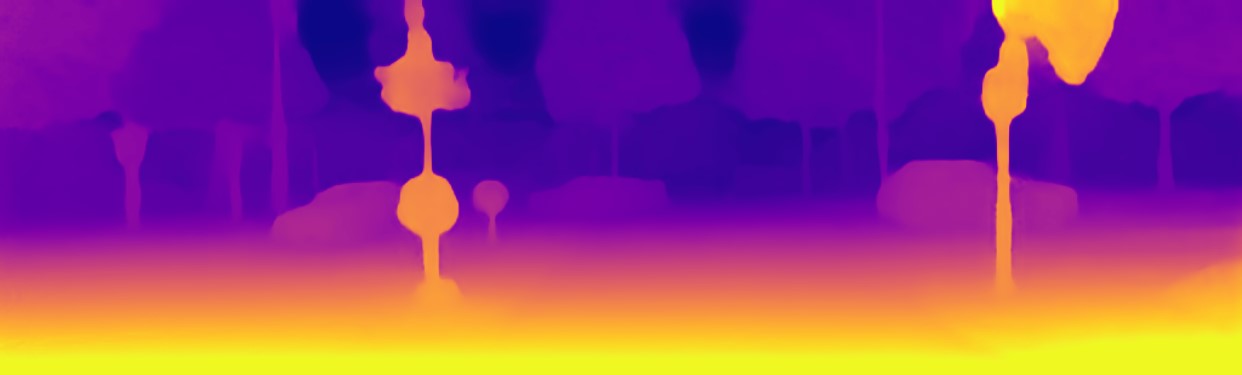}\\

\includegraphics[height=\turnheightnew]{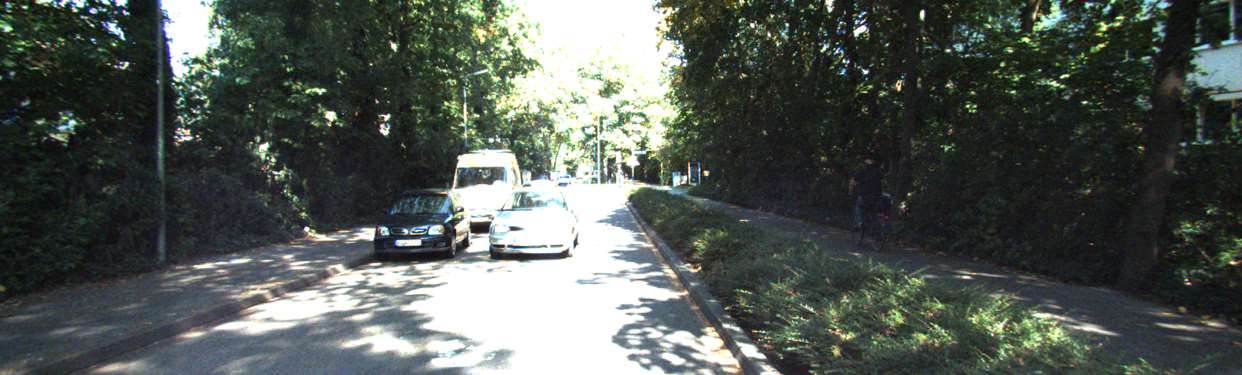} &
\includegraphics[height=\turnheightnew]{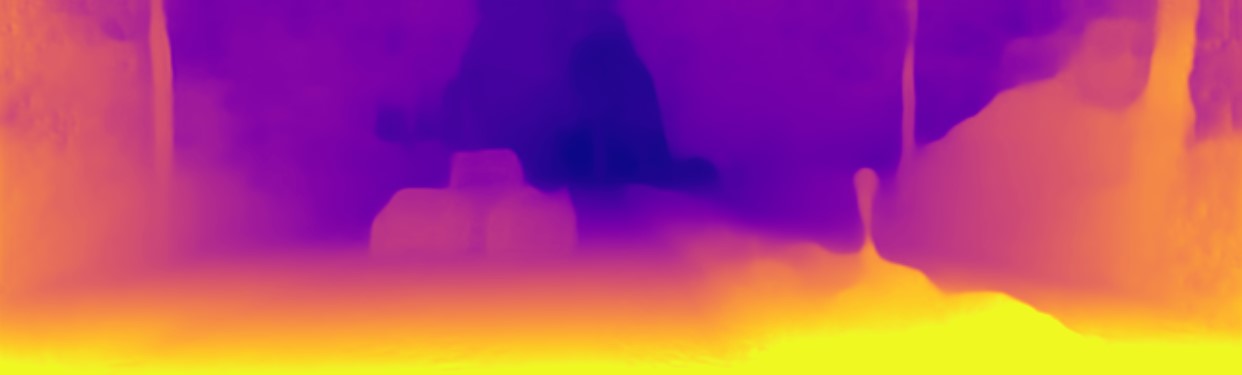} &
\includegraphics[height=\turnheightnew]{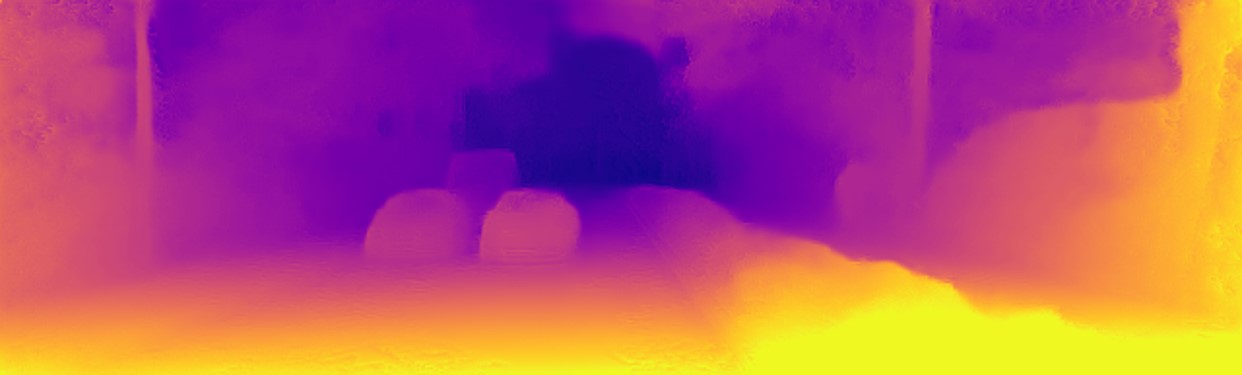} &
\includegraphics[height=\turnheightnew]{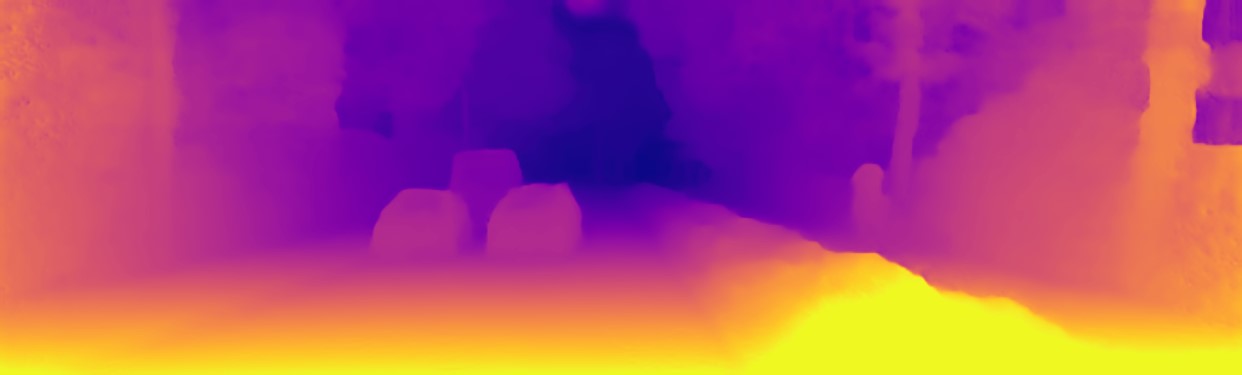}\\

\includegraphics[height=\turnheightnew]{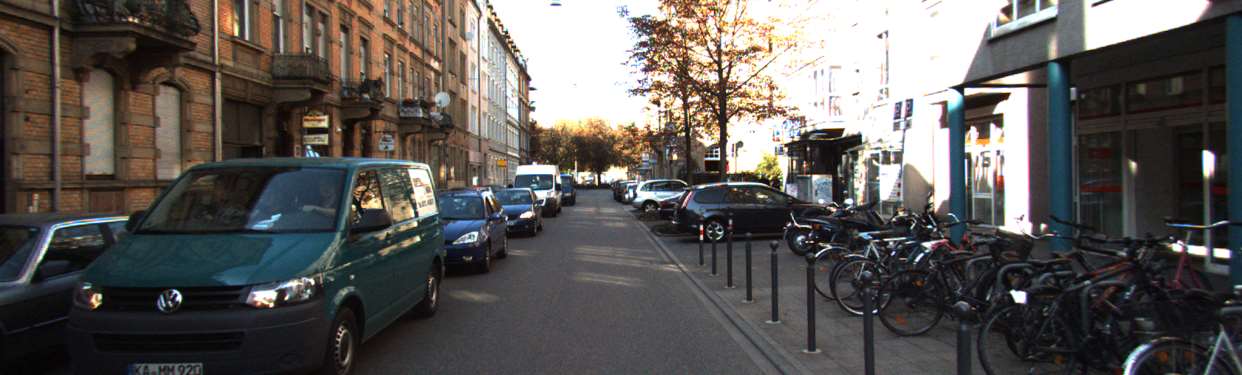} &
\includegraphics[height=\turnheightnew]{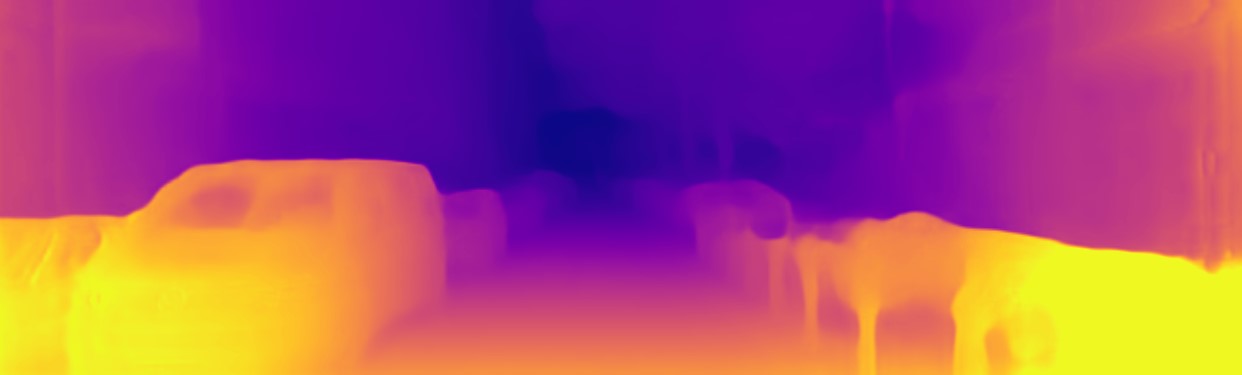} &
\includegraphics[height=\turnheightnew]{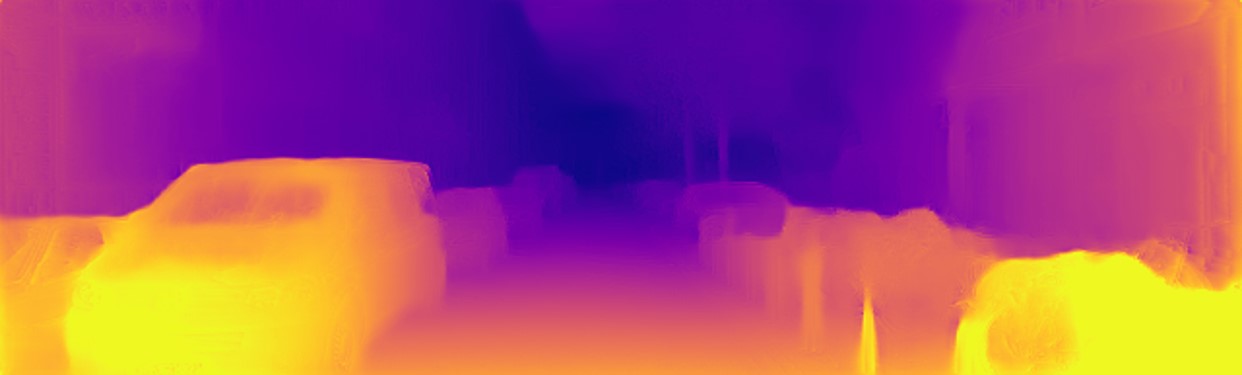} &
\includegraphics[height=\turnheightnew]{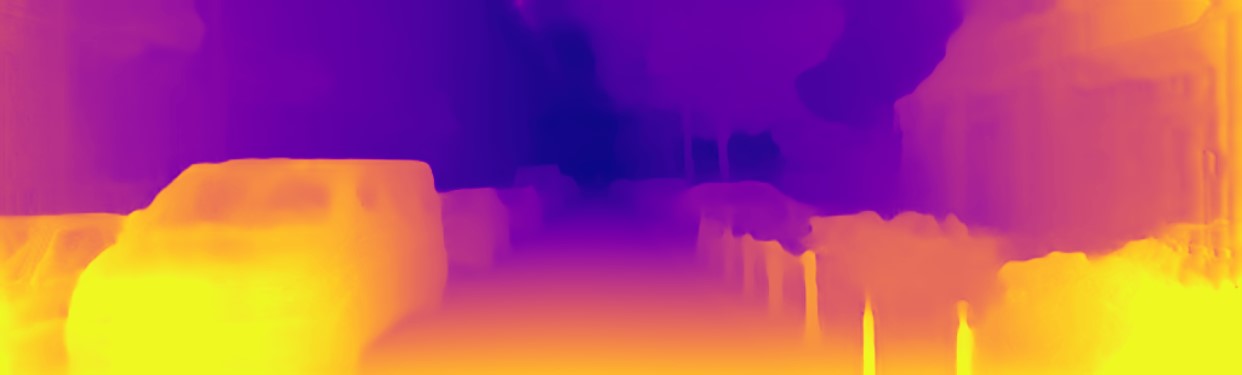}\\

\includegraphics[height=\turnheightnew]{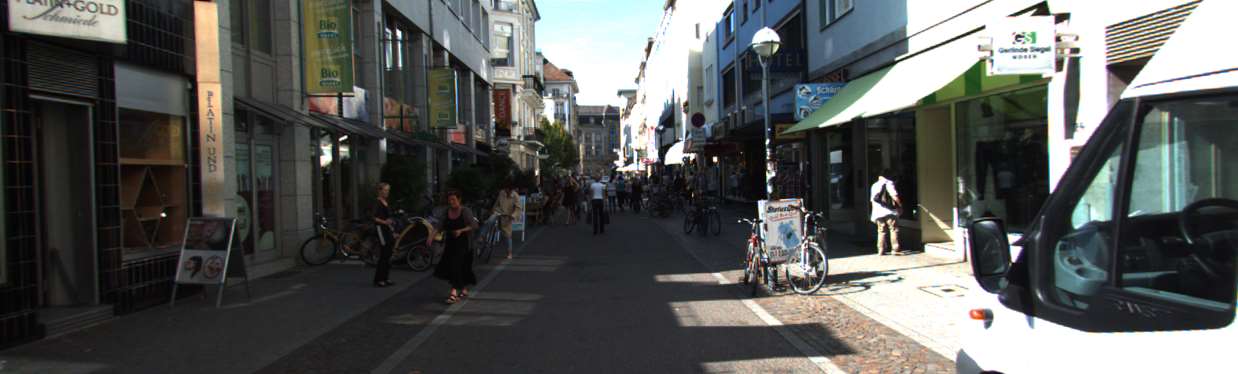} &
\includegraphics[height=\turnheightnew]{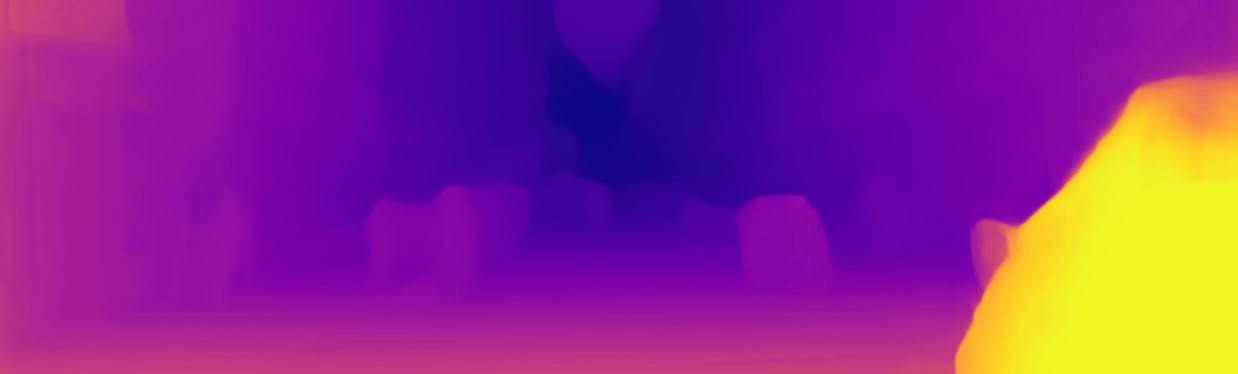} &
\includegraphics[height=\turnheightnew]{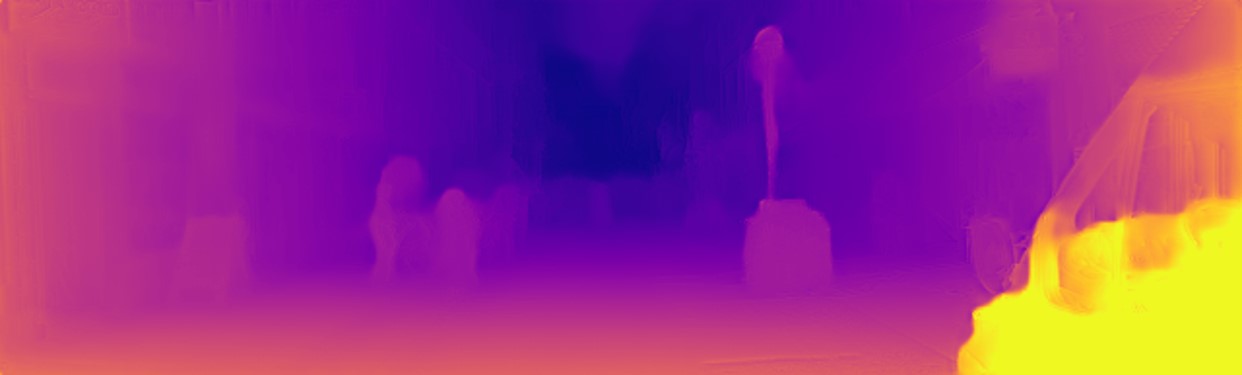} &
\includegraphics[height=\turnheightnew]{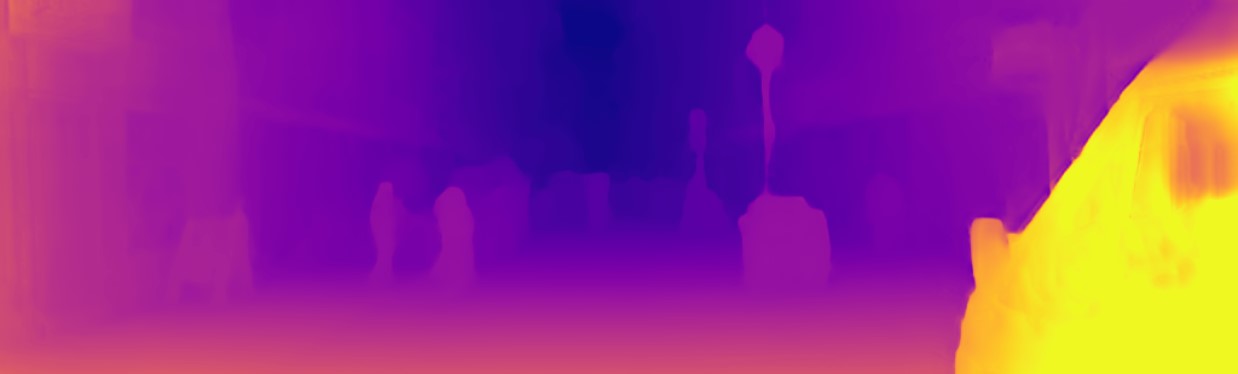}\\

\includegraphics[height=\turnheightnew]{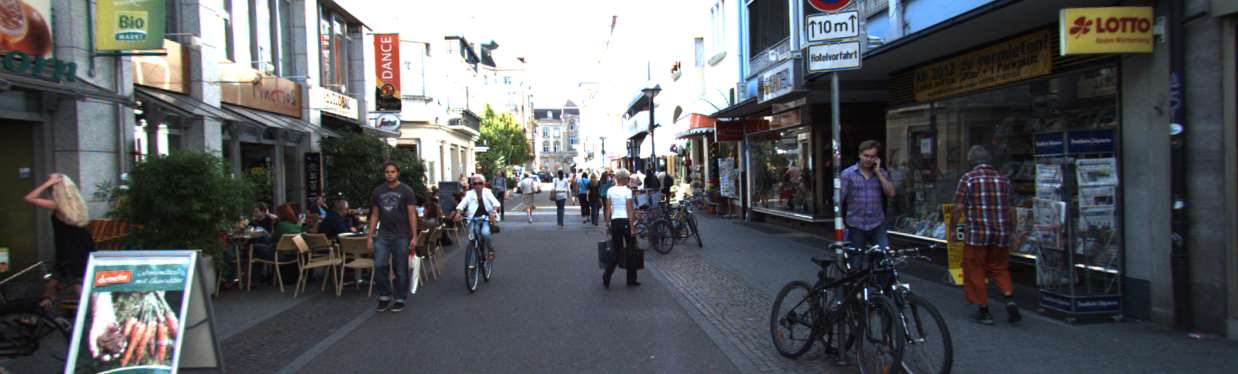} &
\includegraphics[height=\turnheightnew]{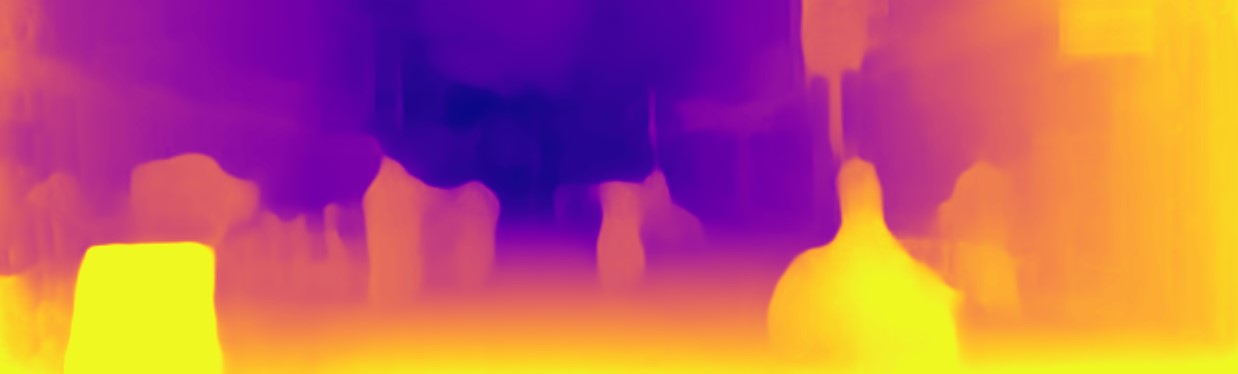} &
\includegraphics[height=\turnheightnew]{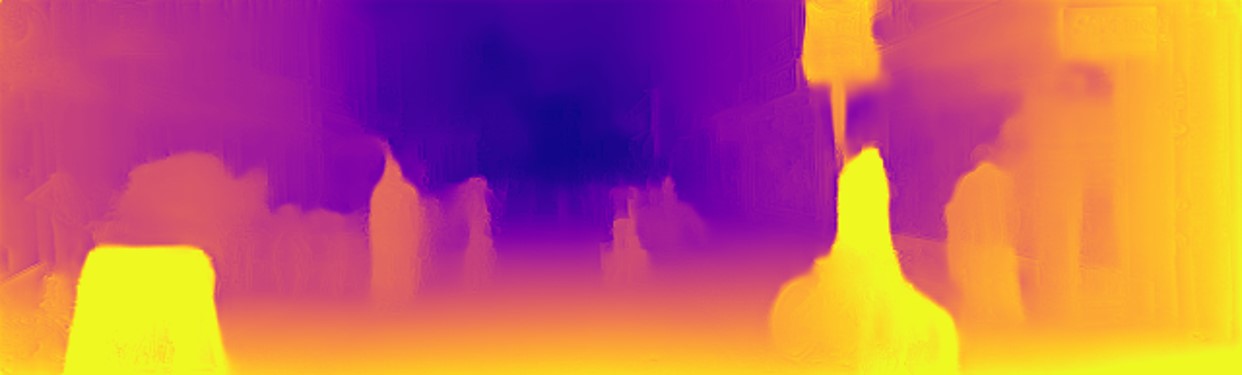} &
\includegraphics[height=\turnheightnew]{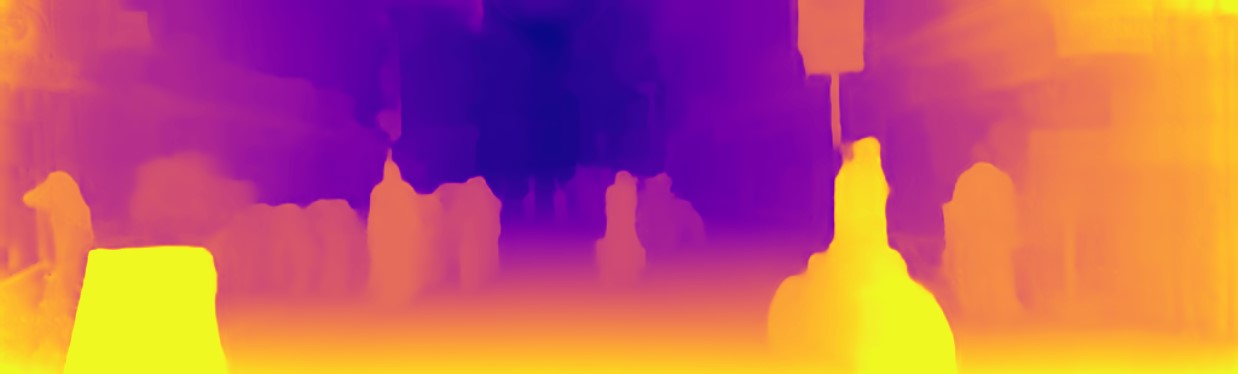}\\

\includegraphics[height=\turnheightnew]{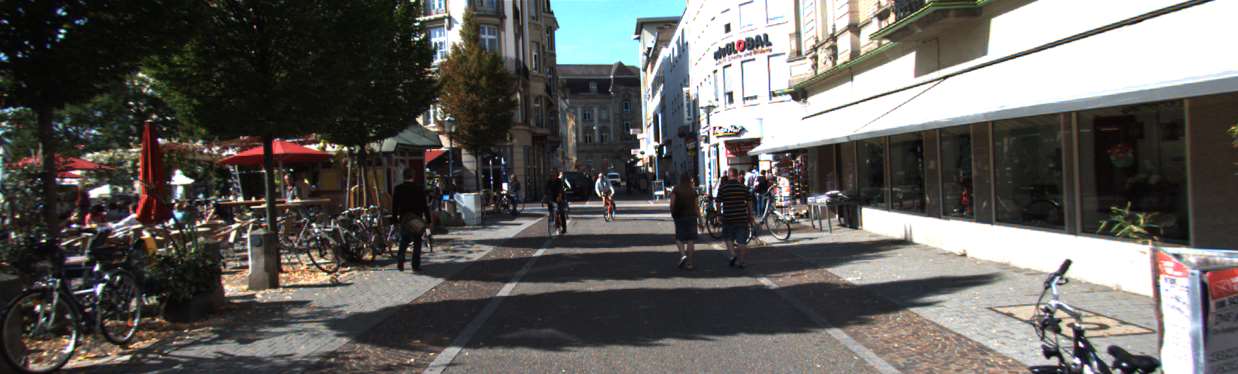} &
\includegraphics[height=\turnheightnew]{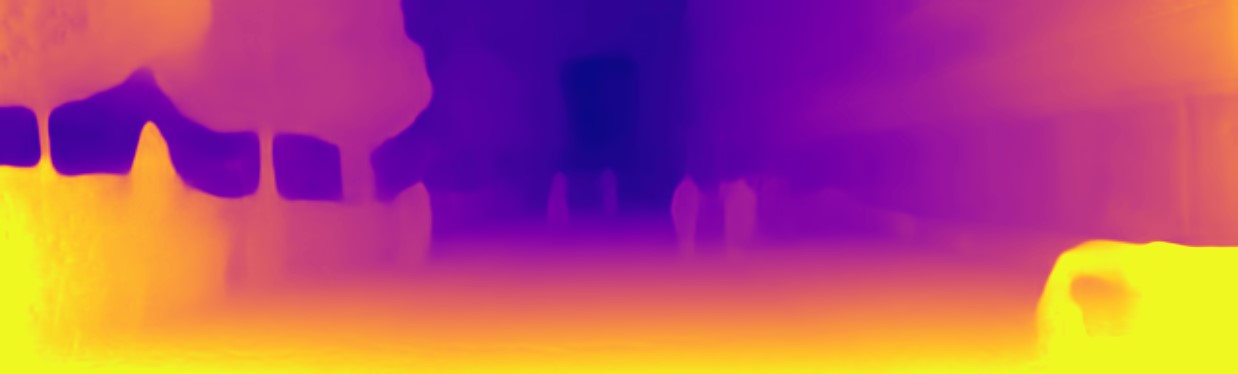} &
\includegraphics[height=\turnheightnew]{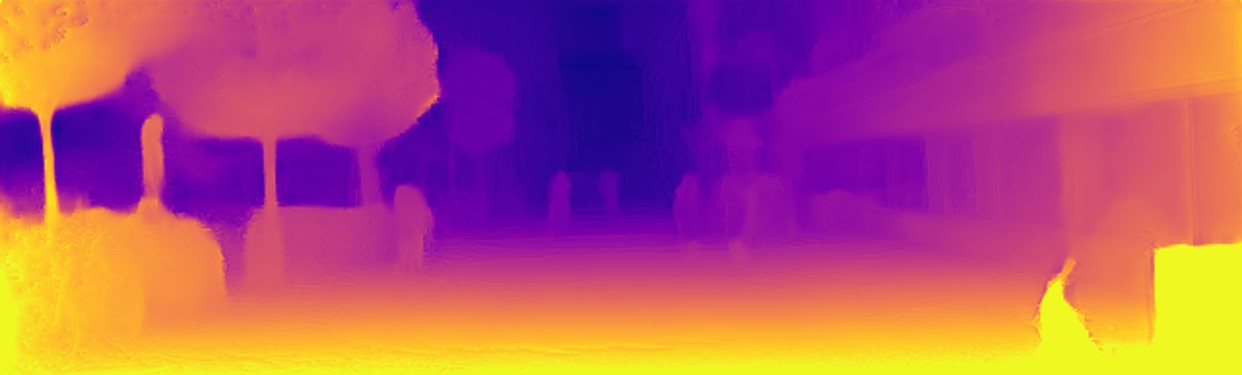} &
\includegraphics[height=\turnheightnew]{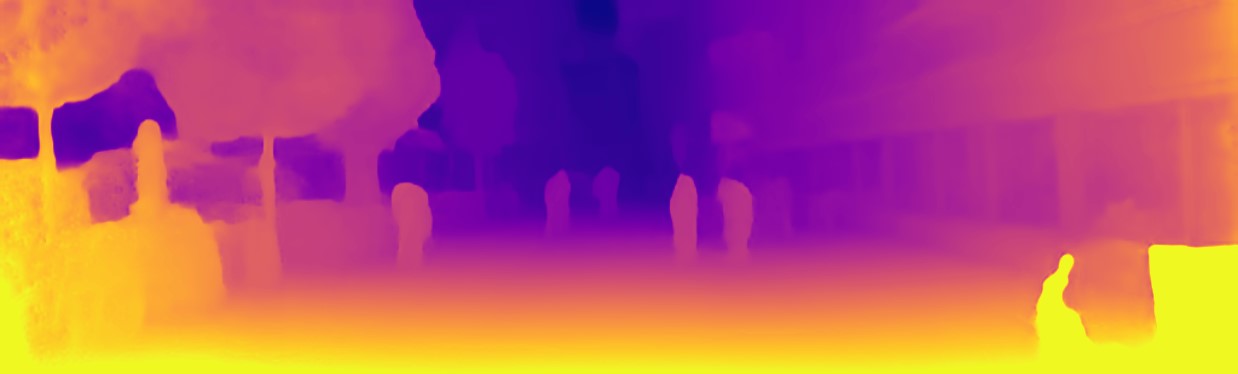}\\

\includegraphics[height=\turnheightnew]{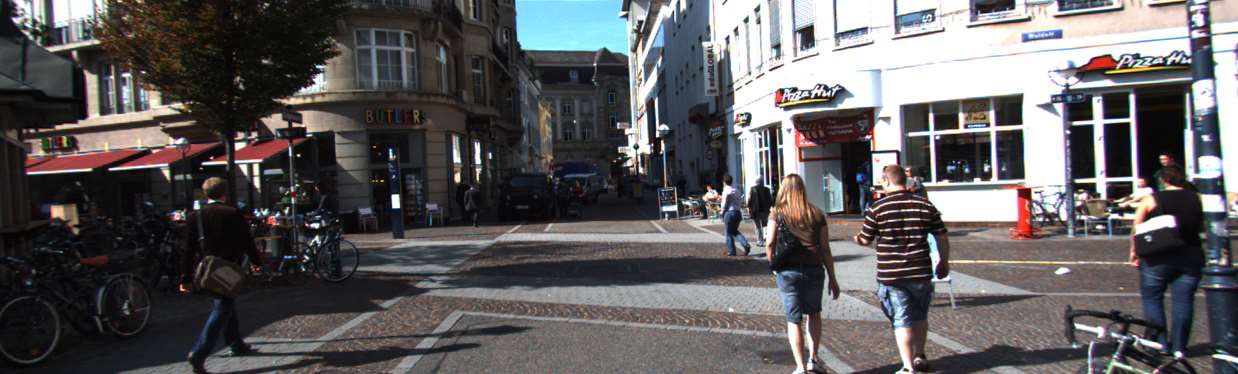} &
\includegraphics[height=\turnheightnew]{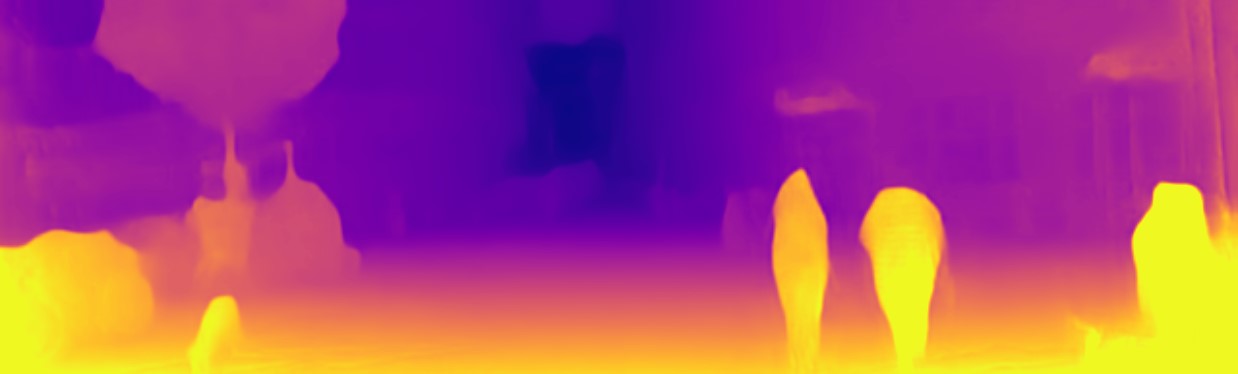} &
\includegraphics[height=\turnheightnew]{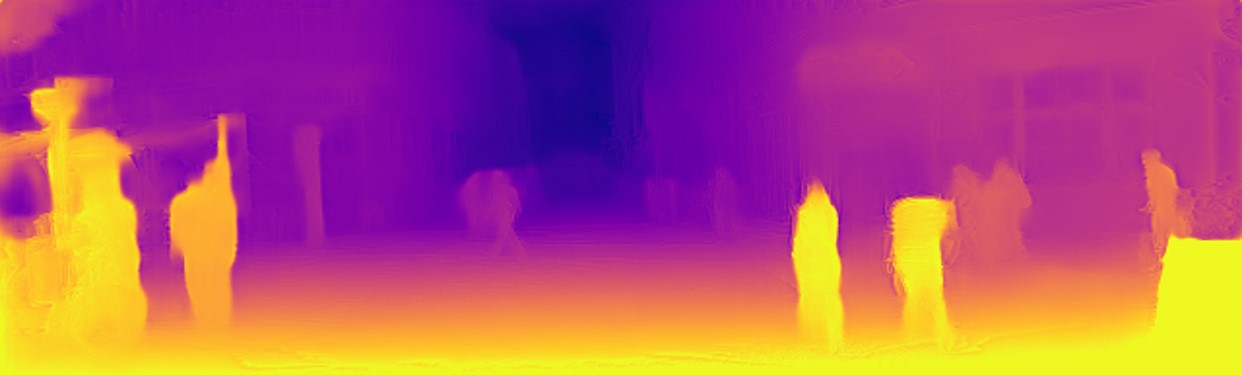} &
\includegraphics[height=\turnheightnew]{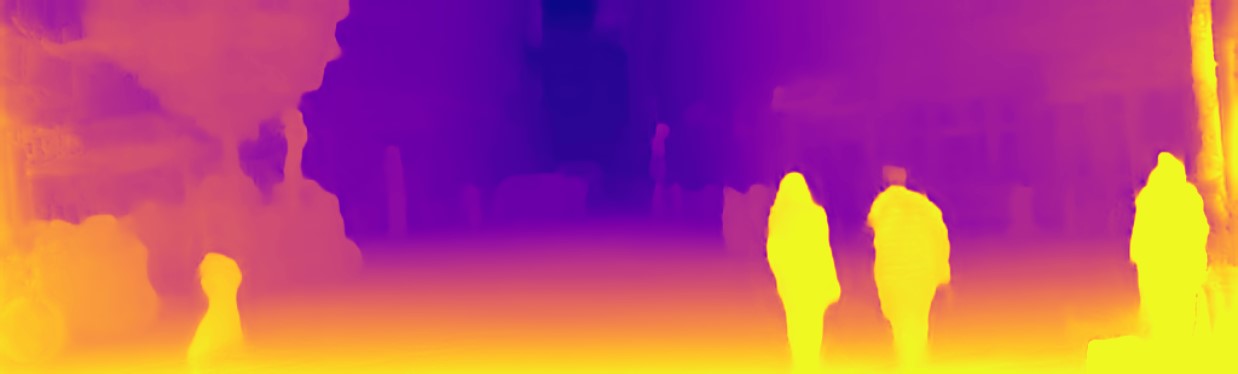}\\

\includegraphics[height=\turnheightnew]{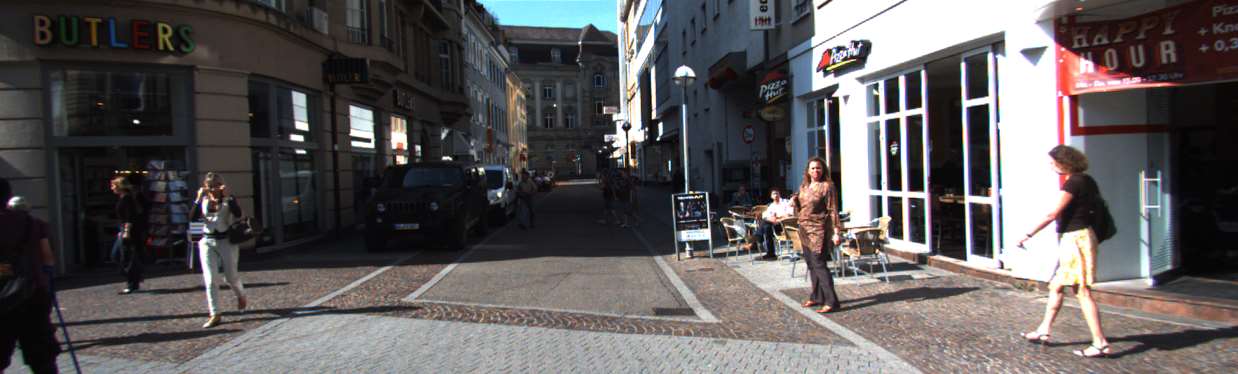} &
\includegraphics[height=\turnheightnew]{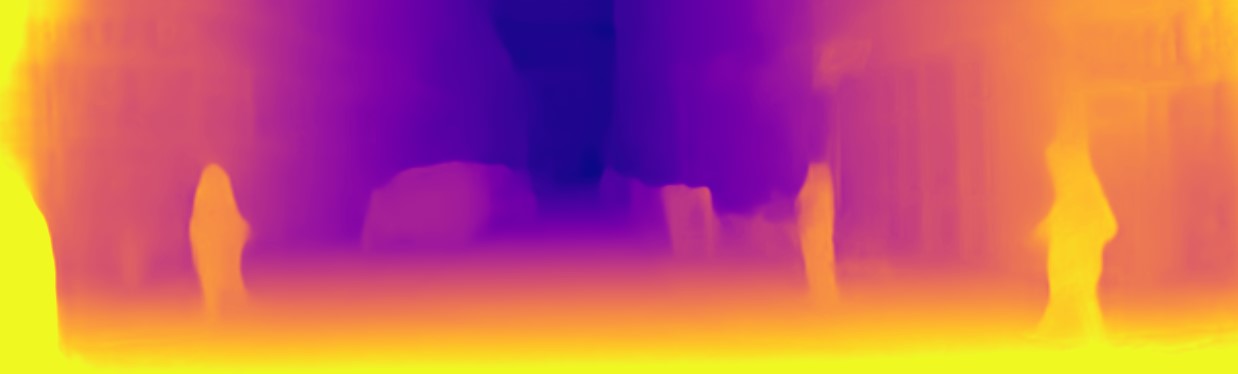} &
\includegraphics[height=\turnheightnew]{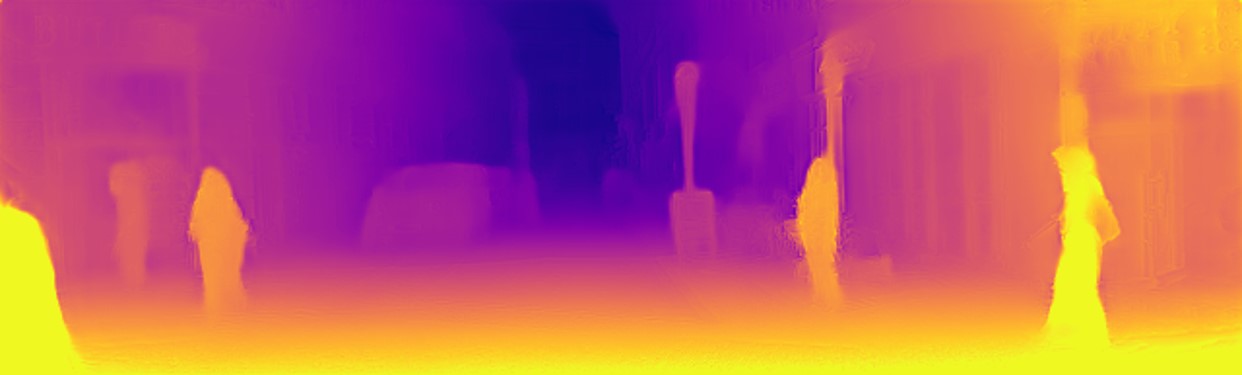} &
\includegraphics[height=\turnheightnew]{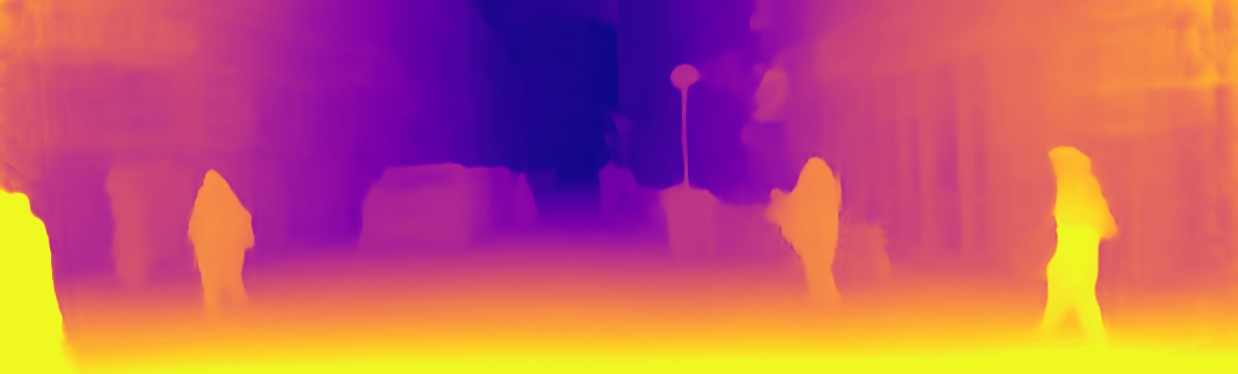}\\

\includegraphics[height=\turnheightnew]{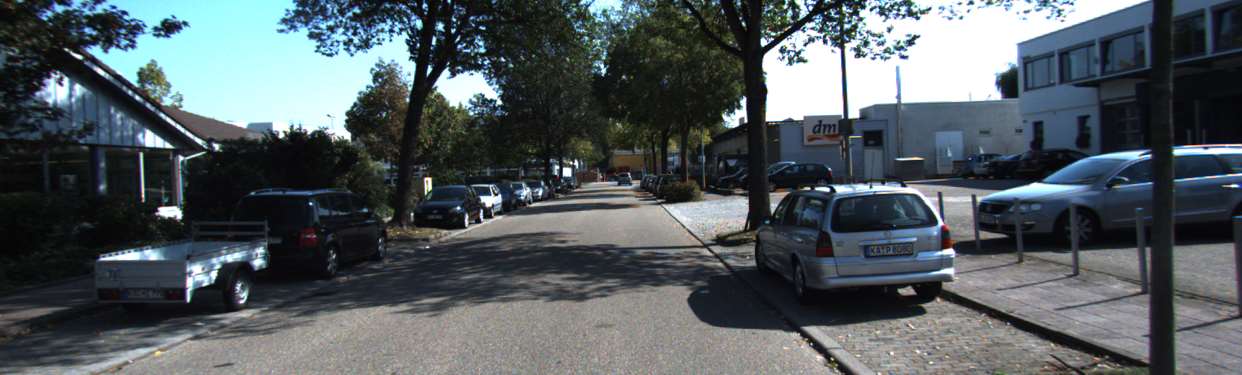} &
\includegraphics[height=\turnheightnew]{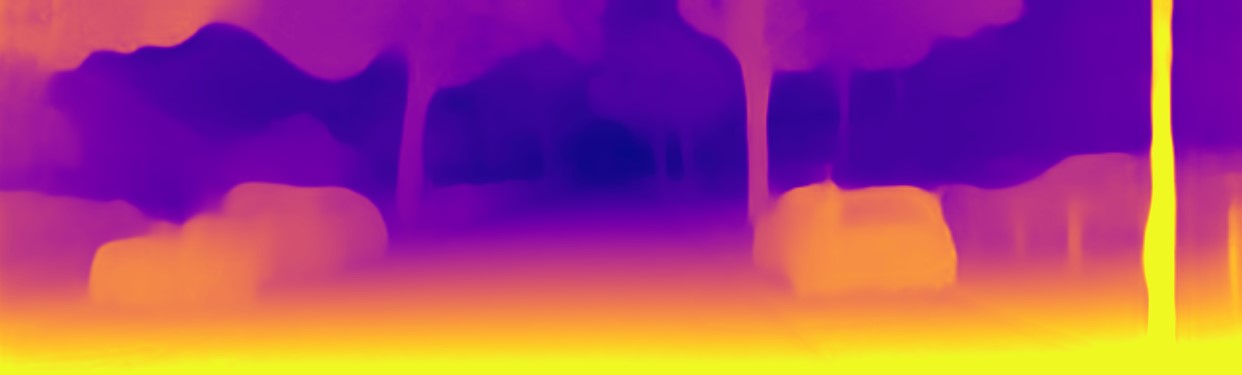} &
\includegraphics[height=\turnheightnew]{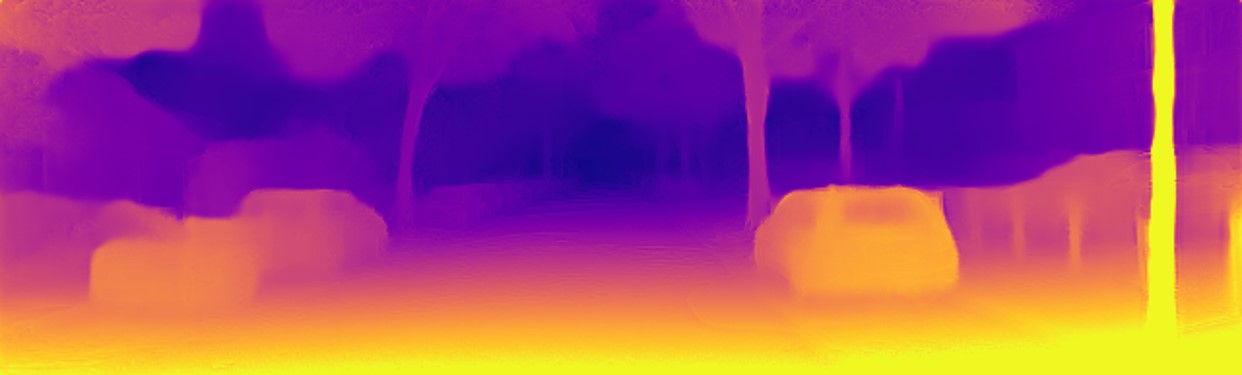} &
\includegraphics[height=\turnheightnew]{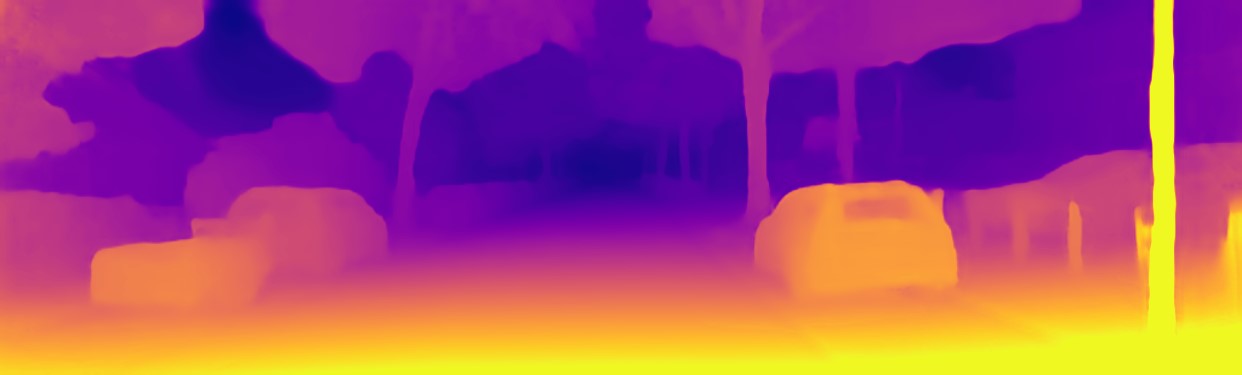}\\

\includegraphics[height=\turnheightnew]{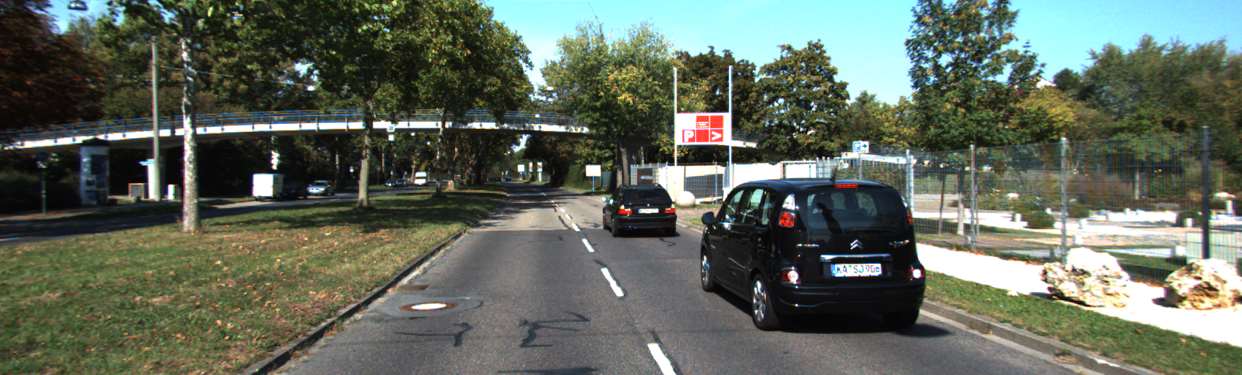} &
\includegraphics[height=\turnheightnew]{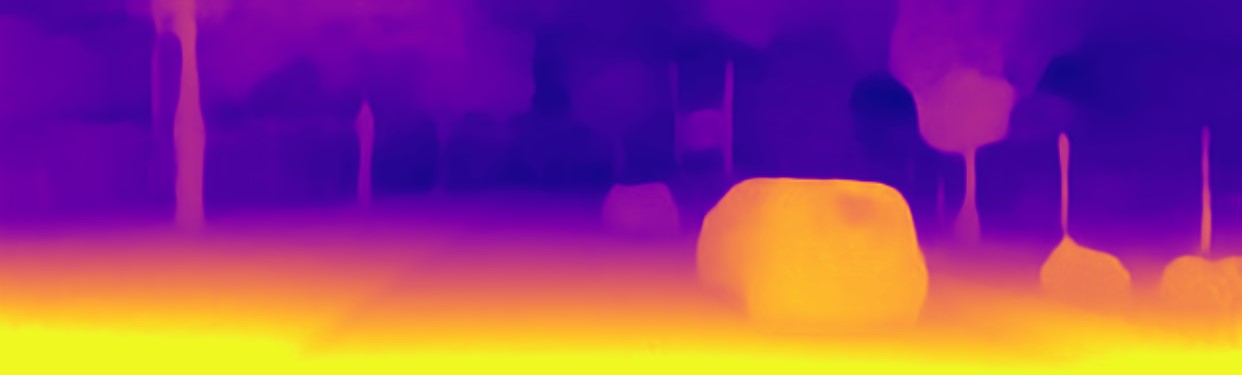} &
\includegraphics[height=\turnheightnew]{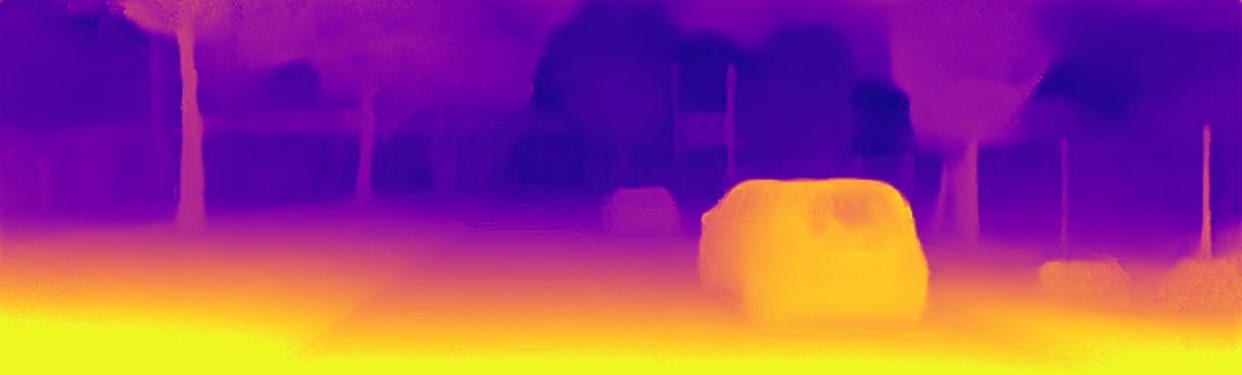} &
\includegraphics[height=\turnheightnew]{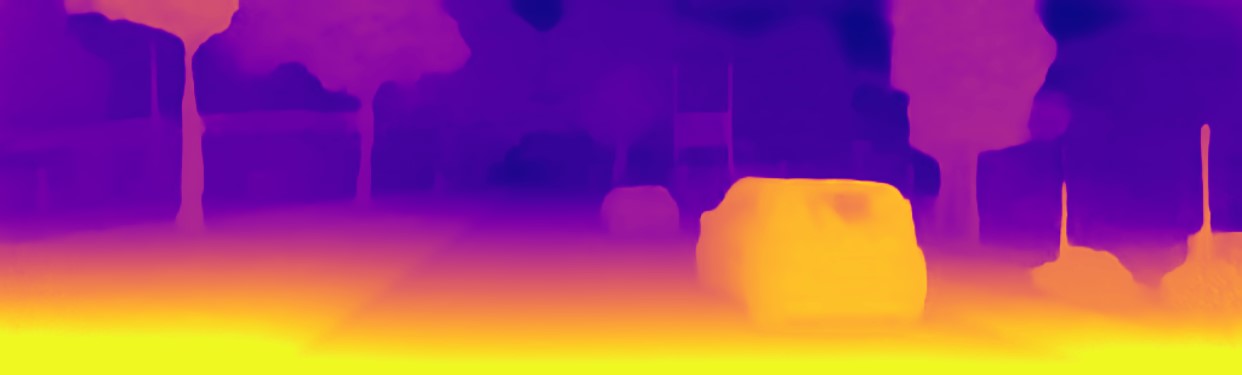}\\

\includegraphics[height=\turnheightnew]{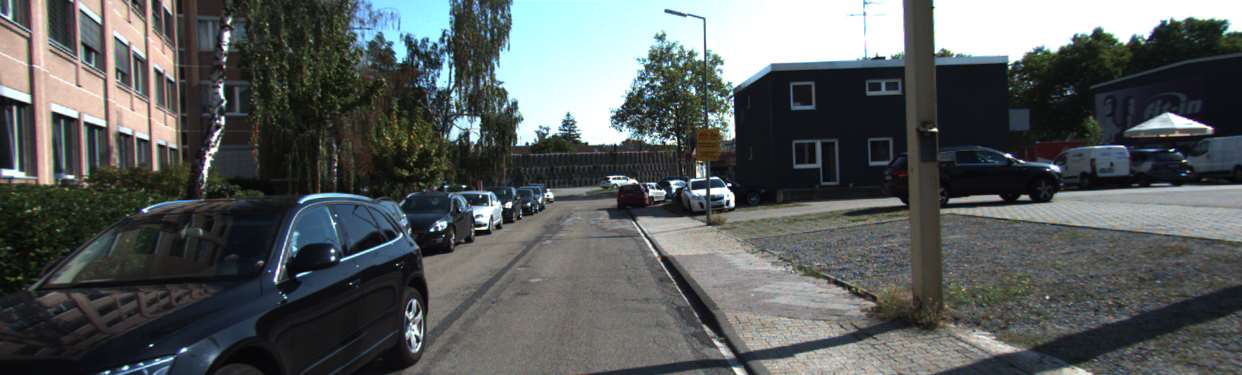} &
\includegraphics[height=\turnheightnew]{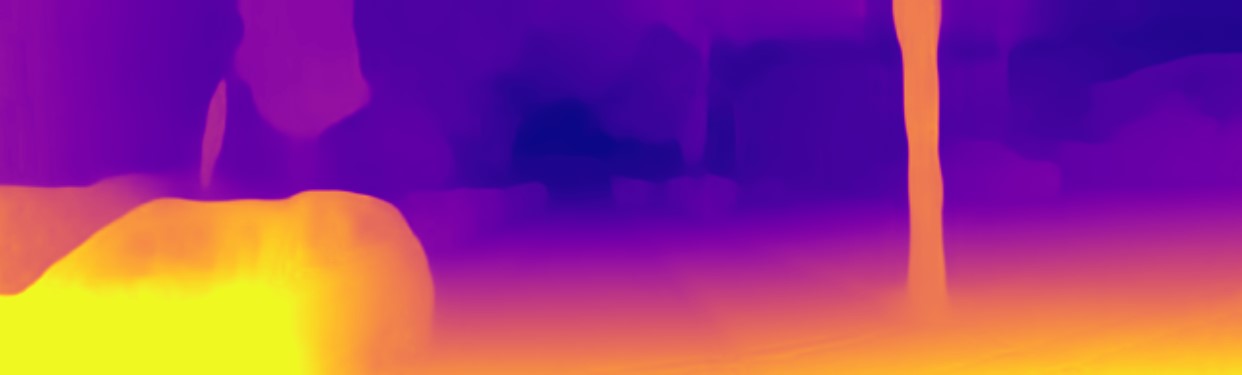} &
\includegraphics[height=\turnheightnew]{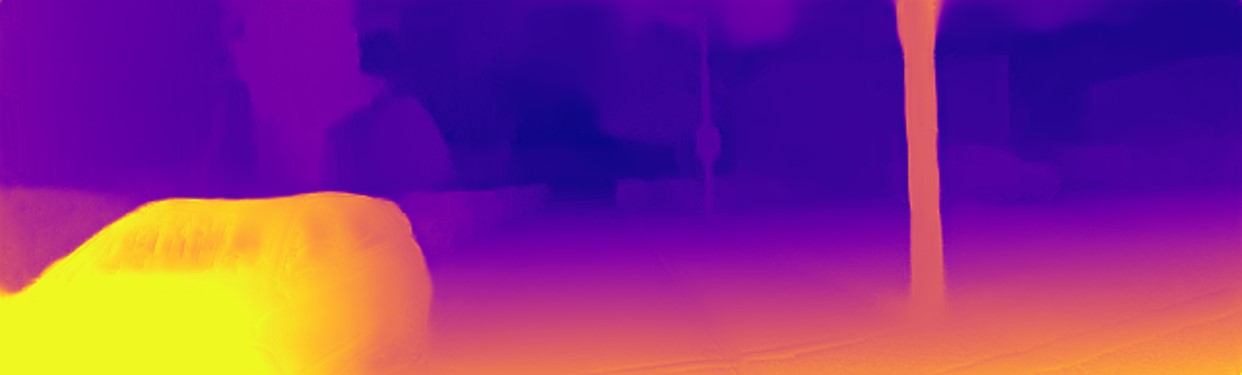} &
\includegraphics[height=\turnheightnew]{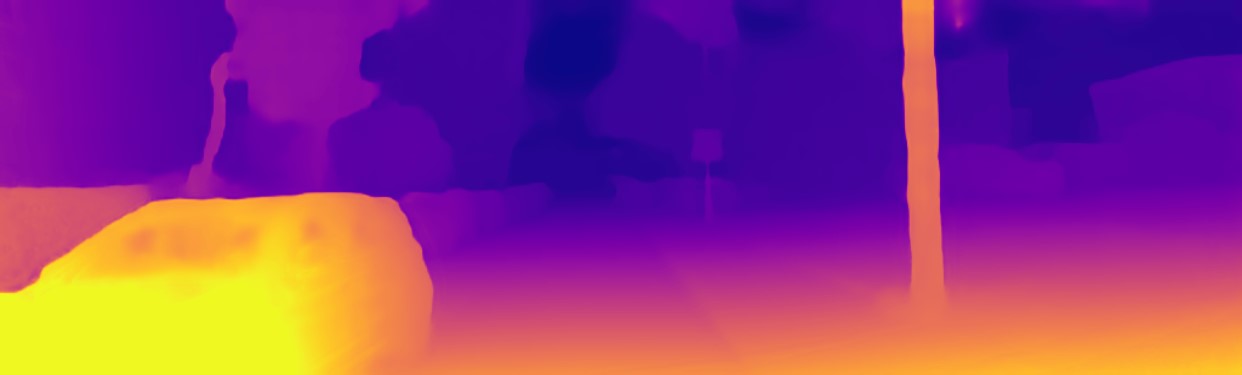}\\

\includegraphics[height=\turnheightnew]{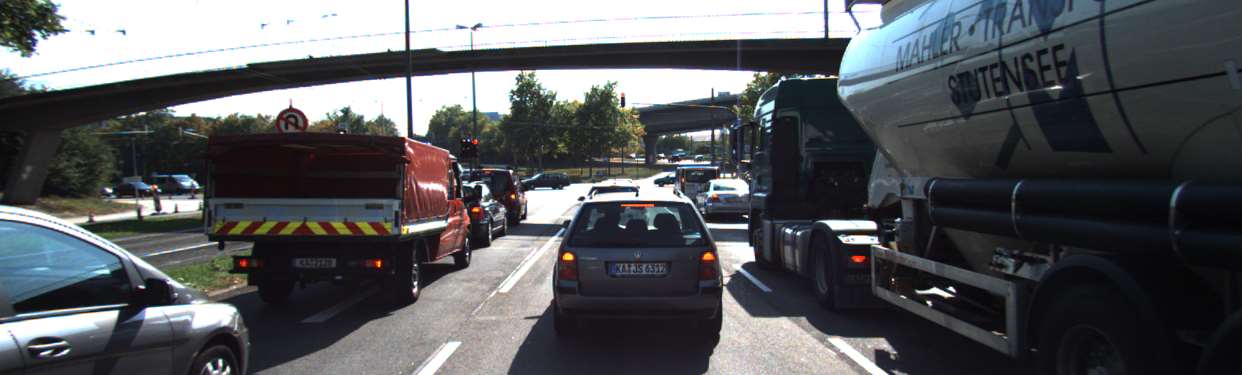} &
\includegraphics[height=\turnheightnew]{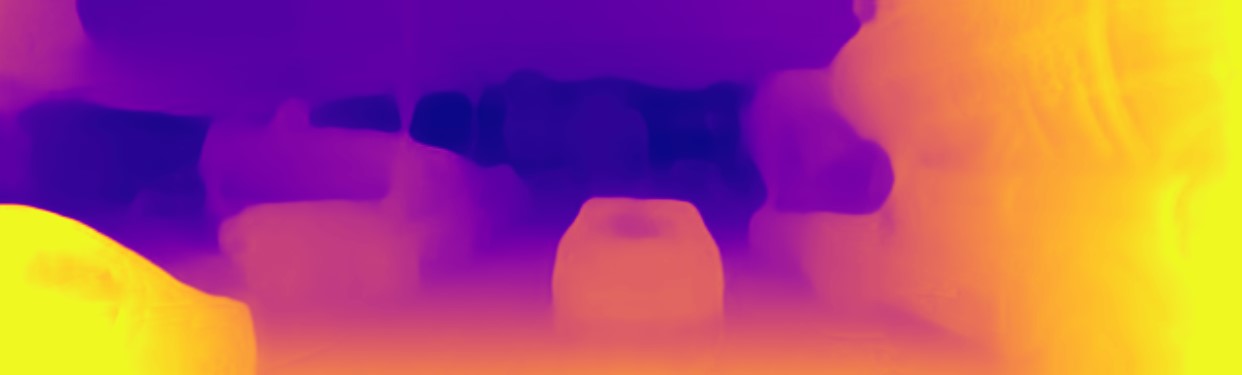} &
\includegraphics[height=\turnheightnew]{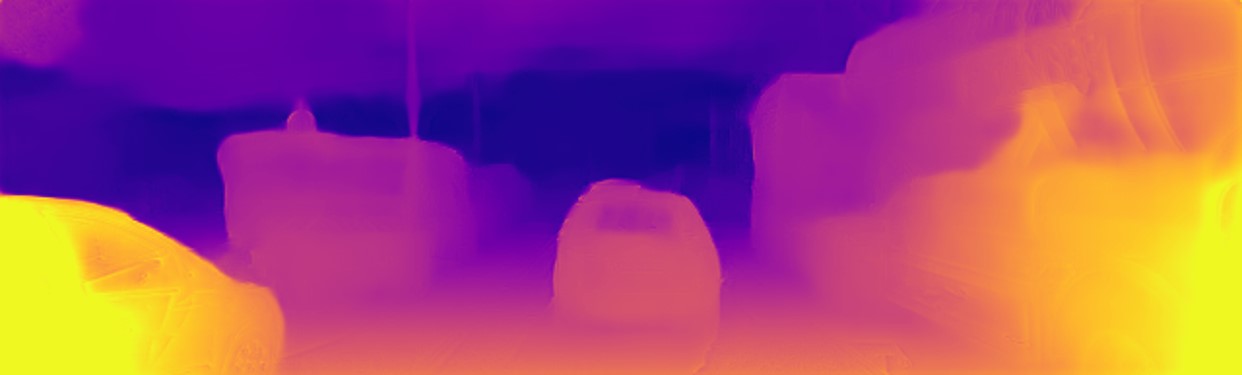} &
\includegraphics[height=\turnheightnew]{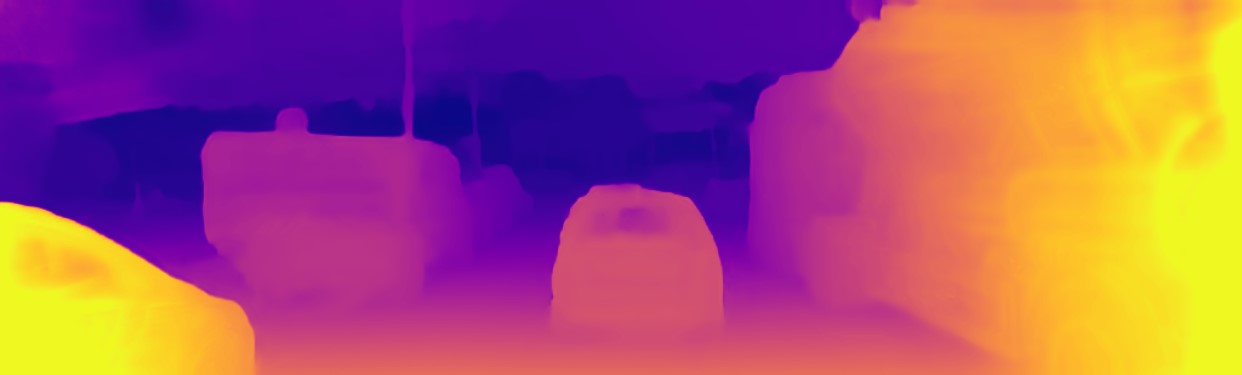}\\

\includegraphics[height=\turnheightnew]{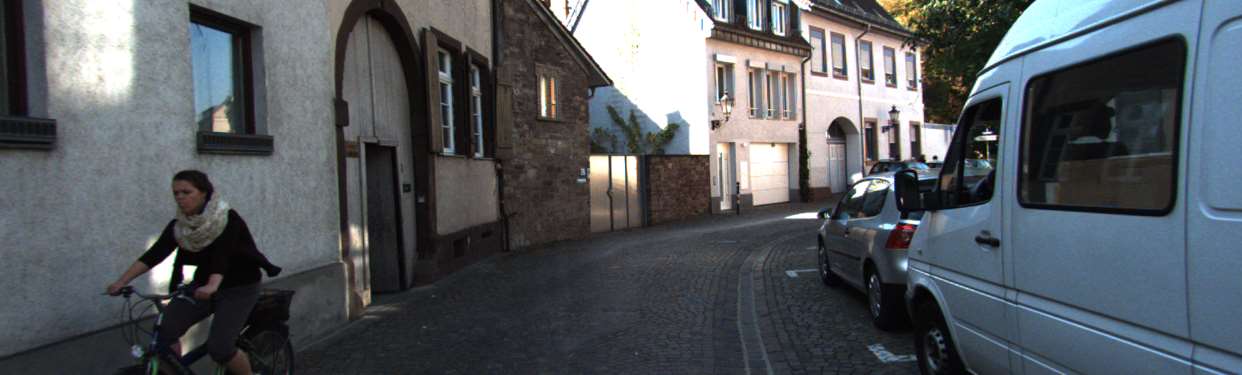} &
\includegraphics[height=\turnheightnew]{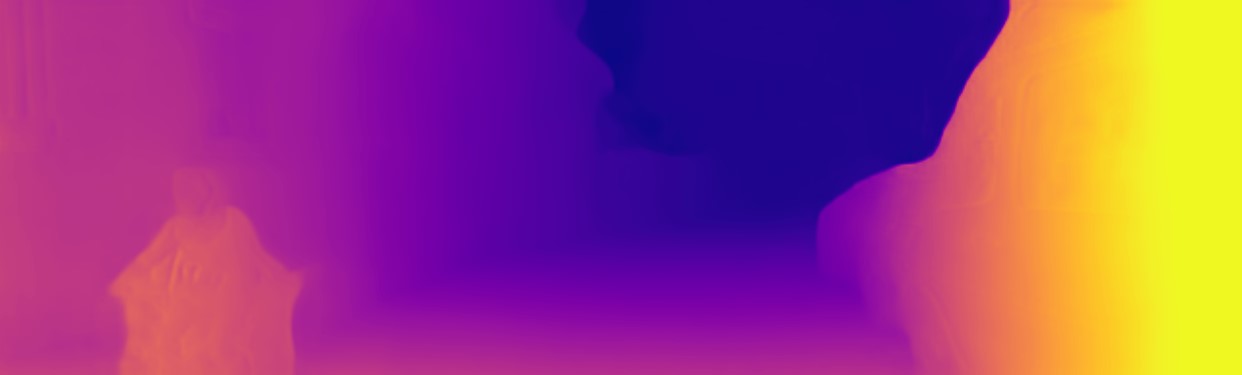} &
\includegraphics[height=\turnheightnew]{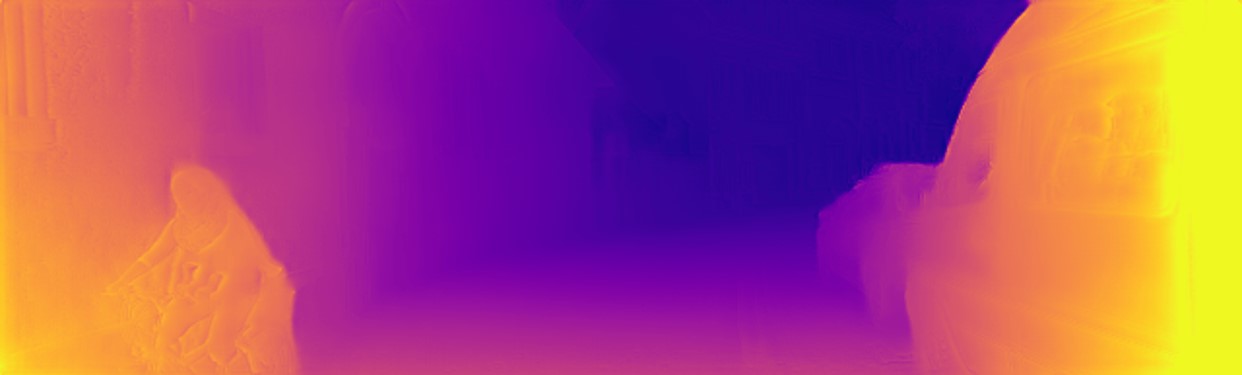} &
\includegraphics[height=\turnheightnew]{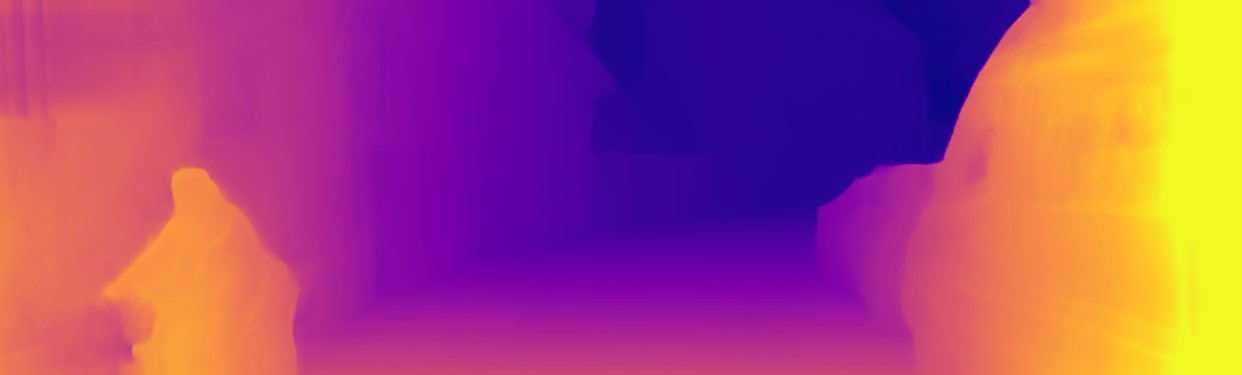}\\

\includegraphics[height=\turnheightnew]{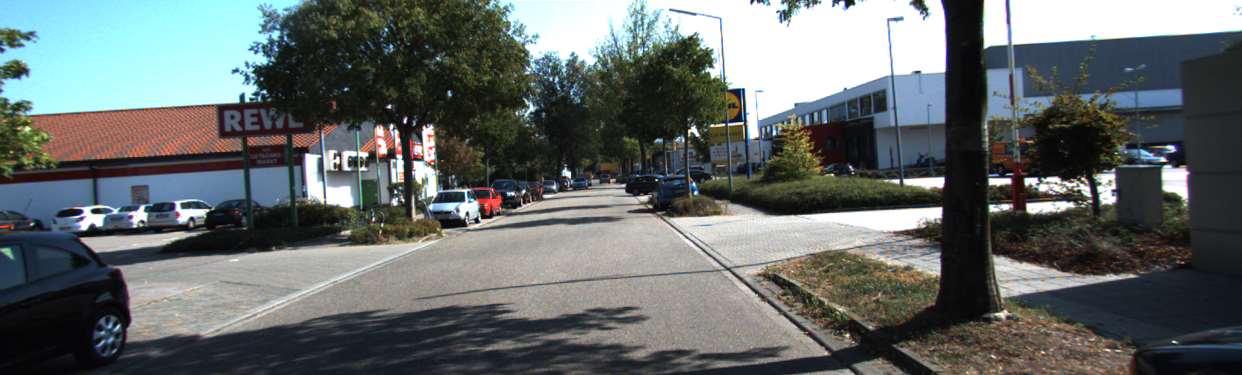} &
\includegraphics[height=\turnheightnew]{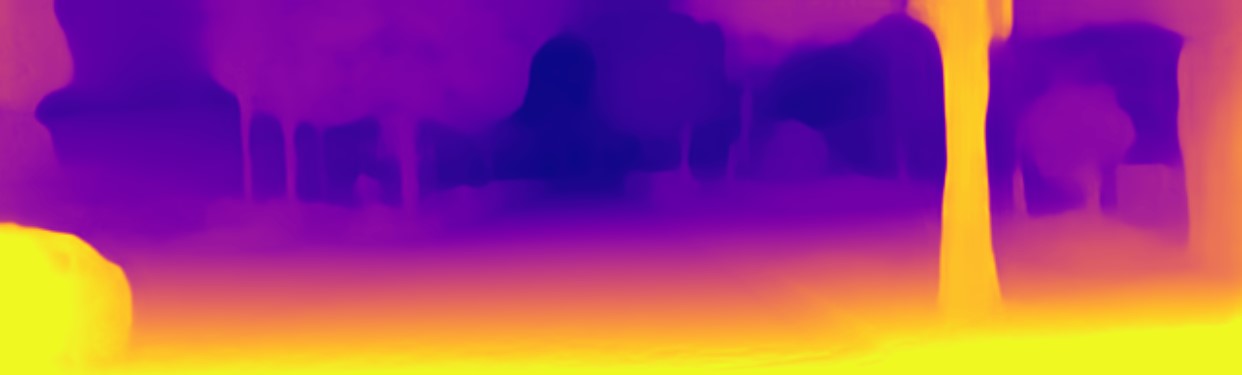} &
\includegraphics[height=\turnheightnew]{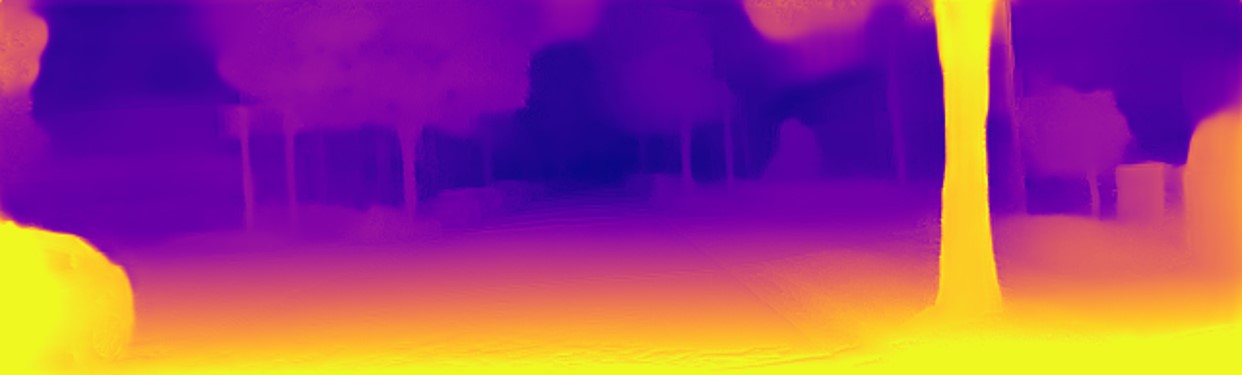} &
\includegraphics[height=\turnheightnew]{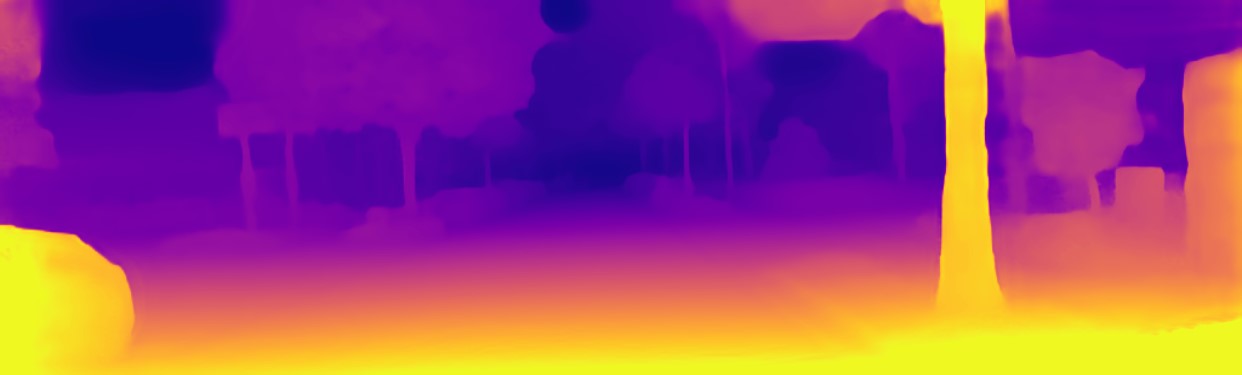}\\

\end{tabular}
 }
	\caption{\textbf{Qualitative results on the KITTI Eigen split.} Our model produces higher quality outputs and possesses the most clear border.}
	\label{fig:supp_kitti_eigen_qual}
\end{figure*}

We provide additional qualitative results from the KITTI datasets in Fig. \ref{fig:supp_kitti_eigen_qual}. We can observe that compared to existing baselines \eg~\cite{godard2019digging, guizilini20203d}, our CADepth-Net produces higher quality outputs and possesses the clearest border overall. We also show additional results from Make3D datasets~\cite{saxena2008make3d} in Fig. \ref{fig:supp_make3d_results}, our methods preserve sharp discontinuities in depth prediction results.

\section{Additional Visualization Results}

\begin{figure*}
	\centering
	\includegraphics[width=\linewidth]{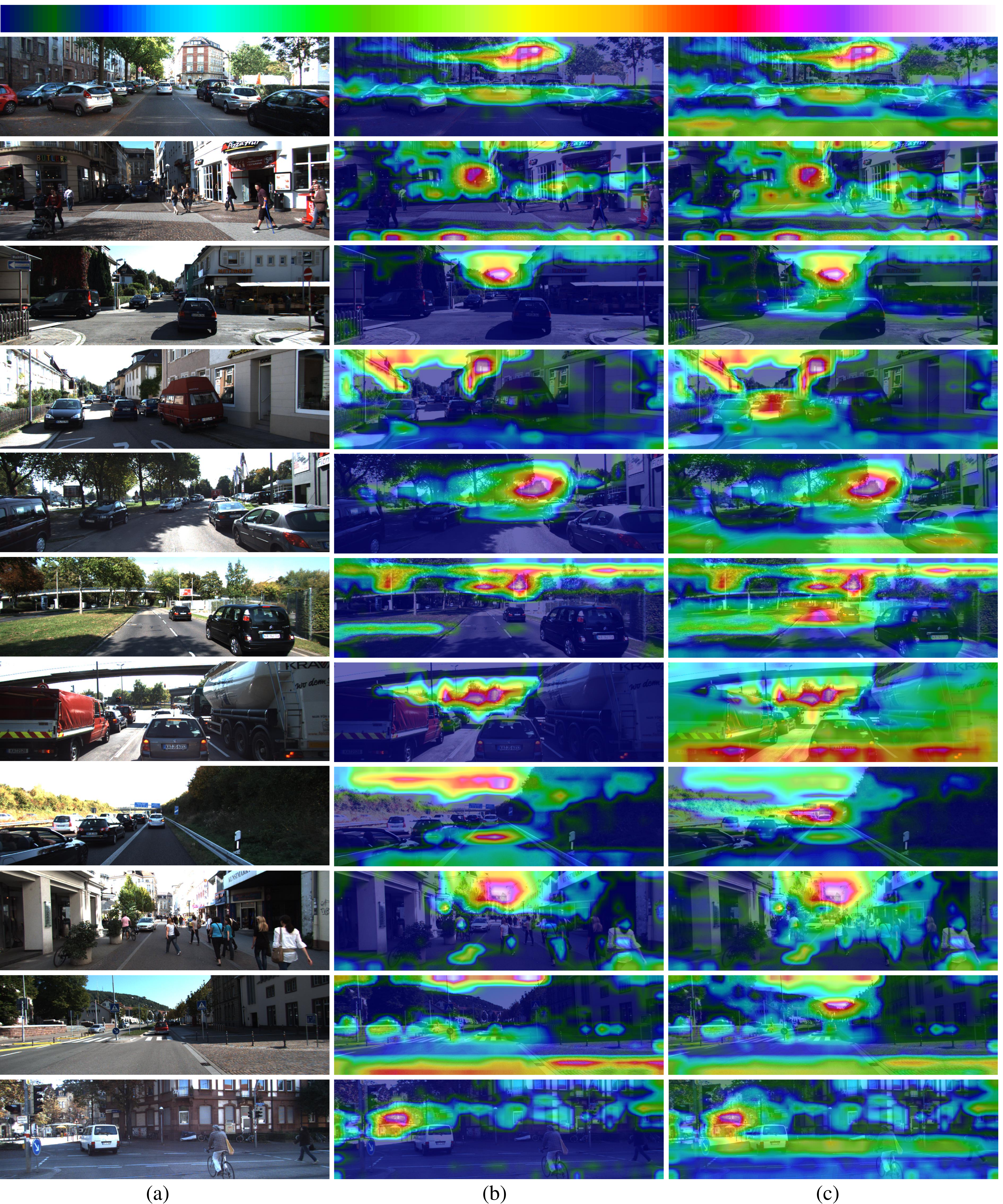} 
	\caption{\textbf{The visualizations of structure perception module.} (a) Input Image. (b) Input feature maps. (c) Output feature maps. All feature representation are projected onto original image for clear visualizations. Our structure perception module explicitly enhances the scene understanding and feature representation.
	}
	\label{fig:supp_SPM_vis}
\end{figure*}

\begin{figure*}
	\centering
	\includegraphics[width=\linewidth]{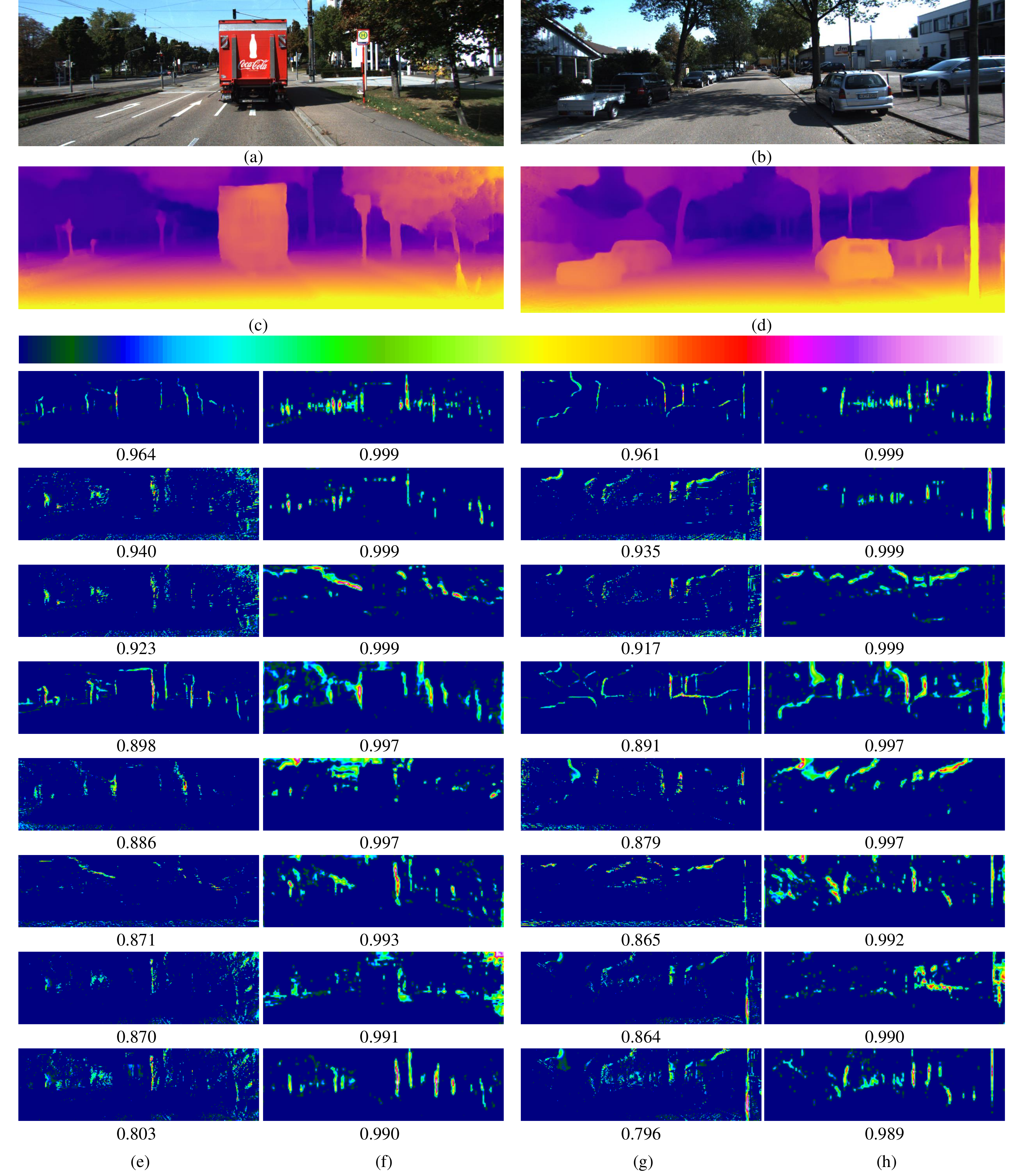} 
	\caption{\textbf{The visualizations of detail emphasis module.} (a)(b) Input Image. (c)(d) Predicted depth maps. (e) $\sim$ (h) Top $n$ feature maps with highest weights score. Here $n$ is 8 and the weights value range is between 0 and 1. Specifically, (e)(g) are produced by $dem2$ (see Table \ref{tab:network}) and (f)(h) are generated from $dem3$. The detail emphasis module mainly highlights the crucial local details. 
	}
	\label{fig:supp_DEM_vis}
\end{figure*}

To better understand our main contributions, Fig. \ref{fig:supp_SPM_vis} and Fig. \ref{fig:supp_DEM_vis} introduce additional visualization results of intermediate features. As shown in Fig. \ref{fig:supp_SPM_vis}, each feature map obtains more region responses from the distant regions and aggregates relative depth relationships over 3D scene. By doing so, our model produces better scene understanding and rich feature representation. Fig. \ref{fig:supp_DEM_vis} lists the top $n$ ($n=8$) feature maps with the highest scores in the detail emphasis module, and we can see that our model highlights critical local details at multiple scales, by assigning higher scores to them.

\end{appendices}

\end{document}